\documentclass[journal,twoside]{IEEEtran} 
\usepackage[acronym,nonumberlist]{glossaries}
\usepackage{amsmath}
\usepackage{amssymb}
\usepackage{xfrac}
\usepackage{mathtools}
\usepackage{relsize}
\newcommand{\R}{\mathbb{R}}
\usepackage{xspace}
\usepackage[shortlabels]{enumitem}
\usepackage{hyperref}
\usepackage{soul}

\usepackage{threeparttable}

\usepackage{calrsfs}
\DeclareMathAlphabet{\pazocal}{OMS}{zplm}{m}{n}
\newcommand{\gammassl}{$\mathrm{\gamma}\text{-}\mathrm{SSL}$\xspace}

\DeclarePairedDelimiterX\set[1]\lbrace\rbrace{#1}

\newcommand{\nossl}{$\mathrm{NoSSL}$\xspace}
\newcommand{\nofiltering}{$\mathrm{M_{\gamma=-\infty}}$\xspace}
\newcommand{\nosax}{$\mathrm{NoSAX}$\xspace}

\newcommand{\Amd}{$\mathrm{A_{MD}}$\xspace}
\newcommand{\maxAmd}{$\mathrm{Max\mathrm{A_{MD}}}$\xspace}

\newcommand{\aupr}{$\mathrm{AUPR}$\xspace}
\newcommand{\auroc}{$\mathrm{AUROC}$\xspace}
\newcommand{\tp}{$\mathrm{TP}$\xspace}
\newcommand{\fn}{$\mathrm{FN}$\xspace}
\newcommand{\tn}{$\mathrm{TN}$\xspace}
\newcommand{\fp}{$\mathrm{FP}$\xspace}
\newcommand{\fhalf}{$\mathrm{F_{0.5}}$\xspace}
\newcommand{\maxfhalf}{$\mathrm{MaxF_{0.5}}$\xspace}
\newcommand{\certain}{$\mathrm{certain}$\xspace}
\newcommand{\uncertain}{$\mathrm{uncertain}$\xspace}
\newcommand{\pac}{$\mathrm{p(a,c)}$\xspace}
\newcommand{\gammasslil}{$\mathrm{\gamma}\text{-}\mathrm{SSL_{iL}}$\xspace}

\newcommand{\mcd}{$\mathrm{MCD}$\xspace}
\newcommand{\ensemble}{$\mathrm{Ens}$\xspace}

\newcommand{\softmax}{$\mathrm{Softmax}$\xspace}
\newcommand{\softmaxA}{$\mathrm{Softmax_A}$\xspace}
\newcommand{\featdist}{$\mathrm{FeatDist}$\xspace}
\newcommand{\featdistA}{$\mathrm{FeatDist_A}$\xspace}
\newcommand{\vim}{$\mathrm{ViM}$\xspace}
\newcommand{\dum}{$\mathrm{DUM}$\xspace}

\newcommand{\enspefive}{$\mathrm{Ens}$-$\mathrm{PE}_{5}$\xspace}
\newcommand{\ensmifive}{$\mathrm{Ens}$-$\mathrm{MI}_{5}$\xspace}
\newcommand{\enspeten}{$\mathrm{Ens}$-$\mathrm{PE}_{10}$\xspace}
\newcommand{\ensmiten}{$\mathrm{Ens}$-$\mathrm{MI}_{10}$\xspace}
\newcommand{\mcdpepointtwo}{$\mathrm{MCD}$-$\mathrm{PE}_{0.2}$\xspace}
\newcommand{\mcdmipointtwo}{$\mathrm{MCD}$-$\mathrm{MI}_{0.2}$\xspace}
\newcommand{\mcddistil}{$\mathrm{MCD}$-$\mathrm{DSL}$\xspace}

\usepackage{multirow}
\usepackage{multicol}

\usepackage{glossary-mcols}
\setglossarystyle{mcolindex}

\newacronym{ood}{OoD}{Out-of-Distribution}
\newacronym{uda}{UDA}{Unsupervised Domain Adaptation}
\newacronym{dum}{DUM}{Deterministic Uncertainty Method}
\newacronym{gan}{GAN}{Generative Adversarial Network}
\newacronym{mi}{MI}{mutual information}
\newacronym{pe}{PE}{predictive entropy}
\newacronym{roc}{ROC}{receiver operating characteristic}
\newacronym{pr}{PR}{precision-recall}
\newacronym{odd}{ODD}{operational design domain}
\newacronym{av}{AV}{autonomous vehicle}
\newacronym{mcd}{MCD}{Monte Carlo Dropout}
\newacronym{rbf}{RBF}{radial basis function}
\newacronym{bdd}{BDD}{Berkeley DeepDrive 10k}

\usepackage[binary-units]{siunitx}
\usepackage{eucal}
\usepackage{bbm}
\usepackage{graphicx}
\usepackage{algorithm}
\usepackage{algpseudocode}
\algrenewcommand\textproc{\texttt}      % define font for function name in algorithm
\usepackage{bbm}
\usepackage{amsmath}
\usepackage{stackengine}
\makeatletter
\let\MYcaption\@makecaption
\makeatother

\usepackage[font=footnotesize]{subcaption}

\makeatletter
\let\@makecaption\MYcaption
\makeatother

\usepackage{cuted}

\usepackage{cleveref}
\crefname{table}{Tab.}{Tabs.}
\crefname{figure}{Fig.}{Figs.}
\crefname{section}{Sec.}{Secs.}
\crefname{equation}{Eq.}{Eqs.}

\usepackage[usernames, dvipsnames]{xcolor}
\IEEEoverridecommandlockouts
\usepackage[noadjust]{cite}

\usepackage{ifthen}
\newboolean{review}
\setboolean{review}{false} 

\ifthenelse{\boolean{review}}{
\usepackage{ulem}

\newcommand{\fix}[1]{{\color{blue}#1}}
\newcommand{\fixS}[1]{{\color{blue}\sout{#1}}}
\newcommand{\fixM}[2]{\fix{#1\marginpar{\color{red}#2}}}
}{
\newcommand{\fix}[1]{#1}
\newcommand{\fixS}[1]{}
\newcommand{\fixM}[2]{#1}
}

\let\emph\textit

\begin{document}

\ifthenelse{\boolean{review}}{
\pagenumbering{roman}
\clearpage
\newpage
\input{replies/reply_revs}
}{}

\pagenumbering{arabic}
%------------------------------------------------------------------
% 	\title{\Large \bf Unsupervised Domain Adaptation for Selective Semantic Segmentation}
% \title{\Huge \bf Semi-Supervised Prototype Networks for Learned Uncertainty Estimation}
\title{
% $\gamma$-SSL: Mitigating Distributional Shift in Semantic Segmentation via Semi-Supervised Uncertainty Estimation}
% $\gamma$-SSL: Mitigating Distributional Shift in Semantic Segmentation via Uncertainty Estimation from Unlabelled Data}
% Uncertainty Estimation for Semantic Segmentation from Unlabelled Data}
\huge{
Mitigating Distributional Shift in Semantic Segmentation via Uncertainty Estimation from Unlabelled Data
}
}
% $\gamma$-SSL: Uncertainty Estimation from Unlabelled Data in Semantic Segmentation}
% \title{\huge \bf $\gamma$-SSL: Semi-Supervised Uncertainty Estimation for Mitigating Distributional Shift in Semantic Segmentation}
\author{David S. W. Williams$^*$, Daniele De Martini, Matthew Gadd, and Paul~Newman\\
% \thanks{Manuscript received \today.}
\thanks{This work was supported by EPSRC Programme Grant EP/M019918/1.
The authors would like to acknowledge the use of Hartree Centre resources and the University of Oxford Advanced Research Computing (ARC) facility in carrying out this work {\tt\small{}http://dx.doi.org/10.5281/zenodo.22558}.
}
\thanks{The authors are with the Oxford Robotics Institute, Department of Engineering Science, University of Oxford, UK. Emails: \texttt{\{dw,daniele,mattgadd,pnewman\}@robots.ox.ac.uk}}
\thanks{$^*$\footnotesize{\textit{Corresponding author}}}
% 	\thanks{
% 		$^*$ These authors contributed equally to this work.
% 		% $^{\dagger}$ Supported by the Assuring Autonomy International Programme, a partnership between Lloyd’s Register Foundation and the University of York, and UK EPSRC Programme Grant EP/M019918/1.
% 	}
}
\maketitle
%------------------------------------------------------------------
% \copyrightnotice
% \renewcommand{\baselinestretch}{0.97}
%------------------------------------------------------------------

\begin{abstract}
Knowing when a trained segmentation model is encountering data that is different to its training data is important. 
% It is important to know when a trained segmentation model is experiencing data that is different to its training data, known as \textit{distributional shift}.
Understanding and mitigating the effects of this play an important part in their application from a performance and assurance perspective -- this being a safety concern in applications such as \glspl{av}.
% Distributional shift greatly decreases the performance of segmentation networks and thus must be mitigated for safe deployment into diverse real-world environments.
This work presents a segmentation network that can detect errors caused by challenging test domains without any additional annotation in a single forward pass.
% As annotation costs prevent obtaining labels for many possible deployment domains, we use easy-to-obtain, uncurated unlabelled data and learn to perform uncertainty estimation by selectively enforcing consistency over data augmentation.
As annotation costs limit the diversity of labelled datasets, we use easy-to-obtain, uncurated and unlabelled data to learn to perform uncertainty estimation by selectively enforcing consistency over data augmentation.
% To this end, a novel segmentation benchmark based on the SAX Dataset, is used, which includes unlabelled \emph{training} data and labelled \emph{test} data, spanning three autonomous-driving domains, ranging in appearance from dense urban to off-road.
To this end, a novel segmentation benchmark based on the SAX Dataset is used, which includes labelled \emph{test} data spanning three autonomous-driving domains, ranging in appearance from dense urban to off-road.
The proposed method, named \gammassl, consistently outperforms uncertainty estimation and \gls{ood} techniques on this difficult benchmark -- by up to \SI{10.7}{\%} in area under the \gls{roc} curve and \SI{19.2}{\%} in area under the \gls{pr} curve in the most challenging of the three scenarios.
\end{abstract}


\glsresetall

%------------------------------------------------------------------
\section{Introduction} \label{sec:introduction}

\IEEEPARstart{S}{emantic} segmentation is crucial for visual understanding, as semantic information is useful for many robotics tasks, e.g. planning, localisation, and mapping \cite{semantic_planning, boxgraph, semantic_mapping}.
Significant progress has been made on \textit{supervised} semantic segmentation, where accuracy has significantly improved over the years.
However, this has mostly involved test datasets from the \textit{same} underlying data distribution as the training data.
% It is still a significant challenge to train a segmentation network that can (1) retain its accuracy on data with a \emph{distributional shift} from the labelled training data, and (2) report accurate uncertainty estimates in the face of a distributional shift.
Considering data with a \emph{distributional shift} from the labelled training data, it is still a significant challenge to train a segmentation network to (1) retain its accuracy on this data (2) report accurate uncertainty estimates.

\begin{figure}
\centering
\includegraphics[width=0.41\textwidth]{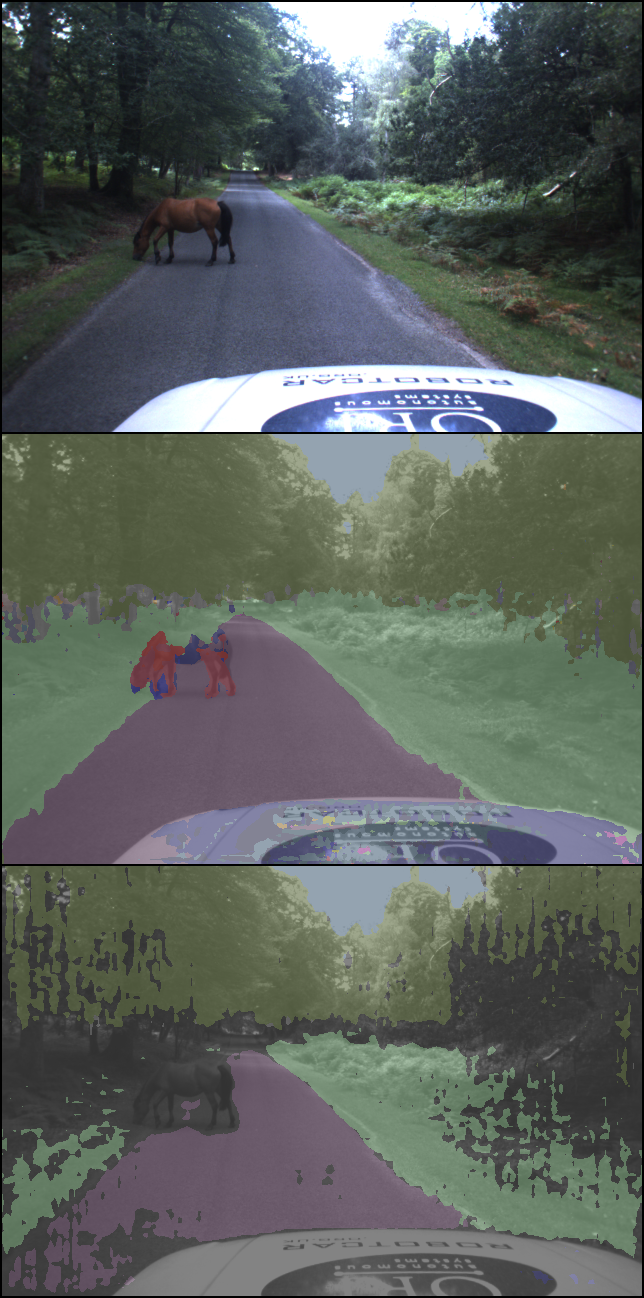}
\caption{
% A qualitative result for a scene from the \textit{SAX Segmentation Test Dataset}.
% Note that at no point were labels in this domain used to train the segmentation network.
In an image from the SAX project~\cite{sax} (top), a horse (an object of unknown class) can be seen on the road.
In the central image, this horse is poorly segmented leading to a dangerous driving situation.
However, the model proposed in this work expresses pixel-wise uncertainty (blacked pixels on the bottom image), thereby mitigating the poor segmentation and the dangerous situation more generally.
Uncertainty is also expressed over unfamiliar greenery that the model struggles to consistently segment as either $\mathrm{vegetation}$ or $\mathrm{terrain}$ (classes defined in Cityscapes).
% \textcolor{red}{Uncertainty around ill-defined borders (e.g. vegetation) between known classes is well documented~\cite{bressan2022semantic}.}
% The model prevents itself from poorly segmenting the horse in the road by segmenting it as unknown (right, seen in 
}
\label{fig:front_page}
\end{figure}
% and (2) usefully report calibrated uncertainty when degrading in the face of a domain shift.

% For a robot traversing an uncontrolled outdoor environment, it is inevitable that it will encounter instances of
% never-before-seen classes, and instances of known classes, but with a different appearance to the labelled examples.
% For \glspl{av}, this more broadly motivates the requirement that a \textit{\gls{odd}} is defined in the safety case for an \gls{av}, i.e. that so far as is possible, the deployment domain is controlled so as to prevent error leading to accidents. 

% For this reason, the safety of \glspl{av} is ensured for an \textit{\gls{odd}}, i.e. an environment with specific characteristics, e.g. location, weather, illumination, sensor setup.
% The controlling of these characteristics is an attempt to control the data distribution input to the \gls{av}; however in dynamic outdoor environments this cannot be guaranteed, e.g. weather changes, dynamic objects of unknown class or appearance appear, illumination varies etc.

% Mobile robots, such as \glspl{av}, are designed to operate safely for an environment with specific conditions, named the \gls{odd}.

The \gls{odd} for a mobile robot is defined as the set of operating conditions under which the robot has been designed to operate safely \cite{pas1881}.
However, due to the dynamic nature of uncontrolled outdoor environments, these operating conditions are liable to change: e.g. weather changes, dynamic objects of unknown classes or appearance are seen, illumination varies, etc.
This is exacerbated by the significant cost of labelling images for semantic segmentation, as it is intractably difficult to anticipate and represent the full breadth of possible situations in the 
sample distribution of the labelled training dataset.
Therefore, it is crucial that robots are able to verify whether they are in their \gls{odd} and can operate safely or whether the domain -- and thus the data distribution -- has deleteriously changed.
Standards for autonomous vehicles \cite{ISO21448, pas1881} cite this as critical for safe deployment.

This work therefore answers the question: given a labelled training dataset in one domain (a.k.a. the \emph{source domain}), how can the segmentation error rate be mitigated on a shifted unlabelled domain (a.k.a. the \emph{target domain})?
In answering the question, a model is presented that can learn to perform high-quality uncertainty estimation from an uncurated unlabelled dataset of the target domain, without the prohibitive cost of labelling. 
An example of the system working is shown in \Cref{fig:front_page}.
Here, the vast majority of the image pixels are segmented correctly but the unknown class ``horse'' is segmented poorly.
This is dangerous: e.g. if identified as some other \textit{static} class, the robot may drive forward, or if identified as some other \textit{dynamic} class, downstream prediction or tracking systems may be affected.
Both situations mean the robot would act unsafely around a wild animal.
Our model however accompanies this prediction with high uncertainty.
Critically, this is learned from unlabelled examples in this domain.

This is accomplished by training a segmentation network using a semi-supervised task, where -- in lieu of labels -- segmentation consistency in the target domain is selectively maximised across data augmentation.
The intuition is that the performance on the semi-supervised task can be considered a proxy for segmentation accuracy, which is then used to train the network to express uncertainty on regions of images with poor performance.
The proposed network expresses uncertainty in feature space with a single forward pass, thus satisfying run-time requirements for a robotics deployment.

The contributions of this work are as follows:
\begin{itemize}
    \item It proposes a training method that leverages an unlabelled dataset to learn pixel-wise uncertainty estimation \textit{alongside} segmentation. % via a small labelled dataset;
    \item It evaluates the robustness of the proposed method against several uncertainty estimation and \gls{ood} detection techniques.
    \item
    It presents a new semantic segmentation benchmark based on images belonging to the Sense-Assess-eXplain (SAX) project \cite{sax}.
    This work proposes over \SI{700}{} pixel-wise labels for a manually curated set of images spanning three domains, that are used for testing both semantic segmentation and uncertainty estimation. \footnote{The labels can be found at \url{https://ori.ox.ac.uk/publications/datasets/}.}
    In addition to the labelled test data, we also propose a set of metrics to evaluate the quality of a model's uncertainty estimates.
    % \item It discusses the appropriate evaluation metrics for uncertainty estimation given the robotics context of this work.
\end{itemize}

% To investigate the use of real-world unlabelled data, this work proposes a novel benchmark based on the SAX Semantic Segmentation Dataset.
% The images for this dataset belong to the Sense-Assess-eXplain (SAX) project \cite{sax}, and this work proposes a curated set of images, with corresponding pixel-wise labels, that are used for testing both semantic segmentation and uncertainty estimation.
% In addition to the labelled test data, we also propose a set of metrics to evaluate the quality of a model's uncertainty estimates.
% This benchmark introduces three geographic domains and, for each domain, contains over \SI{100000}{} unlabelled \emph{training} images and over \SI{200}{} pixel-wise annotated \emph{test} images.

%------------------------------------------------------------------

%------------------------------------------------------------------
\section{Preliminaries on Uncertainty Estimation}
%\subsection{Epistemic vs. Aleatoric Uncertainty Estimation} 
\label{sec:which_uncertainty} 
A given model's error on test data is often described as originating from two different sources: \textit{epistemic} or \textit{aleatoric} uncertainty.
The distinction is that error due to epistemic uncertainty is \emph{reducible}, meaning the modeller can reduce the model's test error e.g. by labelling more data or improving network architecture.
In contrast, aleatoric uncertainty is \emph{irreducible} as the uncertainty is inherent in the test data.
This means it is impossible to train a model that fully reduces the error on this test data, as aleatoric uncertainty is not under the control of the modeller.

Error due to distributional shift, i.e. distributional uncertainty, has two components: (1) uncertainty due to unknown classes, i.e. those not defined in the training data (2) uncertainty due to known classes with unfamiliar appearance.

Given the constraints of this work, it can be argued that (1) is aleatoric as no model parameterisation will fully mitigate the error in the target domain.
This is because our model estimates the probability of a pixel belonging to a \emph{fixed} number of defined classes, meaning that class assignment to a novel class is impossible.
However, it can be estimated directly from data whether a pixel \textit{does not} belong to a defined class.
% whether a pixel belongs to an unknown class.

% Component (2) is caused by the visual dissimilarity between source and target images.
Component (2) is caused by the intra-class visual dissimilarity between source and target images.
This relates to epistemic uncertainty as it can be reduced by a more diverse labelled training dataset.
It can however also be estimated directly from the input target image.
% However as this uncertainty is caused by a clear visual difference
% In spite of this being epistemic uncertainty, this means there is scope to estimate error due to (2) directly from an input image.
% which is indeed epistemic.
% As this arises from the visual difference between the two domains, not just in the sub-optimality of the model weights, there is also scope to estimate error due to (2) directly from the input image.

It is therefore true that the entirety of distributional uncertainty, i.e. components (1) and (2), can be estimated directly from the data.
This motivates this work to draw upon methods for aleatoric uncertainty estimation, as discussed further in \cref{subsec:epistemic_ue} and \cref{subsec:aleatoric_ue}.

%------------------------------------------------------------------
\section{Related Work} \label{sec:related}

\subsection{Epistemic Uncertainty Estimation}  \label{subsec:epistemic_ue}
Epistemic uncertainty estimation considers the weight posterior distribution $p(\mathrm{w|D})$, with $\mathrm{w}$ and $\mathrm{D}$ defining the model weights and training data respectively.
This distribution over possible model parameterizations given the training dataset is then related to the distribution over possible segmentations for an image.
% This allows us to understand at inference time the likelihood of a given pixel being incorrect by looking at the entropy of the latter distribution. 
However, this Bayesian analysis is intractable to perform exactly, and so approximations are made, such as \gls{mcd} \cite{dropout}.
Alternatively, $p(\mathrm{w|D})$ can be defined using an ensemble of models \cite{ensembles}, each trained independently but with the same labelled dataset.

For training and inference, these methods estimate uncertainty by perturbing the model weights to induce a distribution of segmentations for a given image.
% As discussed in \cref{sec:which_uncertainty}, we treat distributional uncertainty as inherent to the data, thereby motivating the hypothesis that it is better estimated by the model's invariance to the data augmentation rather than model perturbation.
For this reason, they are computationally expensive at inference time, as they require multiple forward passes of a network to produce an uncertainty estimate.
When considering the deployment of a segmentation network, both latency and memory usage are critical criteria.

This issue is mitigated in \cite{mcd_distillation} by distilling a \gls{mcd} model into a deterministic network that can estimate uncertainty in a single pass.
% Our proposed model also uses a single pass, making it suitable for many deployment situations.
During the training of our proposed model, a segmentation distribution is instead obtained by applying perturbations to the input data while keeping the model weights constant; at inference time, our model produces segmentation uncertainties in a single pass.

\subsection{Deterministic Uncertainty Methods}  \label{subsec:dum}
Noting the computational requirements of many epistemic uncertainty estimation techniques, \glspl{dum} design networks to estimate uncertainty in a single pass using spectral-norm layers~\cite{spectral_norm}, which constrain the network's Lipschitz constant, ensuring that semantic differences in the input produce proportionally-scaled differences in feature space.
Uncertainty can then be estimated as distance in feature space, i.e. the semantic dissimilarity, between a given input and the labelled training data.
\glspl{dum} differ in how they measure uncertainty in feature space, either with Gaussian Processes \cite{sngp}, a post-hoc Gaussian Mixture Model \cite{dum_baseline}, or \gls{rbf} kernels \cite{duq}.

\glspl{dum}, like this work, turn uncertainty estimation into a representation learning problem, rather than an study of the model weights.
Both methods measure uncertainty in feature space, but \glspl{dum} regularise the feature space with layer normalisation, whereas this work leverages unlabelled data.

\subsection{Aleatoric Uncertainty Estimation}    \label{subsec:aleatoric_ue}
As aleatoric uncertainty is inherent to the data (rather than the model), aleatoric techniques are designed to estimate uncertainty purely as a function of the input data.
This is typically achieved by supervised training.
For the training images, the variability in network output with respect to the ground-truth is approximated by a distribution, \cite{what_uncertainties, probably_unknown}.

% for input images with respect to the ground-truth is approximated by a distribution \cite{what_uncertainties, probably_unknown}.

% The objective function is changed according to the distribution of choice, such that distribution closely approximates the network output. 

For a regression task in \cite{what_uncertainties}, with network estimate $f(x^{(i)})$ and ground-truth $y^{(i)}$, the following loss for a pixel $i$ is used to distribute the output as Gaussian:
\begin{equation*}
    L_{NLL}^{(i)} = \frac{1}{2\sigma(x^{(i)})^2} \|y^{(i)} - f(x^{(i)})\|^2 +\frac{1}{2} \mathrm{log}[\sigma(x^{(i)})]^2
\end{equation*}

% While the network is minimising the squared error between its estimate, $f(x_i)$, and a ground-truth value, $y_i$, it is also estimating the likely error on its estimate, i.e. the variance, $\sigma(x_i)^2$.
Intuitively, this loss function gives the network two paths to minimise the loss.
Its estimate can either be closer to the ground-truth or it can mitigate the large squared error $\|y^{(i)} - f(x^{(i)})\|^2$ by expressing a large variance $\sigma(x^{(i)})^2$.
In \cite{what_uncertainties}, this style of objective is given the name \textit{learned-loss attenuation}.

Furthermore, \cite{light_prob_nets} approximates every layer output as a distribution, using assumed density filtering.
\cite{aleatoric_geo_stable} learns features for geometric matching in a self-supervised manner, while expressing uncertainty over poor geometric matches.

% and also uncertainty estimates associated with the match. 

% This work draws inspiration from the idea of learned loss attenuation; yet, uncertainty is expressed as distance in feature space, rather than as a network output.
This work and \cite{aleatoric_geo_stable}, unlike many techniques, do not require labels to calculate the learned loss-attenuation objective.
%, i.e. they do not use a fully supervised task.
This work differs from \cite{aleatoric_geo_stable}, as it is designed to extract semantics rather than just geometry, meaning labels are required to define the semantic classes.
This work also expresses uncertainty as a feature-space distance rather than as a network output.

% \begin{figure*}[t]
% \centering
% \includegraphics[width=\textwidth]{Figures/training_diagram.png}
% \caption{
% The model parameters are updated by four losses: (a) $L_c$ (b) $L_{uniform}$ (c) $L_{proto}$ (d) $L_{sup}$.
% (a) For the pixels deemed certain by $M_\gamma$, $L_c$ maximises the consistency of segmentations $s'_{T}$, $s_{T}$ of augmentations of the original target image  $x_{T}$.
% (b) $L_{uniform}$ softly constrains the features $z_{T}$ to be uniformly distributed on the unit-hypersphere.
% (c) $L_{proto}$ maximises the distance between source prototypes $p_S$, i.e. the mean embeddings of each class in the source dataset.
% (d) $L_{sup}$ maximises the segmentation accuracy on the source images $x_S$ with respect to ground-truth labels.
% For each diagram, only the networks coloured in \textcolor{Aquamarine}{\texttt{aquamarine}} are updated by the losses.
% \label{fig:system_detail}
% }
% \end{figure*}

\begin{figure*}
\centering
\includegraphics[width=0.8\textwidth]{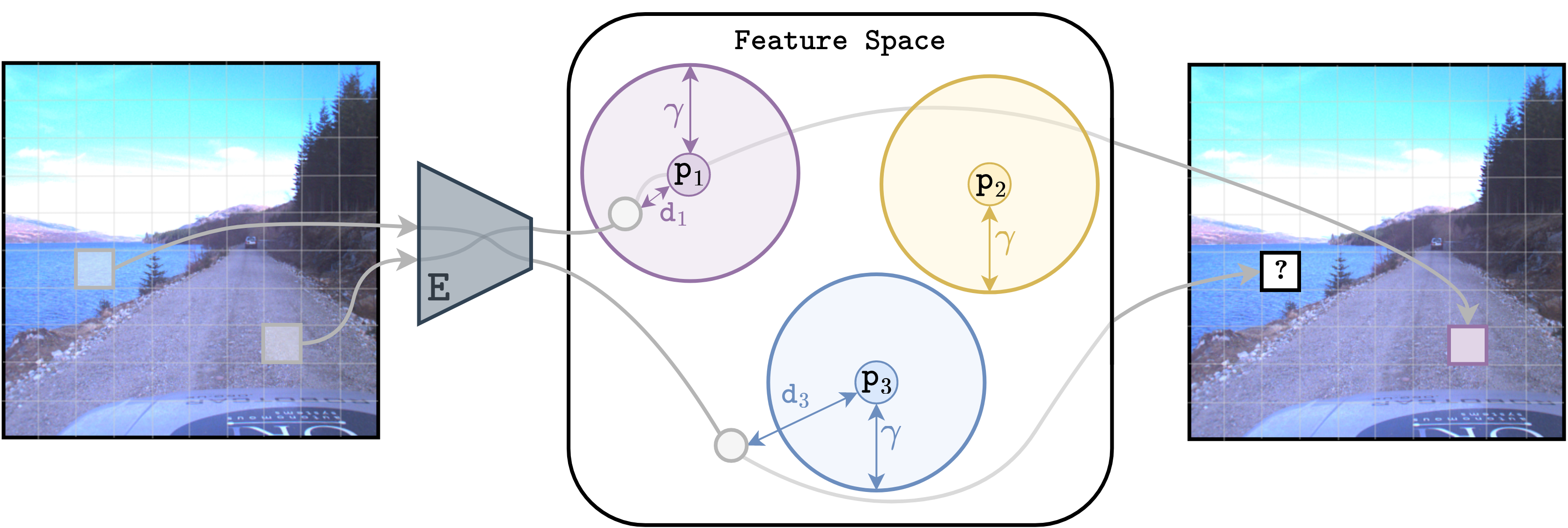}
\caption{
Depiction of simultaneous segmentation and uncertainty estimation for the model presented in this work.
Pixel-wise features are extracted from an image by encoder \texttt{E}.
Distances $\texttt{d}_{1:3}$ are calculated between each feature and prototypical features from each class $\texttt{p}_{1:3}$, known as prototypes.
If one of $[\texttt{d}_1, \texttt{d}_2, \texttt{d}_3] < \gamma$, the feature is $\mathrm{certain}$ and assigned the class of its closest prototype (denoted by the coloured pixel overlaid on the right), and if not, the feature is assigned $\mathrm{uncertain}$ (denoted as the question mark in white pixel). 
In this way, a `safe region of operation' is defined in feature space, where inside pixels are accurate and certain, and outside they are uncertain and inaccurate.
% This feature representation is learned using an uncurated unlabelled dataset.
}
\label{fig:system_overview}
\end{figure*}

\subsection{Out-of-Distribution Detection} \label{ood_lit}
\gls{ood} detection attempts to identify instances that appear distributionally distinct from the labelled training data.
The difference with uncertainty estimation is that the focus is on the data, rather than on mitigating model error.
% , i.e. it explicitly estimates distributional uncertainty.
One set of techniques train a network in a supervised manner on source data and then calculates an \gls{ood} score from the network's learned representation. 
% that describes how dissimilar the input is from the training data.
% The methods differ in how this score is calculated, with the simplest being to take the maximum softmax-score \cite{maxsoftmax}.
The simplest method calculates the maximum softmax-score \cite{maxsoftmax}.
% This is extended by using adversarial perturbations to the input in \cite{odin}.
\cite{odin} also adds adversarial perturbations to the input.
In \cite{mahalanobis}, the Mahalanobis distance between the input and training data in a series of feature spaces is combined with logistic regression. 
In \cite{vim} a score which is a function of both the features and the logits is used. 

Other methods introduce proxy-training tasks to generate an \gls{ood} score.
% In \cite{geometric_ood} a classifier to select which transformation has been performed on the input image is learned, where the max-softmax score for the classification is used as an \gls{ood} score.
In \cite{geometric_ood} a classifier learns to classify the transformation applied to the input image, using the max-softmax score as an OoD score.
% In \cite{self_supervised_ood} the auxiliary self-supervised task of orientation prediction is used, with the result being improved \gls{ood} performance.
In \cite{self_supervised_ood} the task of orientation prediction is utilized, resulting in improved \gls{ood} performance.

Our work also introduces a task to learn an \gls{ood} score, but uses the task to learn a separable representation of in-distribution and \gls{ood} pixels, so the task does not have to be run at inference time.
Additionally, the aforementioned works train only on in-distribution data, instead of  improving robustness via leveraging \gls{ood} data.
Finally, the proposed model performs pixel-wise, rather than image-wise, \gls{ood} detection.

% operate on pixel-wise segmentation rather than image-wise classification.

% Density estimation is also used for \gls{ood} detection, whereby deep generative models learn a structured representation of the in-distribution data.
% Examples include using variational auto-encoders, \cite{vaerecon2015} and energy-based models \cite{classifierEBM}.
% The performance of these types of approaches for \gls{ood} has been mixed, with works such as \cite{bad_generative} showing that the feature density producing a poor performance.
% These works use only in-distribution data, and so struggle to produce a separable representation of \gls{ood} data.

Both \cite{outlierexposure, prior_nets} use a curated \gls{ood} dataset to explicitly separate in-distribution and \gls{ood} data.
\cite{ood_gan, fool_me_once} generate \gls{ood} datasets containing \textit{near-distribution} images, i.e. those at the edge of the training distribution, with a \gls{gan} and data augmentation respectively. 
Near-distribution images are used to achieve more robust \gls{ood} detection than clearly \gls{ood} images, as separating in-distribution and near-distribution images is much more challenging.

% \cite{ood_gan} generates an \gls{ood} dataset with a \gls{gan} optimised to produce images at the fringes of the labelled in-distribution dataset, suggesting that boundary \gls{ood} examples are more informative. 
% \cite{fool_me_once} uses an \gls{ood} dataset; however, it uses data augmentation to reduce the difference between \gls{ood} and in-distribution data and a contrastive loss to learn a separable representation.

Although unlabelled, a purely \gls{ood} dataset can require costly curation.
They are also less informative for pixel-wise \gls{ood} detection, where in testing, \gls{ood} instances are contained in in-distribution scenes (and vice versa).
In contrast, our work's target-domain dataset does not require curation and contains near-distribution \textit{and} in-distribution \emph{and} \gls{ood} instances within the same image.
For these reasons, this data represents the opportunity to learn a more robustly separable representation, at the cost of being more challenging to use.

\subsection{Semi-Supervised Learning} \label{subsec:semi-supervised}
Semi-supervised methods extend supervised methods by including a loss on unlabelled data, and are evaluated on their ability to reduce the test error rate.
In contrast, our approach seeks to \textit{detect} test errors.
These approaches operate orthogonally, but both prevent unknowingly erroneous predictions.
% \dw{talk about this ^^ here on in intro???}

Semi-supervised approaches often maximise the consistency of a model's representation across perturbations to the input, model, or both \cite{meanteacher, temporalensembling, consistency_training}.
Similar to this work, \cite{paws} uses data augmentation and a cross-entropy objective between the class assignment distributions.
In contrast, we apply the objective selectively and per-pixel.
To appropriately represent \gls{ood} data, we also introduce additional objectives and procedures in place of the regularisation objective seen in \cite{paws}.

% This work uses a similar framework to \cite{paws}, where two views on unlabelled images are generated with data augmentation.
% Similarly, a representation of the target data is learned by maximising the consistency in feature embedding, but, differently, only on pixels estimated to be in distribution.

\subsubsection*{Prototypes}
Related to \cite{paws}, \cite{swav} maximises consistency using prototypes as part of its mechanism, also seen in few-shot learning literature \cite{prototypicalnets, matching_nets}.
A prototype is calculated for each class as the centroid of all source embeddings for that class.
Prototypes thus compactly represent the high-level semantics for a given class by averaging over the intra-class factors of variation.
Additionally, \glspl{dum} calculate class mean features (i.e. prototypes) and measure uncertainty as a distance between features and class centres, as mentioned in \cref{subsec:dum}.

\subsubsection*{Unsupervised Domain Adaptation}
A specific instance of semi-supervised learning is \gls{uda}, which specifically targets the distributional shift between the labelled (a.k.a. \emph{source}) and unlabelled (a.k.a. \emph{target}) data.
Still, \gls{uda} methods \cite{kang2019contrastive,tzeng2017adversarial} are designed purely to increase test accuracy by learning an improved representation of the target domain and not to detect the errors arising from the shift, as in our method.
% \gls{uda} methods~\cite{kang2019contrastive,tzeng2017adversarial} use the unlabelled target data to learn an improved representation of the target domain.
% This work also seeks to learn an improved representation of the target domain; however, we focus on the separability of in-distribution and \gls{ood} instances, rather than solely increasing accuracy on the segmentation task.

% One class of method achieves this by aligning the feature distributions of source and target data.
% This can be done with a hand-crafted distance function, such as maximum mean discrepancy \cite{mmd_domain_adaptation}, or by learning a distance function in an adversarial setting, \cite{advent}.

% An alternative to this is to align the source data to the target data in image space.
% This is useful as then the source labels can be used to train the network on images with the appearance of the target domain, thus improving accuracy in the target domain.
% These techniques often use a CycleGAN \cite{cyclegan} as the means to transfer appearance from one domain to another \cite{cycada}.

%------------------------------------------------------------------
%------------------------------------------------------------------

\section{System Overview}
\label{sec:system_ov}

\Cref{fig:system_overview} shows an overview of the proposed system.
At inference, its goal is to segment each pixel of an image of size $H\times W$, $x \in \mathbb{R}^{3 \times W \times H}$ into $K$ known classes $\mathcal{K} = \set{k_1, \dots k_K}$ or flag them as uncertain.
This is done by producing both a categorical distribution $p(y | x) = [p(y=k_1|x), \dots p(y=k_K|x)] \in \mathbb{R}^{K \times W \times H}$ and an uncertainty mask $M_\gamma \in \mathbb{B}^{W \times H}$, which assigns to each pixel $1$ for \certain or $0$ for \uncertain through a threshold $\gamma$ in feature space.

\subsection{Segmentation Using Prototypes} \label{subsec:proto_seg}
An encoder, $\texttt{E}: \mathbb{R}^{3 \times H \times W} \rightarrow \mathbb{R}^{F \times h \times w}$, and projection network $\texttt{g}_{\rho}: \mathbb{R}^{F \times h \times w} \rightarrow \mathbb{R}^{F \times h \times w}$ calculate embeddings $z \in \mathbb{R}^{F \times h \times w}, \| z \|_2 = 1$ of an image $x$, where $F$ is the feature length and $h$ and $w$ are the downsampled spatial dimensions.%, as per the bottom branch in \cref{fig:system_detail}.
% batch of $N$ images are calculated by an encoder network $\texttt{E}$ as follows, $E: \R^{N \times 3 \times H \times W} \rightarrow \R^{N \times M \times h \times w}$, and, to calculate cosine similarity, are normalised: $\|z^{(i)}\|_2=1$, where $z^{(i)} \in z$.

Let $X_S \in \mathbb{R}^{N \times 3 \times W \times H}$ be a batch of $N$ source images and $Y_S \in \R^{N \times h \times w \times K}$ their corresponding one-hot labels downsampled to the size of the embeddings $Z_S \in \R^{N \times h \times w \times F}$.
Prototypes $p_S \in \R^{F \times K}$ are then calculated for each of the $K$ classes from the embeddings $Z_S$:
\begin{equation}\label{eqn:prototypes}
    p_S = \frac{Z_S^{\top} Y_S}{\| Z_S^{\top} Y_S \|_2}
\end{equation}
% including normalisation such that: $\| p_S^{(k)} \|_2=1$
% \begin{equation*}
%     \boldsymbol{p_S^{(k)}} = \frac{1}{m} \sum_{k \in K} \boldsymbol{z_S^{(k)}}
% \end{equation*}
% where the normalising factor is $m=\| \sum_{k \in K} \boldsymbol{z_S^{(k)}} \|_2$.
Here, $Z_S^{\top} Y_S\in\R^{F \times w \times h \times N}\times\R^{N \times h \times w \times K}=\R^{F \times K}$, giving us a normalised, aggregate feature per class.
%$k\in\mathcal{K}$.
%, which is then also normalised.
A segmentation $p(y|x)$ is obtained by projecting each pixel embedding of $x$ onto the prototypes.
Specifically, the classification scores, $\tilde{s}^{(i)} \in \mathbb{R}^{1 \times K}$, for the $i$-th pixel embedding $z^{(i)} \in \R^{F \times 1}$ are:
\begin{equation}
    \tilde{s}^{(i)} = z^{(i)\top} p_{S}
\end{equation}

% Here, $z^{(i)\top} p_{S} \in \mathbb{R}^{K\times{}F}\times\mathbb{R}^{F\times{}1}=\mathbb{R}^{K\times{}1}$.
Here, $z^{(i)\top} p_{S} \in \mathbb{R}^{1\times{}F} \times \mathbb{R}^{F\times{}K}=\mathbb{R}^{1\times{}K}$.
For this segmentation method, the scores $\tilde{s}$ represent the cosine similarity between a feature and the prototype for each class.
These scores over the downsampled spatial resolution $\tilde{s}\in\mathbb{R}^{h\times{}w{}\times{}K}$ are then bilinearly upsampled to $s \in \mathbb{R}^{H \times W \times K}$.
The categorical distribution over the $K$ classes, $p(y|x)^{(i)} \in \R^K$, is given by
\begin{equation}\label{eq:segProto}
p(y|x)^{(i)} = \sigma_{\tau}(s^{(i)})
\end{equation}
where $\sigma_\tau(.)$ is the softmax function with temperature $\tau$ (in this work, $\tau=0.07$).

\subsection{Uncertainty Estimation using $\gamma$}  \label{subsec:gamma}
% As the prototypes represent the mean embedding for each class for the source images, the distance between the source prototypes and the target data should encode a meaningful difference in the semantics between source and target.
% Therefore this distance is taken to estimate the uncertainty.
% The other option would be for the network to directly express uncertainty values as an output tensor, as done in aleatoric uncertainty estimation.

% The problem with this lies in the fact that ground-truth does not exist for the target data.
% This means that the network is prone to ``cheating'' at the objective and either expressing low uncertainty for everything, or high uncertainty for everything.
% Calculating the uncertainty as a latent space characteristic is therefore preferable as it makes it harder to cheat at the objective. 
% \dw{Can explain this in more detail.}

% For distance in a latent space to encode meaningful differences in the input, regularisation is needed.
% This is the motivation behind the cited works in \glspl{dum}, see \cref{subsec:dum}.
% This work instead provides regularisation using unlabelled target data, rather than spectral normalisation layers.

Uncertainty is expressed in this work as the probability that a pixel belongs to \textit{no known class}, i.e. $p(y \notin \mathcal{K} | x)$.
% In order to express uncertainty as a feature space distance, $p(y \notin \mathcal{K} | x)$ is defined by the parameter $\gamma$, by appending it to the classification scores as the $(K+1)$-th score:
Taking inspiration from \cite{gamma_uber,gamma_wayve}, we append a parameter $\gamma$ to the classification scores as the $(K+1)$-th score:

% This is defined by the parameter $\gamma$ which is appended to the classification scores as the $(K+1)$-th score:
\begin{equation}
    % s^{(i)} = \sigma_{\tau}([z^{(i)} \boldsymbol{p_{L}}^{\top}, \gamma])
    % p(y^{(i)}|x) = \sigma_{\tau}([z^{(i)} \boldsymbol{p_{L}}^{\top}, \gamma])
    p(y|x)^{(i)} = \sigma_{\tau}(s^{(i)} \oplus \gamma) \in \R^{K+1}
\end{equation}
\fixM{Where $\oplus$ denotes vector concatenation.}{2.1\label{comm:2.1_1}}
% Now, $p(y|x) \in \R^{K+1}$ can be seen as concatenation of two terms: $p(y\in \mathcal{K} |x) \in \mathbb{R}^K$ and $p(y \notin \mathcal{K}|x)] \in \R$.

The largest score, $\mathrm{max}(s^{(i)})$, is the cosine similarity between an embedded pixel and its closest prototype.
This is a measure of the model's confidence, upon which $\gamma$ operates as a threshold
\footnote{
%Let $s_{1:K{+}1}= [s_1, s_2, ...\ s_K, \gamma]$.
For example, if $s^{(i)}_{k=2}$ is larger than $\gamma$, after a low temperature softmax, ${p(y=2|x)^{(i)}}$ will be high.
However, if none of $s^{(i)}$ surpass $\gamma$, then $p(y=K{+}1|x) \equiv p(y \notin \mathcal{K}|x)$ will be high, (see \cref{fig:system_overview} for a depiction).}.
Certainty mask, $M_\gamma$, is given by:
\begin{equation}\label{eq:certainMask}
    % M_\gamma^{(i)} = \mathbbm{1}[s_{1:K}^{(i)}>\gamma]
    M_\gamma^{(i)} = 1 - \mathbbm{1}[\mathrm{argmax}(s^{(i)} \oplus \gamma) = K{+}1]
\end{equation}
where $\mathbbm{1}[\cdot]$ is the indicator function, and so certainty is given by $M_{\gamma}^{(i)}=1$ if the highest score is a known class, otherwise $M_{\gamma}^{(i)}=0$ if $\gamma$ is the highest score.
%, indicating uncertainty.
This way, $\gamma$ defines a region around each prototype, outside of which the feature is considered uncertain.

% \subsection{Regularisation with unlabelled data}
% As the prototypes represent the mean embedding for each class for the source domain images, the distance between the source prototypes and the unlabelled target data should encode a meaningful difference in the semantics between source and target.
% Therefore this distance is taken to estimate the uncertainty.
% The other option would be for the network to directly express uncertainty values as an output tensor, as done in aleatoric uncertainty estimation~\cite{aleatoric_geo_stable, what_uncertainties}.
% The problem with this lies in the fact that ground-truth does not exist for the target data.
% This means that the network is prone to ``cheating'' at the objective -- either expressing low uncertainty for everything, or high uncertainty for everything.
% Calculating the uncertainty as a latent space characteristic is therefore preferable as it makes it harder to cheat at the objective. 
% % \dw{Can explain this in more detail.}

% However, for distance in a latent space to encode meaningful differences in the input, regularisation is needed.
% This is the motivation behind the cited works in \glspl{dum}, see \cref{subsec:dum}.
% This work instead provides regularisation using unlabelled target data, rather than spectral normalisation layers.

% \cref{sec:training} thus further describes the theoretical and practical aspects of getting this system to train in a well-behaved way.

% \section{Proposed Training Regime}
\section{Training objectives}
\label{sec:training}

\begin{figure*}[h]
\centering
\includegraphics[width=\textwidth]{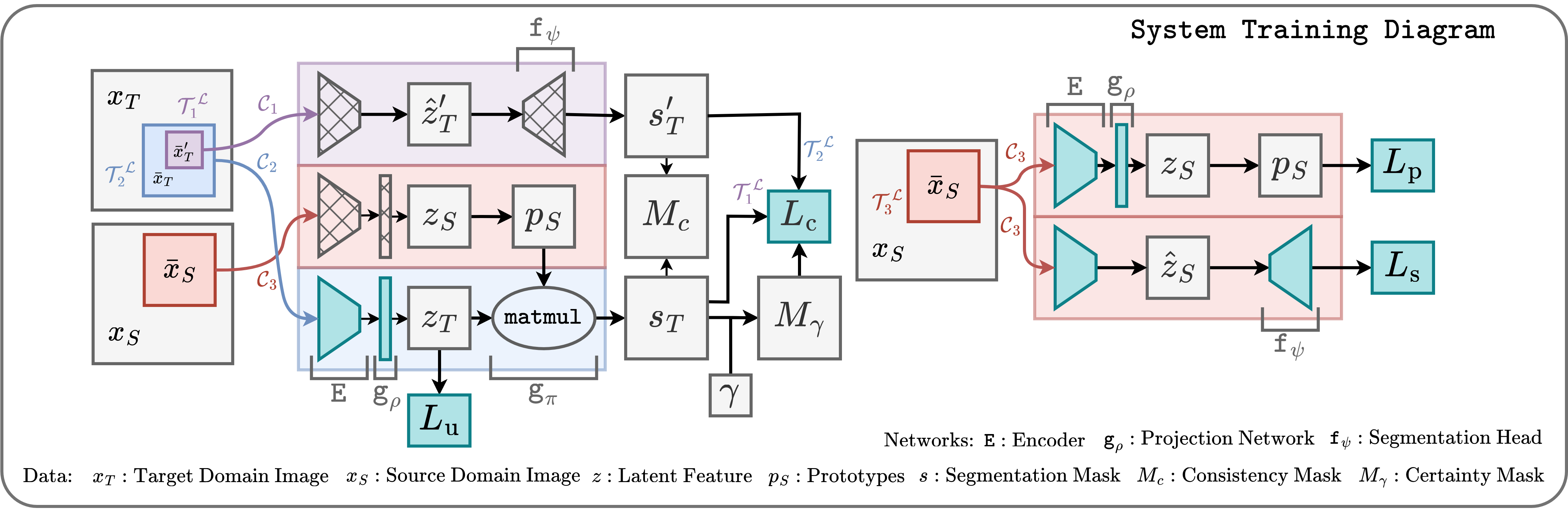}
\caption{
The training regime of the proposed approach.
The model parameters are updated by four losses: (a) $L_{\text{c}}$ (b) $L_{\text{u}}$ (c) $L_{\text{p}}$ (d) $L_{\text{s}}$.
(a) For the pixels deemed certain by $M_\gamma$, $L_{\text{c}}$ maximises the consistency -- a proxy for accuracy -- over the segmentations $s'_{T}$, $s_{T}$ of augmented versions $\bar{x}'_T,\bar{x}_T$ of the original target domain image $x_{T}$.
(b) $L_{\text{u}}$ softly constrains the features $z_{T}$ to be uniformly distributed on the unit-hypersphere.
(c) $L_{\text{p}}$ maximises the distance between source prototypes $p_S$, i.e. spreads the mean embeddings of each class in the source domain dataset over the unit-sphere uniformly.
(d) $L_{\text{s}}$ maximises the accuracy for the segmentations of the source images $x_S$ with respect to ground-truth labels.
% For each diagram, only the networks coloured in \textcolor{Aquamarine}{\texttt{aquamarine}} are updated by the losses\textcolor{blue}{, while the cross-hatched networks are not}.
For each diagram, the networks coloured in \textcolor{Aquamarine}{\texttt{aquamarine}} are updated by the losses, while the cross-hatched networks are not.
% Note that as described in \cref{subsec:ssl_task},  $\bar{x}'_T,\bar{x}_T$ are defined as $\bar{x}'_T = \pazocal{C}_1 \circ \pazocal{T}^{\pazocal{L}}_{1} \circ \pazocal{T^{G}}(x_{T})$ and $\bar{x}_T = \pazocal{C}_2 \circ \pazocal{T}^{\pazocal{L}}_{2} \circ \pazocal{T^{G}}(x_{T})$, 
Note that for diagrammatic clarity, the colour transforms are depicted as following $\bar{x}'_T,\bar{x}_T, \bar{x}_S$, whereas in reality and as described in \cref{subsec:ssl_task}: $\bar{x}'_T = \pazocal{C}_1 \circ \pazocal{T}^{\pazocal{L}}_{1} \circ \pazocal{T^{G}}(x_{T})$, $\bar{x}_T = \pazocal{C}_2 \circ \pazocal{T}^{\pazocal{L}}_{2} \circ \pazocal{T}_1^{\pazocal{G}}(x_{T})$, $\bar{x}_S = \pazocal{C}_3 \circ \pazocal{T}^{\pazocal{L}}_{3} \circ \pazocal{T}_2^{\pazocal{G}}(x_{S})$.
Best viewed in color.
\label{fig:system_detail}
}
\end{figure*}

Ultimately, we aim to train a model that can express uncertainty with uncertainty map $M_{\gamma}$ via the threshold $\gamma$.
Yet, in the absence of labels, we draw on the hypothesis that
% To train such a system, a method is presented in \cref{fig:system_detail} that imposes the hypothesis that accurate pixels must be consistent across perturbation and inaccurate pixels are inconsistent.
% This method draws on the following hypothesis to train such a system: in the absence of labels, 
segmentation consistency over image augmentation approximates ground-truth accuracy.
Consider two corresponding pixels from two augmentations: if both are assigned the same class, this is likely to be accurate; otherwise, it is not. 
Embedding consistency, whether feature distance \cite{simclr, moco} or implicit class assignment \cite{swav}, is a good proxy for classification accuracy, shown by the success of linear probe experiments in training classifiers from models trained with an embedding consistency objective.
% However, such methods do not directly apply to our problem, where we instead focus on learning a representation for uncertainty estimation. 
Instead, we focus on learning a representation for uncertainty estimation, and use pixel-wise alignment, rather than image-wise supervision.
% Additionally, this work uses pixel-wise rather than image-wise supervision, requiring pixel-wise alignment to compute loss terms.

Seen in \cref{fig:system_detail}, consistency between two augmented target-domain images $\bar{x}'_T$ and $\bar{x}_T$, is represented by consistency map $M_c \in \mathbb{B}^{H \times W}$, where each pixel is $1$ if consistent or $0$ otherwise.

% This work relates the consistency and uncertainty maps in the following two-step process: 
In this work, consistency maps are related to uncertainty maps in the following two-step training process:
% in feature space, where the threshold $\gamma$ is used to generate the uncertainty map $M_{\gamma}$.
% To maximise supervision for this task, we make sure there is a good deal of pixel overlap between $\bar{x}'_T$ and $\bar{x}_T$, see~\cref{subsec:ssl_task} below.
% The uncertainty associated with a segmented image is represented in two ways in the proposed system: (a) $M_c \in \{0,1\}$, returning $1$ iff the segmentations $s_{T1, T2}$ are consistent across augmented versions $x_{T1, T2}$ of the input image
% (b) $M_{\gamma} \in \{0,1\}$, returning $1$ iff the pixel-wise features $z_{T}$ are within a given distance threshold $\gamma$ around the source prototypes $p_S$ (see \cref{fig:front_page}).
% The system training aims to improve these uncertainty estimates by using unlabelled target domain images.
% Consider a two-step process for the uncertainty and consistency masks:
\begin{enumerate}%[(a)]

% \item To improve the estimate of this consistency map, $M_c$, the consistency needs to be maximised for accurate pixels and minimised for inaccurate pixels.
% This is achieved by using the uncertainty mask $M_\gamma$ to approximately determine which pixels are accurate and inaccurate.
% % (in conjunction with objectives shown in \cref{fig:system_detail} and described in~\cref{add_objectives}).

% \item In the other direction, to improve $M_\gamma$, the threshold $\gamma$ needs to be refined.
% $M_c$ is used for this, where we solve for $\gamma$ such that $M_\gamma$ approximates $M_c$ -- which is to say that certain pixels should be consistent, and uncertain pixels should be inconsistent.

\item 
$\gamma$ is solved for such that the mean certainty of segmentations according to $M_\gamma$ is equal to the mean consistency according to $M_c$. 
$\gamma$ therefore separates pixels into \certain and \uncertain broadly according to consistency.

\item 
% The model parameters are then updated such that the estimate of this consistency map $M_c$ is improved.

% The model parameters are then updated such that the accurate and inaccurate pixels are separated in the latent space, which in turn improves the $M_c$ as an estimate of ground-truth accuracy. 

The model parameters are then updated by maximising the consistency of pixels deemed \certain by $M_\gamma$, as a proxy for maximising the accuracy of the \certain pixels.
This both improves the estimate of $M_{\mathrm{c}}$ and separates features of pixels assigned \certain and \uncertain.

% This would ideally be achieved by maximising the consistency of accurate pixels, and minimising the consistency for the rest, but instead the uncertainty mask $M_\gamma$ is used to approximately determine which pixels are accurate and inaccurate, i.e. optimising the model so that certain pixels are consistent and uncertain pixels are inconsistent.

\end{enumerate}
This method therefore establishes a \textit{positive feedback loop}, where $M_c$ and $M_\gamma$ continually improve each other's estimates of which pixels are segmented accurately.
The following sections describe how, with care, the training dynamics can be conditioned such that this leads to simultaneous high-quality uncertainty estimation and segmentation.

\subsection{Semi-Supervised Task}\label{subsec:ssl_task}
% Due to the lack of target labels, a self-supervised objective is defined using data augmentation.
% The absence of labels in the target domain makes direct supervision of the class assignment unviable.
% Direct supervision is not possible due to the absence of labels in the target domain. 

Augmentations transform one image into two with distinct appearances but contain the same underlying semantics.
% Hence, the distance between embeddings of these images is a measure of the network's invariance to the appearance of visual semantic concepts, which empirically has been shown to be a good proxy metric for classification accuracy \cite{simclr,moco,swav}.
% This is achieved with the following process: starting with two colour-augmented versions, $\bar{x}'_T$ and $\bar{x}_T$, of initial target domain image $x_{T}$, where $\bar{x}'_T = \pazocal{C}_1 (x_T),\ \bar{x}_T = \pazocal{C}_2 (x_T)$ :
% For a set of colour-transformations $\pazocal{C}_1, \pazocal{C}_2 \sim \pazocal{C}$, two colour-augmented versions, $\bar{x}'_T$ and $\bar{x}_T$, of initial target domain image $x_{T}$ are obtained, $\bar{x}'_T = \pazocal{C}_1 (x_T),\ \bar{x}_T = \pazocal{C}_2 (x_T)$.
% Two colour-augmented versions, $\bar{x}'_T = \pazocal{C}_1 (x_T),\ \bar{x}_T = \pazocal{C}_2 (x_T)$, of the initial unlabelled target image $x_{T}$ are obtained, where $\pazocal{C}_1, \pazocal{C}_2 \sim \pazocal{C}$ are randomly sampled colour-space transformations.
% Pixel-wise aligned segmentations of these two images are then obtained via the following:
Firstly, a target image $x_{T}$ is randomly cropped with transform $\pazocal{T}_1^\pazocal{G}$.
This global crop is transformed by $\pazocal{T}^{\pazocal{L}}_{1}$ and $\pazocal{T}^{\pazocal{L}}_{2}$, which are sampled such that one is always a local crop and resize, and the other an identity transform.
Finally, $\bar{x}'_{T}$ and $\bar{x}_{T}$ are obtained by applying colour-space transforms $\pazocal{C}_1$ and $\pazocal{C}_2$.
At the end, $\bar{x}'_{T}$ and $\bar{x}_{T}$ are images of the same spatial dimensions, but of different appearance and where one is an upsampled crop of a region within the other.

Both of these images are then segmented by functions $\texttt{f}$ and $\texttt{g}$.
Pixel-wise segmentations $s'_{T}$ and $s_{T}$ are obtained by applying the opposite local cropping transform to the one applied to the input image, as follows:
% \begin{equation*}
%     t_1, t_2 \sim T \\
% \end{equation*}
% \begin{equation}\label{eq:sstask_augment}
%     p(y | \bar{x}'_T)^{(i)} = f(T_1(x_1))^{(i)},\ p(y | \bar{x}_T)^{(i)} = f(T_2(x_2))^{(i)}
% \end{equation}
\begin{equation}\label{eq:sstask_augment}
\begin{split}
    s'_{T} = \pazocal{T}^{\pazocal{L}}_{2} \circ \mathlarger{\texttt{f}} \circ \pazocal{C}_1 \circ \pazocal{T}^{\pazocal{L}}_{1} \circ \pazocal{T}_1^\pazocal{G}(x_{T}) \\ s_{T} = \pazocal{T}^{\pazocal{L}}_{1} \circ \mathlarger{\texttt{g}} \circ \pazocal{C}_2 \circ \pazocal{T}^{\pazocal{L}}_{2} \circ \pazocal{T}_1^\pazocal{G}(x_{T})
\end{split}
\end{equation}
% i.e. if $\pazocal{T}^{\pazocal{L}}_{1}$ was applied to the image, then $\pazocal{T}^{\pazocal{L}}_{2}$ is applied to the corresponding segmentation.
$s'_{T}$ and $s_{T}$ are therefore pixel-wise aligned as they are both segmentations of the region of the smaller local crop. 
$\texttt{f}(\cdot)$ and $\texttt{g}(\cdot)$ represent the top and bottom branches in \cref{fig:system_detail} and are distinct functions that both return a segmentation for a given image.
Insights for why $\texttt{f} \neq \texttt{g}$ are given in \cref{subsec:asymm}.

Transformations are applied to $x_S$ in the following order to obtain $\bar{x}_S$: $\pazocal{T}_2^\pazocal{G}, \pazocal{T}^{\pazocal{L}}_{3}, \pazocal{C}_3$, where $\pazocal{T}_2^\pazocal{G}$ is a global crop, $\pazocal{T}^{\pazocal{L}}_{3}$ is the identity transform or a local crop and resize, and $\pazocal{C}_3$ is a colour-space transform.

% Here we have as per~\cref{fig:system_detail} the augmentations $\bar{x}'_T=T_1(x_1)$ and $\bar{x}_T=T_2(x_2)$ where $\pazocal{T}_A, \pazocal{T}_B \sim \pazocal{T}$ are random cropping and rescaling transformations with the first augmentation completely inside the second.
% These are applied to target domain images.
% For $\bar{x}'_T$, however, they are produced by a parallel branch, as motivated in in~\cref{subsec:asymm}.
% with a colour space transform as well as a geometric transform -- a crop and rescale.

%%%%%%%%%%%%%%%%%%%%%%%% Calculating gamma %%%%%%%%%%%%%%%%%%%%%%%%
\subsection{Calculating $\gamma$: Making inconsistent pixels uncertain}     \label{subsec:calculating_gamma}
% The parameter gamma is used 

% In~\cref{eq:Lc}, consistency was represented by the cross-entropy term, i.e. the distributional dissimilarity between $p(y|\bar{x}'_T)$ and $p(y|\bar{x}_T)$.
% The model was updated such that consistency was only maximised for \certain pixels, with uncertainty expressed as $M_{\gamma}$ via $\gamma$.

% In this, consistency approximated accuracy.

% Now, we introduce another inductive bias which ensures in a \textit{converse} fashion that consistent pixels are certain, and inconsistent pixels are uncertain -- in this case by calculating $\gamma$.

Let us consider batches of target domain images, $\bar{X}'_T, \bar{X}_{T} \in \mathbb{R}^{N \times 3 \times H \times W}$, and segmenting them to obtain $S^{'}_{T}, S_{T} \in \mathbb{R}^{N \times K \times H \times W}$.
The consistency mask $M_c \in \mathbb{B}^{N \times H \times W}$, where $M_c^{(i)} = 1$ if consistent, else $M_c^{(i)} = 0$, is given by:
\begin{equation}\label{eqn:Mc}
M_c^{(i)}=\begin{cases}
1 & \underset{k \in \mathcal{K}}{\mathrm{argmax}}({S}_{T}^{'(i)})=\underset{k \in \mathcal{K}}{\mathrm{argmax}}(S_{T}^{(i)})\\
0 & \text{otherwise}
\end{cases}
\end{equation}

$\gamma$ is then calculated such that the $\mathrm{p(certain)}$ (i.e. the proportion of pixels that are certain according to $M_{\gamma}$) is equal to $\mathrm{p(consistent)}$ (i.e. the proportion of pixels consistent according to $M_c$), as detailed in \Cref{alg:gamma_calc}.
Here, $\mathrm{Max}S_{T}$ contains the largest similarity with prototypes for each pixel.
$(1-p_c)$ is the proportion of inconsistent pixels in the batch.
Line 9 then chooses $\gamma$ so that certain pixels have the same proportion as consistent.

% Note that the use of $\gamma$ means that 
% Note that consistency and certainty are not enforced at a pixel-wise level, but at a batch-wise level.

% This completes our feedback between $M_\gamma$ and $M_c$, the remaining sections focused on methods to tame the training dynamics introduced by this cycle and ensuring that the implicit biases expressed in~\cref{eq:Lc} and~\cref{eqn:Lgamma} are well exploited.

\begin{algorithm}
\caption{
Algorithm to calculate $\gamma$.
} \label{alg:gamma_calc}
\begin{algorithmic}[1]
\State{\textbf{Inputs:}}
\State{Consistency mask: $M_c \in \R^{N \times H \times W} $}
\State{Classification scores: $S_{T} \in \R^{N \times K \times H \times W} $}
\Function{calculate\_gamma}{$M_c,S_{T}$}
% \State $N, H, W = \texttt{shape(}M_{c} \texttt{)}$
\State $\mathrm{Max}S_{T} =  \texttt{flatten(}\texttt{max(}S_{T},\ \texttt{dim="K"))}$
\State $\mathrm{Max}S_{T} = \texttt{sort(}\mathrm{Max}S_{T},\ \texttt{ascending=True)}$
\State $p_c = \texttt{mean(}M_{c} \texttt{)}$ \Comment{\% consistent pixels}
\State $\mathrm{R} = (1-p_c)*N*H*W$ \Comment{Num. uncertain pixels}
\State $\gamma = \mathrm{Max}S_{T}\texttt{[int(}\mathrm{R}\texttt{)]}$
\State \textbf{return} $\gamma$
\EndFunction
\end{algorithmic}
\end{algorithm}

\subsection{Learning \large{$\texttt{E}$}\normalsize{: Making certain pixels consistent}}    \label{subsec:learning_E}
% \subsection{Learning $f$: Making uncertain pixels inconsistent}    \label{subsec:learning_E}
Similar to learned loss attenuation, the ultimate objective is to train a model that produces either high-quality segmentations or expresses high uncertainty.
% To this end, the consistency objective, $L_c$, maximises the segmentation consistency across both views, but \emph{only} for pixels that are considered certain by $\gamma$.
% To this end, a consistency objective, $L_\text{c}$, maximises the quality of the segmentation 
To this end, a consistency objective, $L_\text{c}$, maximises the quality of the segmentation, but \emph{only} for pixels that are deemed certain by $M_\gamma$.
Segmentation quality is represented by the consistency, which in this case is calculated as the cross-entropy between the pixel-wise categorical distributions across views, as follows:
\begin{equation}\label{eq:Lc}
    % L_{DA} = - \frac{1}{N\sum_{j}^{N}M_\gamma^{(j)}}\sum_{i}^{N}\sum_{k}^{K} M_\gamma^{(i)}\ p(y_i=k|I_{T1})\ log(p(y_i=k|I_{T}))
    % L_{c} = - \frac{\tfrac{1}{NHW}}{\sum_{j}^{NHW}M_\gamma^{(j)}}\sum_{i}^{NHW} M_\gamma^{(i)}\ H[p(y|\bar{x}'_T)^{(i)}, p(y|\bar{x}_T)^{(i)}]
    L_\text{c} =  \frac{\sum_{i}^{NHW} M_\gamma^{(i)}\ \mathcal{H}[p(y|\bar{x}'_T)^{(i)}, p(y|\bar{x}_T)^{(i)}]}{\sum_{j}^{NHW}M_\gamma^{(j)}}
\end{equation}
where $\mathcal{H}$ is the cross-entropy function.
% Importantly, after segmentation in~\cref{eq:sstask_augment}, $M_\gamma$ is fixed as per~\cref{eq:certainMask}.
% Indeed, in this step, $\gamma$ is fixed and the segmentation network ($f$) parameters are updated.
As shown in  \cref{fig:system_detail}, only the encoder $\texttt{E}$ is updated by this loss function.

% Thus the model is incentivised in minimising~\cref{eq:Lc} to update $\texttt{E}$ to make uncertain pixels become inconsistent.

% with the model embedding a pixel at a distance greater than $\gamma$ from the prototypes
% Oppositely, greater consistency in segmentation is seen as a decrease in $\mathbb{H}[p(y|\bar{x}'_T)^{(i)}, p(y|\bar{x}_T)^{(i)}]$, optimal as   $M_\gamma^{(i)}$ being high indicates certainty.
% with the model embedding a pixel at a distance greater than $\gamma$ from the prototypes -- i.e. making $\mathrm{argmax}(s^{(i)} \oplus \gamma) = K{+}1$ and thus $M_\gamma^{(i)}$ zero in~\cref{eq:certainMask} as the pixel embedding will have low similarity $s^{(i)}$ to any prototype.

$L_{\text{c}}$ causes an entropy decrease for $p(y|x)$ of \certain pixels, but not for \uncertain pixels.
% This means that the entropy of the categorical distributions for \certain pixels decreases with respect to those for the $uncertain$ pixels.
% and so as the loss is only computed for $\mathrm{certain}$ pixels, the categorical distributions for $\mathrm{certain}$ pixels are 
% A contributing factor to this is that, on average, $p(y|\bar{x}'_T)$ has sharp peaks.
% This is because, in contrast to $p(y|\bar{x}_T)$, it is calculated as the softmax function applied to logits produced by a segmentation network trained with a supervised loss, which is frequently observed that produce high confidence \cite{on_the_calibration}.
% The categorical distributions are much sharper for segmentation networks due to the unbounded nature of the logits and the supervised objective, as opposed to prototype segmentation, where the logits are bounded in the range of cosine similarities, $[-1, 1]$.
% This further separates the $\mathrm{certain}$ and $\mathrm{uncertain}$ target features $z_{T}$, as $\mathrm{certain}$ features are pulled closer to the prototypes, and thus pushed away from the $\mathrm{uncertain}$ features.
Given that $p(y|\bar{x}_T)$ is produced via prototype segmentation (see \cref{subsec:proto_seg}), this relates to an increase in the separation between the features $z_{T}$ of \certain and \uncertain pixels.
% is produced via prototype segmentation (see \cref{subsec:proto_seg}), this increases the separation between pixel-wise target features $z_{T}$ that are $\mathrm{certain}$ and $\mathrm{uncertain}$.
This is because \certain features, i.e. those with a cosine distance less than $\gamma$ to a prototype (see \cref{fig:system_overview}), have been pulled closer to their closest prototype, and thus further from the \uncertain features.
% This corresponds to pushing the $\mathrm{certain}$ target features, $z_{T}$, closer to the prototypes, and further from 
% Through optimising $L_\mathrm{c}$, both consistent and inconsistent pixels, and thus accurate and inaccurate pixels are separated in feature space and, through $\gamma$, assigned as certain and uncertain respectively.
% This has the effect of both embedding the certain target features closer to the prototypes and improving segmentation quality over known pixels.
% Importantly, it does this without penalising the model for expressing uncertainty.

\begin{figure*}[h!]
     \centering
     \begin{subfigure}[b]{0.24\textwidth}
         \centering
         \frame{\includegraphics[width=\textwidth]{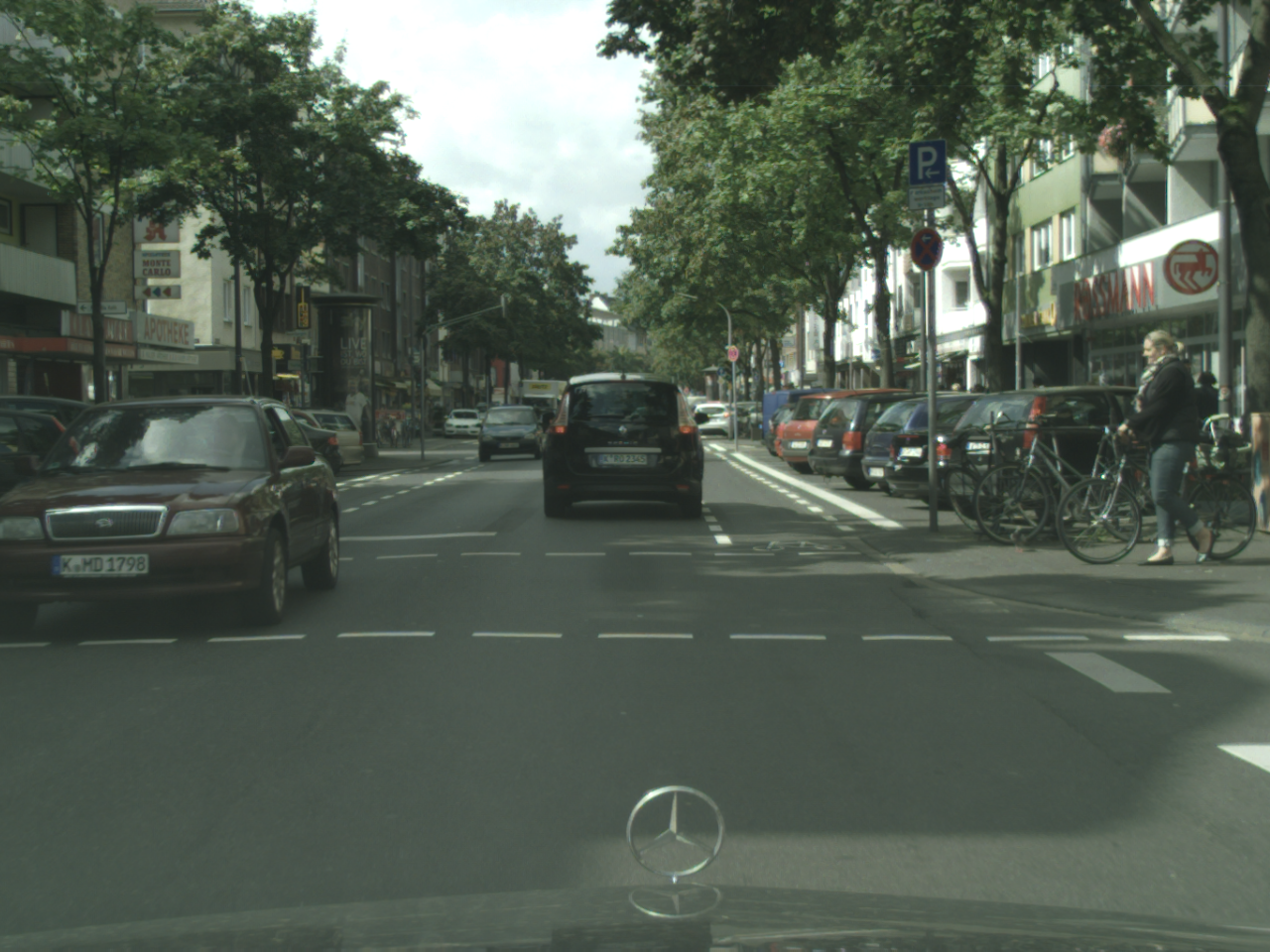}}
         % \caption{Cityscapes}
     \end{subfigure}
     \hfill
     \begin{subfigure}[b]{0.24\textwidth}
         \centering
         \frame{\includegraphics[width=\textwidth]{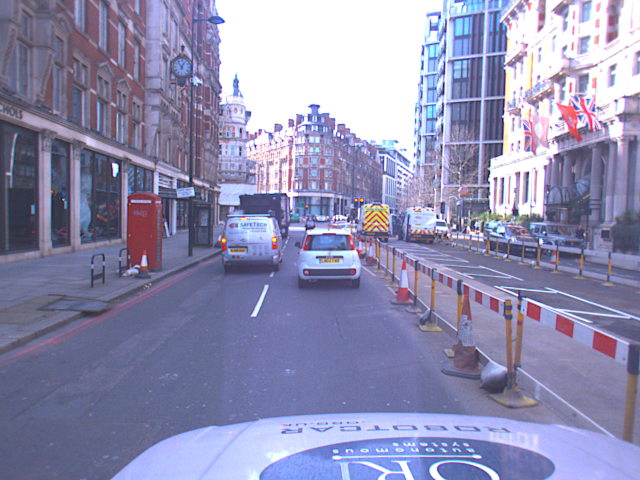}}
         % \caption{SAX London}
     \end{subfigure}
     \hfill
     \begin{subfigure}[b]{0.24\textwidth}
         \centering
         \frame{\includegraphics[width=\textwidth]{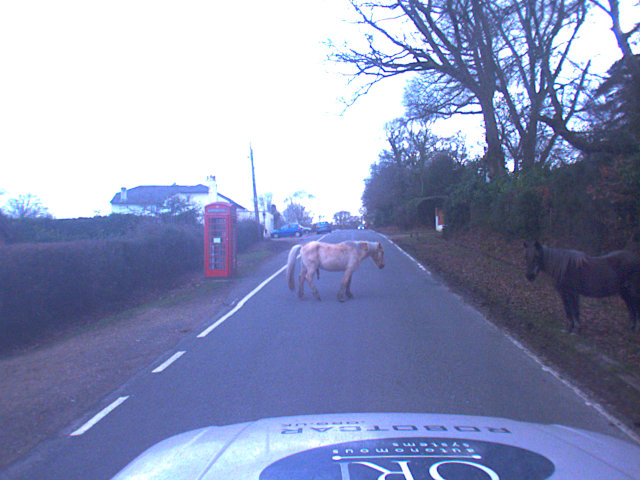}}
         % \caption{SAX New Forest}
     \end{subfigure}
     \hfill
     \begin{subfigure}[b]{0.24\textwidth}
         \centering
         \frame{\includegraphics[width=\textwidth]{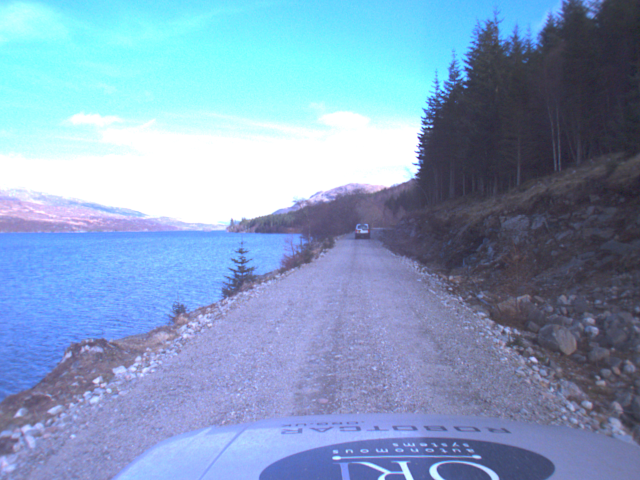}}
         % \caption{SAX Scotland}
     \end{subfigure}
     
     \begin{subfigure}[b]{0.24\textwidth}
         \centering
         \frame{\includegraphics[width=\textwidth]{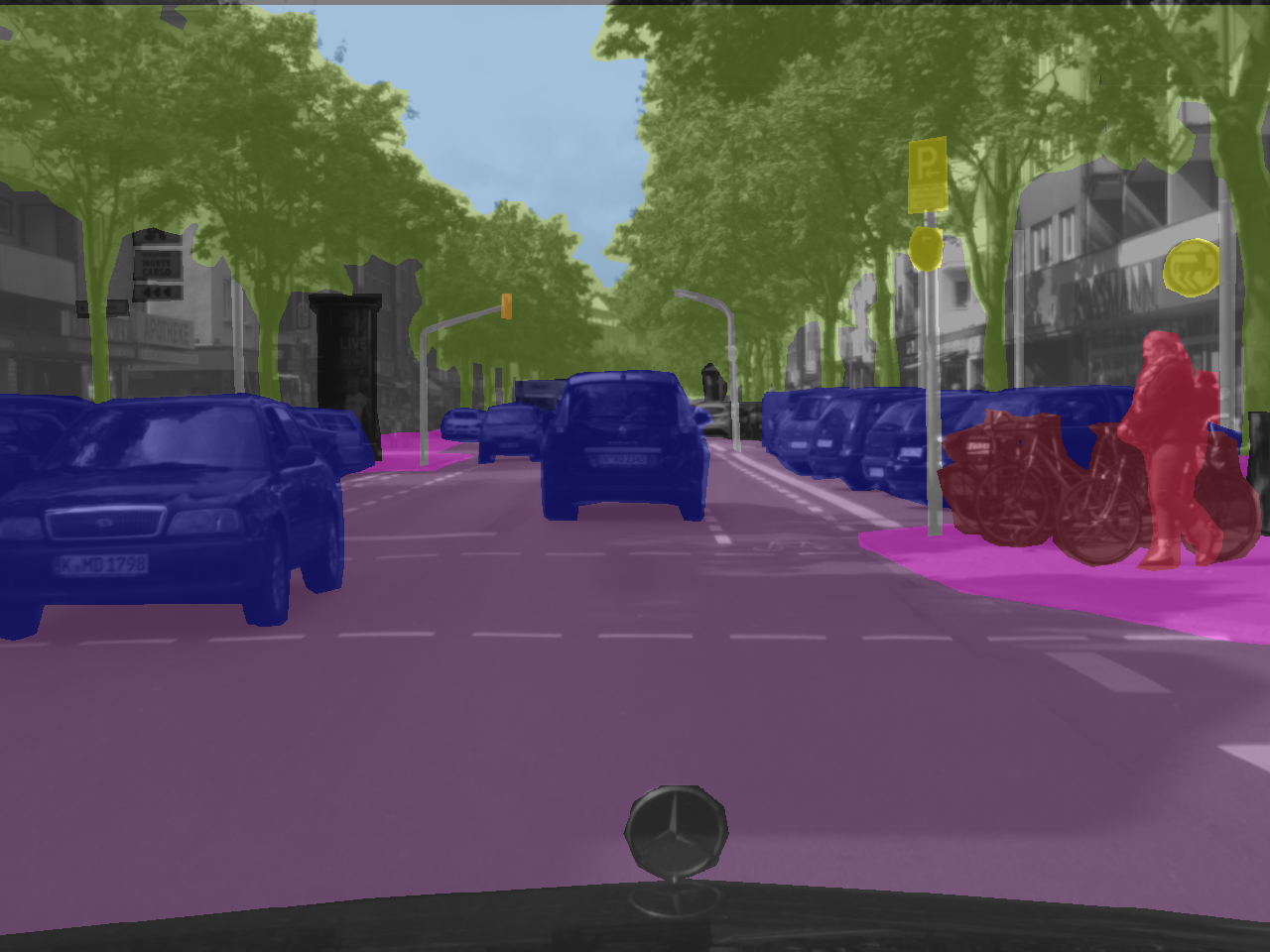}}
         \caption{Cityscapes}
     \end{subfigure}
     \hfill
     \begin{subfigure}[b]{0.24\textwidth}
         \centering
         \frame{\includegraphics[width=\textwidth]{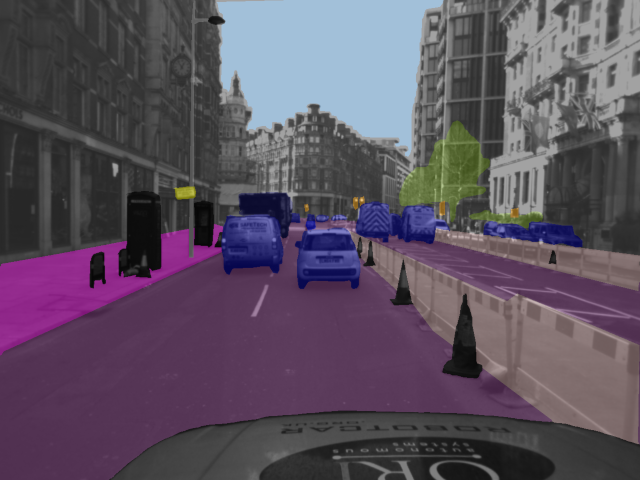}}
         \caption{SAX London}
     \end{subfigure}
     \hfill
     \begin{subfigure}[b]{0.24\textwidth}
         \centering
         \frame{\includegraphics[width=\textwidth]{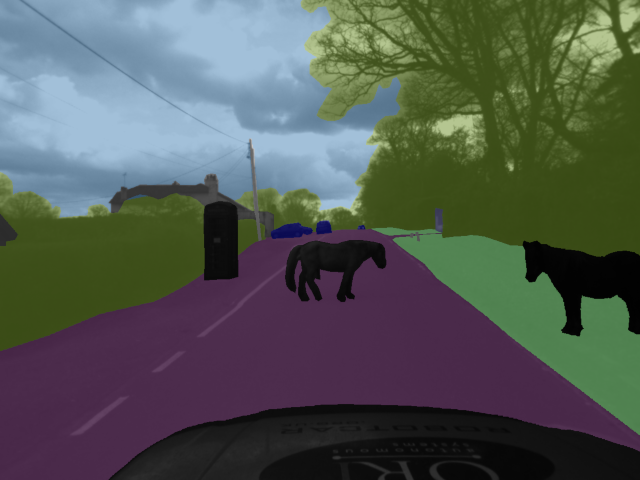}}
         \caption{SAX New Forest}
     \end{subfigure}
     \hfill
     \begin{subfigure}[b]{0.24\textwidth}
         \centering
         \frame{\includegraphics[width=\textwidth]{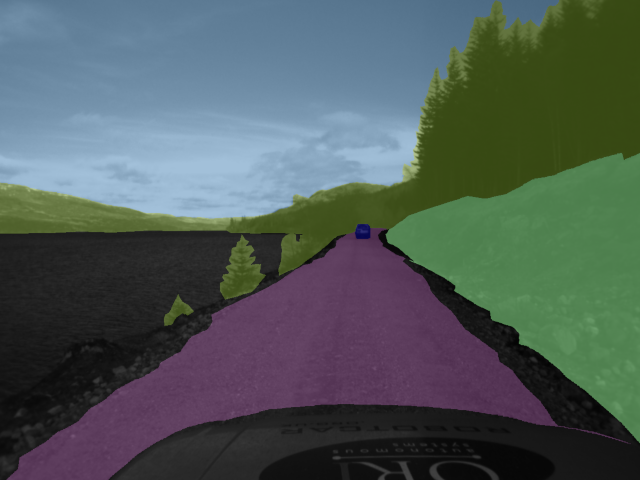}}
         \caption{SAX Scotland}
     \end{subfigure}
        \caption{Example images and relative segmentation masks from Cityscapes -- the source domain -- and the domains in the \textit{SAX Segmentation Test Dataset}.\label{fig:example_images}}
\end{figure*}

\subsection{Additional Objective Functions} \label{add_objectives}
% If the only objectives used were those in the previous section, the achieved uncertainty estimates would be fairly inaccurate.
% This is because the model is strongly encouraged to segment images consistently between views, and thus embed target data close to the source prototypes. 
% This means that all target features can collapse onto a subset of the source prototypes to achieve this objective, without accurate segmentation or uncertainty estimation.
$L_\text{c}$ does \textit{not} maximise consistency for \uncertain pixels.
Despite this, when only $L_\text{c}$ is used to update the model, \emph{all} features in the target domain tend to collapse onto a subset of the source prototypes, thereby achieving near-perfect consistency irrespective of the input image.

This negatively affects uncertainty estimation, as the calculated $\gamma$ cannot effectively separate \certain and \uncertain features, and so $M_{\gamma}$ is a poor uncertainty estimate.
Additionally, few of the prototypes onto which the features collapse correspond to the correct ground-truth class, so the near-perfect consistency does not correspond to near-perfect accuracy.
% On top of resulting in an inaccurate segmentation model, this behaviour means that $M_c$ also approximates accuracy very poorly.
This results both in an inaccurate model and a $M_c$ that approximates accuracy very poorly.

% This sub-optimal solution is easier for the model to find, but does not satisfy the implicit objective of improving 
% This greatly reduces the performance of both segmentation and uncertainty estimation in the target domain, with the latter being the result of consistency no longer being an accurate estimator of ground-truth segmentation accuracy.

% This is because in this situation, all pixels become certain (they are negligibly close to the prototypes, inside the range of $\gamma$).
% Therefore, to minimise~\cref{eq:Lc}, they must all be made consistent -- therefore snapping onto the prototypes themselves.
% When this happens, the proportion of pixels that are certain and consistent across views tends to \SI{100}{\percent}.
% At the same time, the proportion of target pixels that are consistent across views (a.k.a. $p(consistent)$) therefore also increases dramatically, beyond the proportion of pixels that ground-truth would say is accurate (a.k.a. $p(accurate)$). 
% This greatly reduces the uncertainty estimation accuracy, due to the fact that consistency is no longer an accurate estimator of ground-truth segmentation accuracy.

% For this reason, the model is softly constrained to embed a batch of target features in a uniformly distributed manner on the unit hypersphere, as presented in \cite{uniformity}.
The proposed solution to this problematic training dynamic is to softly constrain the model to distribute each batch of target features uniformly on the unit hypersphere, as presented in \cite{uniformity}.
Uniformity prevents feature collapse by constraining the proportion of features near the prototypes, thus making the model more selective about which features are \certain or \uncertain.
$L_\text{u}$ is calculated in the same form as in \cite{uniformity}:
\begin{equation}
    L_\text{u} = \frac{1}{Nh_{\text{u}}w_{\text{u}}} \sum_{i\neq j} e^{-t || \tilde{Z}_{T}^{(i)} - \tilde{Z}_{T}^{(j)}||_{2}^{2} }
\end{equation}
where $t=2$ and $\tilde{Z}_{T} \in \mathbb{R}^ {N \times F \times h_{\text{u}} \times w_{\text{u}}}$ is a batch of target features which has been downsampled by average pooling such that $h_{\text{u}}, w_{\text{u}} = h/4,w/4$. 
Average pooling reduces the number of pairwise distances calculated, reducing memory usage.
% where $N$ is the batch size, and we choose $t=2$.

Simultaneously, another loss term, $L_\text{p}$ \cite{uniform_proto}, maximises the distance between the source prototypes, i.e. spreads them on the unit sphere, preventing $L_\text{u}$ from concentrating the prototypes to maximise the distance between \certain and \uncertain features.
For $K$ class prototypes $p_S \in \R^{F \times K}$:
\begin{equation}
    L_\text{p} = \frac{1}{K}\sum_{i=1}^{K} \max_{\scriptscriptstyle{j \in K}}[p_S^\top p_S - 2\mathrm{I}]_{ij}
\end{equation}
This minimises the similarity between nearest prototypes, as $p_S^\top p_S \in \R^{K \times K}$ contains cosine similarities between prototypes, and subtracting $2\mathrm{I}$ excludes self-similarity terms.

While $L_\text{u}$ and $L_\text{p}$ both maximise the distance between features, $L_\text{u}$ applies this locally using an \gls{rbf} kernel, whereas $L_\text{p}$ maximises nearest-neighbour feature distance, and thus more strongly encourages uniformity.  

Finally, a supervised loss is calculated using the source labels to maintain a good representation of the source domain.
This objective, $L_\text{s}$, is used to update the encoder and segmentation head and is calculated as the cross-entropy:
\begin{equation}
    L^{(i)}_\text{s} = -\sum_{k \in \mathcal{K}} \ y_S^{(i)}\ \mathrm{log}(p(y=k|x_{S})^{(i)})
\end{equation}
where $y_S^{(i)}$ is the ground-truth one-hot label for pixel $i$. %as in~\cref{sec:segprelim}.

\subsection{Asymmetric Branches} \label{subsec:asymm}
% \dw{$p(y_i=k|\bar{x}'_T) = h(E(I_{T1})$ where $h(.)$ is the segmentation head.} 
% \dw{$p(y_i=k|\bar{x}_T) = \sigma_{\tau}(g(E(\bar{x}_T) \boldsymbol{p_{S}})$}
As seen in \cref{fig:system_detail}, the branches segmenting $\bar{x}'_T$ and $\bar{x}_T$ are not identical.
The top branch, $\texttt{f}: \mathbb{R}^{3 \times H \times W} \rightarrow \mathbb{R}^{K \times H \times W}$, segments $\bar{x}'_T$ through $\texttt{E}$ and a segmentation head rather than prototype segmentation; therefore $\texttt{f}(\cdot) = \texttt{f}_{\psi} \circ \texttt{E}(\cdot)$, where $\texttt{f}_{\psi}(\cdot)$ is the segmentation head. 
In contrast, the bottom branch, $\texttt{g}: \mathbb{R}^{3 \times H \times W} \rightarrow \mathbb{R}^{K \times H \times W}$, segments $\bar{x}_T$ as $\texttt{g}(\cdot) = \texttt{g}_{\pi} \circ \texttt{g}_{\rho}\circ\texttt{E}(\cdot)$, where ${\texttt{g}_\rho(\cdot)}$ is a projection network and $\texttt{g}_{\pi}(\cdot)$ performs prototype segmentation as $\texttt{g}_{\pi}(z_{T}) = z_{T}^{\top}p_S$.

% does not use prototypes, but an additional neural network, named the segmentation head, comprising a single $1 \times 1$ convolutional layer. % which is only updated with the supervised objective, $L_\text{s}$.
% Additionally, the encoder on this branch does not include a projection network, as depicted in \cref{fig:system_detail}.

% $\texttt{f(}\cdot\texttt{)} = \texttt{f}_{\psi} \circ \texttt{E(}\cdot\texttt{)}$

% $\texttt{g(}\cdot\texttt{)} = \texttt{g}_{\pi} \circ \texttt{g}_{\rho} \circ \texttt{E(}\cdot \texttt{)}$

Branch asymmetry prevents training from collapsing.
$L_\text{c}$ can be trivially minimised by assigning large regions to the same incorrect class across views, similar to what is described in \cref{add_objectives}.
Similar architectures are found in self-supervised learning methods such as \cite{moco,byol}, also prevent similar failure modes -- \cite{moco} using exponential moving averages of model weights, and \cite{byol} using additional layers.

As a result, the proposed asymmetry is not susceptible to collapse.
Suppose that the encoder $\texttt{E}$ collapses, for $L_{\text{c}}$ to be minimised by $s'_{T}=s_{T}$, the following has to be true: $\texttt{f}_{\psi}(\cdot)=\texttt{g}_{\pi} \circ \texttt{g}_{\rho}(\cdot)$.
This is not observed, which we attribute to the following factors: (1) $\texttt{g}_{\pi} \circ \texttt{g}_{\rho}$ and $\texttt{f}_{\psi}$ are architecturally different (2) neither the segmentation head $\texttt{f}_{\psi}$, nor the prototypes via $\texttt{g}_{\pi}$ can contribute to a degenerate solution, as neither are updated by $L_{\text{c}}$ (3) $\texttt{g}_{\rho}$ is updated with $L_\text{u}$ in addition to $L_\text{c}$.

% This means that even $L_\text{c}$ causes a collapse the the feature representation for the target domain, 
% This means that the entire top branch in \cref{fig:system_detail} is optimised at every iteration to segment the labelled source data, which reduces the like

% and so this degenerate solution no longer exists as a minimum on the loss landscape.

A secondary benefit of using the segmentation head $\texttt{f}_\psi$ is that it naturally produces a low entropy $p(y|\bar{x}'_T)$.
This contributes to decreasing the entropy of $p(y|\bar{x}_T)$, thus further separating \certain and \uncertain pixels in the manner described in \cref{subsec:learning_E}.

\section{Experimental Setup} \label{sec:experiments} 	
%------------------------------------------------------------------
% Sub-Sections are: Data, Performance Metrics, Benchmarks, Experiments
% Each of need to be introduced in this section, with further detail in sub-sections themselves.

% Data
This work introduces a novel test benchmark, building on top of the Sense-Assess-eXplain (SAX) project \cite{sax}.
The benchmark is composed of pixel-wise labels that annotate a set of manually curated images from three domains of the SAX project.
Alongside the test labels, this benchmark also proposes test metrics to evaluate quality of uncertainty estimation, presented in \cref{sec:metrics}.
\subsection{Data}  \label{subsec:data}
% Firstly, the test data in this work is pixel-wise annotated, rather than image-wise, thus making the test task much harder.
% During training, this method uses two types of data.
This work uses three different types of data: (1) labelled training images (2) unlabelled training images (3) labelled test images.
The primary experiments in this work use Cityscapes~\cite{cityscapes} as (1), and a SAX domain provides (2) and (3).
As such, unless otherwise stated, the labelled dataset used is Cityscapes.
However, in order to investigate the generality of the method, \gls{bdd}~\cite{bdd} is also considered as a source domain, and both \gls{bdd} and KITTI~\cite{kitti} are used as target domains.

% These datasets define the visual semantic concepts for the task, and so the annotations for the SAX test dataset conform to Cityscapes' and \gls{bdd}'s semantic definitions.

The SAX dataset comprises data from three domains defined by their location of collection: London, the New Forest (a rural region in southern England), and the Scottish Highlands, ordered by descending similarity with Cityscapes.
Examples from each dataset can be seen in \cref{fig:example_images}.
% These each pose a unique challenge, however one would say that the order stated previously (London $\rightarrow$ New Forest $\rightarrow$ Scottish Highlands) is an order of increasing dissimilarity with the source domain, Cityscapes, which comprises of images of driving scenes in German cites. 
By testing across all three domains, the effect of the magnitude of the distributional shift on uncertainty estimation can be evaluated.

Each domain contains instances that are in-distribution (e.g. cars, road, signs that look very similar to those found in Cityscapes) and \gls{ood} (e.g. classes not defined in Cityscapes such as horses, Scottish lochs, gravel roads).
Classes undefined in source domain are treated such that any assignment to these classes is treated as $\mathrm{inaccurate}$.
Importantly, each domain also contains many instances on the edge of the labelled distribution, i.e. \textit{near-distribution}.
% Notably, each image can contain a combination of in-distribution, \gls{ood} and near-distribution instances.
Each of these instance types are combined within images, causing a significant challenge to uncertainty estimation and \gls{ood} detection models.
% Within each image, a combination of in-distribution, \gls{ood} and near-distribution instances can be found, which causes a significant challenge to uncertainty estimation and \gls{ood} detection models.

For the KITTI labelled test dataset, the 201 labelled training images were used (as only these have downloadable labels), while the unlabelled training dataset come from the published raw data. 
% The raw data is split up into `drives', from which the labelled data was taken.
\gls{bdd} publishes both a labelled training dataset, test dataset and 100,000 driving images without semantic annotation.
This allows us to use \gls{bdd} as both source and target domain separately. 
For both KITTI and BDD, care was taken to prevent any overlap between labelled testing images and the unlabelled training images.
% Care was taken to prevent overlap between the labelled testing images and the unlabelled training images, as such the training images come from the entirely separate `drives' to the test data.
% Care is similarly taken to prevent overlap between the test dataset and the unlabelled images.

% Often testing for \gls{ood} detection or epistemic uncertainty estimation involves a much easier task, such as separating distinct datasets that are mixed together in an image-wise manner.

% In order to overcome that significant challenge of a given SAX test domain, the unlabelled SAX dataset from that same domain is used during training.
% This is because large quantities of unlabelled data can be obtained with no significant cost.
% In preparing the unlabelled SAX dataset, care was taken to avoid having any overlap between the unlabelled training data and the labelled test data.
% The same reason given as to why the SAX test dataset so challenging, makes the SAX training dataset very useful for training, see \cref{ood_lit}.

\subsection{Network Architecture \& Training}   \label{subsec:net_arch}
For every experiment, the segmentation network used has a DeepLabV3+ architecture \cite{deeplabv3plus} with a ResNet18 backbone \cite{resnet}.
More specifically, $\texttt{E}$ is represented by both the ResNet18 and the ASPP module, and so the features $\hat{z}$ are taken from the penultimate layer of DeepLabV3+, with just the segmentation head $\texttt{f}_\psi$ to follow. \footnote{see \url{https://github.com/qubvel/segmentation_models.pytorch} for implementation details}
For prototype segmentation, features $\hat{z}$ are then passed through a projection network $\texttt{g}_{\rho}$, which is 
a two-hidden-layer perceptron -- similar to \cite{simclr}, but applied to each pixel embedding independently.
% The feature space used in this work is taken after the ASPP module and a projection network -- a two-hidden-layer perceptron that projects every pixel feature independently -- as presented in \cite{simclr}.
% This means $\texttt{E}$ represents the ResNet, ASPP module and the projection network: $E: \R^{N \times 3 \times H \times W} \rightarrow \R^{N \times F \times h \times w}$.
% The spatial dimensions of the features are given by ${h=\sfrac{H}{4}}, {w=\sfrac{W}{4}}$, and for the feature length, $F=256$.
The feature dimensions are given by ${h=\sfrac{H}{4}}, {w=\sfrac{W}{4}}$, $F=256$.

Before being updated by $L_{\text{c}}$, the networks $\texttt{E}$ and $\texttt{g}_{\rho}$ are pre-trained using only $L_\text{s}$ and $L_\text{u}$.
Firstly, this means that before semi-supervised training begins, the segmentation head $\texttt{g}_{\rho}$ has already broadly learned the spatial distribution of classes.
Secondly, after a small number training iterations with $L_{\text{c}}$ and $L_{\text{p}}$, the prototypes faithfully represent the semantic classes due to the pre-training of $\texttt{E}$.
In combination, these factors mean that $M_{c}$ starts as a better estimate of ground-truth segmentation accuracy, and thus the system is well-initialised for the positive feedback loop described in \cref{sec:training}.

% the prototypes already faithfully represent the semantic classes, and the segmentation head has learned to decode from

% WHAT DOES THIS MITIGATE? WHAT DOES IT DO?
% Resultantly at the start of training, $s'_{T}$, i.e. the pseudo-label, is already fairly accurate, and the consistency between $s'_{T}$ and $s_{T}$ is very low.
% The latter makes it more difficult for a pair of pixels to become consistent, thus reducing the likelihood of incorrect pixels being consistent.

%------------------------------------------------------------------
\subsection{Use of a Domain-based Curriculum} \label{subsec:curriculum}
% \subsection{$\mathrm{\gamma\text{-}SSL}$ vs. $\mathrm{\gamma}\text{-}\mathrm{SSL_{iL}}$} \label{subsec:curriculum}
% \subsection{$\mathrm{\gamma}\text{-}\mathrm{SSL}$ \& $\mathrm{\gamma}\text{-}\mathrm{SSL_{iL}}$} \label{subsec:benchmarks}
\fixM{Models trained with the method presented in \Cref{sec:system_ov} are given the name \gammassl.}{2.1\label{comm:2.1_2}}
For each domain, a separate \gammassl model is trained, such that testing occurs in the domain of the unlabelled training data.
% The same is broadly true for the $\mathrm{\gamma}\text{-}\mathrm{SSL_{iL}}$ models; however, these are initialised on a model trained with unlabelled SAX London images.
% $\mathrm{\gamma}\text{-}\mathrm{SSL_{iL}}$ models are however also initialised on weights trained with unlabelled SAX London images.
\fixM{Models named $\mathrm{\gamma}\text{-}\mathrm{SSL_{iL}}$ are however also initialised on weights trained with unlabelled SAX London images.}{2.1\label{comm:2.1_3}}

% Given that SAX London is the domain most similar to the source domain (Cityscapes), 
$\mathrm{\gamma}\text{-}\mathrm{SSL_{iL}}$ models are trained under the following hypothesis: splitting training into two chunks (source $\rightarrow$ intermediate target, intermediate target $\rightarrow$ final target), reduces the distributional gap between source and target for each chunk of training, and this improves the quality of the learned representation of the final target domain.
SAX London is this intermediate domain, as it is most similar Cityscapes, while also sharing platform configurations with other SAX domains.
% At the same time, it shares platform configurations with the others.

The motivation for this comes from curriculum learning \cite{curriculum_learning}, where a model's performance is improved by using a training procedure in which the difficulty of training examples increases during training.
In our case, the difficulty of the curriculum is controlled by one high-level characteristic of the domain, i.e. its geographic location, but other characteristics could be used, such as rainy/dry, day/night, sunny/overcast.

Given that robots are designed for a specific \gls{odd}, the source domain is precisely defined by its operating conditions.
By considering what the \gls{odd} considers in-distribution, the curriculum can be designed to include progressively more \gls{ood} conditions.
The added diversity in this curriculum naturally increases the data requirements for \gammasslil.
However, in a robotics context, this work argues that these requirements are not difficult to satisfy.
This is because collecting an uncurated unlabelled dataset for a specific set of operating conditions (e.g. those not contained in the \gls{odd}) merely requires access to a robot and for those conditions to exist in the real world.
This argument also explains why the standard \gammassl models are not significantly more difficult to train than the benchmarks.

% It should be noted that the data requirements \gammasslil with respect to \gammassl, which in some settings could prove challenging.

% $\mathrm{\gamma}\text{-}\mathrm{SSL_{iL}}$ models naturally come with the added cost of the unlabelled data from two domains, rather than one (or zero for the benchmarks).
% % Depending on the context however, this cost may not be that great, due to the lack of labelled requirements and the prevalence 
% In the context of robotics, this work makes the argument that the collection of unlabelled data is extremely cheap (in terms of time and resources).
% This is because it is presumed that access to a relevant robot is not limited (given that a perception system is being designed for it), thus the means to collect data exists.
% The only requirement is therefore to have the robot traverse environments with

% it does not seem that this is exceedingly costly.
% If access to the robot isn't limited, then these additional unlabelled images can be collected by a traversal of 

% then the cost, in terms of time, is simply 

% then unlabelled data can be collected with great ease
% given that the requirement is simply to traverse an 

%------------------------------------------------------------------

%------------------------------------------------------------------
\subsection{Benchmarks} \label{subsec:benchmarks}
% This work evaluates a range of techniques, each producing a scalar value for each pixel that can be interpreted as the estimated likelihood of the model making an error. 
This work evaluates a range of techniques, each producing a likelihood per-pixel of the model making an error. 
As the test data is distributionally shifted from the labelled training data, the proposed method is benchmarked against several epistemic uncertainty estimation and \gls{ood} detection techniques.

Methods are split into \textit{epistemic} and \textit{representation} based methods.
The distinction is that the epistemic methods consider a distribution over the model parameters, sample from that distribution and then calculate the inconsistency in segmentation; this process requires multiple forward passes of the network.
Instead, the representation methods solely leverage a learned representation and compute a metric to determine the uncertainty in a single forward pass, greatly reducing the computational requirements for deployment.
% The downside is that often representation-based methods are shown to achieve lower-quality uncertainty estimates.

The epistemic methods comprise Monte-Carlo Dropout \cite{dropout} (\mcd) and Deep Ensemble \cite{ensembles} (\ensemble).
For both, \gls{pe} and \gls{mi} are used as uncertainty measures, where \gls{mi} is more often used to estimate epistemic uncertainty, as evidenced by its use in active learning \cite{gal_thesis}.
The network for \mcd builds on Bayesian DeepLab \cite{eval_bayesian_seg}, but is adapted for a ResNet18, and is tested over a range of dropout probabilities and numbers of samples, with the best presented ($0.2$ and $8$ respectively).
Different ensemble sizes are also evaluated.
The \mcd model is also distilled into a deterministic network, named \mcddistil, as per \cite{mcd_distillation}.

As for the representation methods, the techniques investigated include several \gls{ood} detection methods \cite{maxsoftmax, odin,mahalanobis,vim}.
\cite{maxsoftmax}  (\softmax) and \cite{odin} (\softmaxA) propose a pretrained segmentation network with tuned softmax temperature parameters, where the latter also leverages adversarially perturbed images.
\cite{mahalanobis} leverages a pretrained network where \gls{ood} score is the Mahalanobis distance in feature space (\featdist), which can also leverage adversarial inputs (\featdistA).
Finally, \cite{vim} (\vim) defines the \gls{ood} score as a function of both the features and the logits. 
The feature space chosen for \cite{mahalanobis,vim} is the same as used for the proposed method.
When using adversarially perturbed inputs, evaluation takes place over a range of step-sizes, $\epsilon$.
The final representation method is a Deterministic Uncertainty method (\dum), presented in \cite{dum_baseline}, and uses the official implementation of \cite{invertible_resnet}.

% As described previously, the error rate can be reduced on the target datasets by both domain adaptation and uncertainty estimation.
% For this reason, both domain adaptation and uncertainty estimation techniques are investigated as benchmarks in the work.
% Additionally, it has been observed that uncertainty estimates degrade as the distributional shift between training data and test data gets larger, see \cite{bad_uncertainty}.
% This means that it might be expected that if domain adaptation is performed and a better representation of the target domain is acquired, then the uncertainty estimates should also be better.
% This is also investigated in this paper.
% Benchmarks for uncertainty estimation comprise of predominantly of epistemic uncertainty estimation methods, Monte Carlo Dropout \cite{dropout} and Ensembles \cite{ensembles}.
% Both of these techniques are computationally expensive at inference time, however they have been shown to be the most effective benchmark methods for estimating distributional uncertainty.
% Using the entropy of model's output categorical distribution over classes as a measure of predictive uncertainty is also investigated as a benchmark.
%------------------------------------------------------------------

%------------------------------------------------------------------
\section{Evaluation Metrics} \label{sec:metrics}
This work evaluates methods on their ability to perform misclassification detection.
This is a binary classification problem, whereby pixels segmented accurately with respect to labels should be classified as \certain, and inaccurate pixels classified as \uncertain.
These states are defined in~\cref{tab:confusion_matrix}.

\begin{table}[!h]
\renewcommand{\arraystretch}{2}
\caption{ Confusion Matrix for Misclassification-Detection \label{tab:confusion_matrix} }
\centering
\begin{tabular}{cccc}
    &   & \multicolumn{2}{c}{$\mathrm{Predicted}$} \\ \cline{3-4} 
    & \multicolumn{1}{c|}{} & \multicolumn{1}{c|}{$\mathrm{[certain]}_t$} & \multicolumn{1}{c|}{$\mathrm{[uncertain]}_t$} \\ \cline{2-4} 
\multicolumn{1}{c|}{\multirow{2}{*}{\rotatebox[origin=c]{90}{$\mathrm{Actual}$}}} & \multicolumn{1}{c|}{$\mathrm{accurate}$}   &   \multicolumn{1}{c|}{\tp}              & \multicolumn{1}{c|}{\fn}                   \\ \cline{2-4} 
\multicolumn{1}{c|}{}  & \multicolumn{1}{c|}{$\mathrm{inaccurate}$} & \multicolumn{1}{c|}{\fp} & \multicolumn{1}{c|}{\tn} \\ \cline{2-4} 
\end{tabular}
\begin{tablenotes}
\item The problem is set as a binary-classification with: True Positive (\tp), False Negative (\fn), False Positive (\fp) and True Negative (\tn).
\item Each of the states $\mathrm{certain}$, $\mathrm{uncertain}$ are determined by a given uncertainty threshold $t$.
\end{tablenotes}
\end{table}

% Each of the possible metrics that can be used to evaluate model's on binary classification define the ideal model in a distinct manner.
Given a set of imperfect models, the best model for a binary classification problem can be selected based on a number of different metrics. 
For uncertainty estimation, the most appropriate metric is dependent on the context in which the uncertainty estimates and segmentation predictions are used.
For this reason, this work considers a range of possible definitions of metrics and justifies each in a robotics context.
% \cref{subsec:metric_definitions} defines each of the metrics used, and \cref{subsec:metric_discussion} justified each metric in a robotics context.

\subsection{Metrics: Definitions} \label{subsec:metric_definitions}
% AUROC & AUPR
Firstly, this work considers \gls{roc} curves and \gls{pr} curves for the evaluation of misclassification detection, based on the prior use of these metrics in~\cite{maxsoftmax}.
\gls{roc} curves plot the true positive rate ($\mathrm{TPR}$) versus the false positive rate ($\mathrm{FPR}$):
\begin{equation*}
	\mathrm{TPR} =  \mathrm{\frac{TP}{TP+FN}},\ \mathrm{FPR} = \mathrm{\frac{FP}{FP+TN}}
\end{equation*}
% Here, $\mathrm{TPR}$ is the probability that an actual positive will test positive, while $\mathrm{FPR}$ is the probability that an actual negative will test negative.
Here, $\mathrm{TPR}$ is the proportion of $\mathrm{accurate}$ pixels detected as \certain, whereas $\mathrm{FPR}$ is the proportion of $\mathrm{inaccurate}$ pixels incorrectly assigned to \certain.
% In this context, the area under this curve, $\mathrm{AUROC}$, can be interpreted as the probability that a model's uncertainty estimates ranks a randomly chosen accurate pixel higher than a randomly chosen inaccurate pixel.
The \gls{roc} curve treats the positive and negative classes separately and is thus independent of the class distribution, i.e. the underlying proportion of pixels segmented as the correct semantic class.
The \gls{roc} curve is summarised by the area under it, the \auroc.%, as reported in \cref{tab:auroc-aupr}.

Precision and recall are defined respectively as:
\begin{equation*}
    \mathrm{Precision} = \frac{\mathrm{TP}}{\mathrm{TP+FP}},\ \mathrm{Recall} = \frac{\mathrm{TP}}{\mathrm{TP+FN}}
\end{equation*}
% Precision considers: of the pixels that are classified as certain, what proportion are accurate? 
% While recall considers: of the pixels that are accurate, what proportion are certain?
% Typically, a balance of these is desired, and so again the area under this curve, \aupr, is reported in \cref{tab:auroc-aupr}.
% Drawing on the information retrieval interpretation of PR curves and the definitions in\cref{tab:confusion_matrix}, models are evaluated with on their ability .
Interpreting misclassification detection as an information retrieval task, \gls{pr} curves evaluate the ability to use uncertainty estimates to retrieve \emph{only} accurate pixels (as the positive class is defined as $\mathrm{accurate}$ in \cref{tab:confusion_matrix}).
% PR curves are used in information retrieval, where it is desired that the positive class is retrieved.
% As defined in \cref{tab:confusion_matrix}, the positive class is defined as $\mathrm{accurate}$, and so models are evaluated on their ability to retrieve accurate pixels, as these are what are ultimately useful downstream.
% PR curves are calculated independent of the number of true negatives, and thus excel when true negatives unbalance the classification problem.
%PR curves are shown in \cref{fig:main_results_plot}, as well as summarised by their area, \aupr, which is reported in \cref{tab:auroc-aupr}.
As for \gls{roc} curves, \gls{pr} curves can be summarised by the area under them, the \aupr.

% F score
Additionally, as an alternative to \aupr, the $\mathrm{F_{\beta}}$ score is also considered, defined as:
\begin{equation*}
	\mathrm{F_{\beta}} = \frac{(1+\beta^{2})\mathrm{TP}}{(1+\beta^{2})\mathrm{TP}+\mathrm{FP}+\beta^{2}\mathrm{FN}}
\end{equation*}
This factors precision and recall into a single metric, weighting their contribution through the scalar $\mathrm{\beta}$.
For $\mathrm{\beta} < 1$ a stronger focus is given to $\mathrm{Precision}$, while for $\mathrm{\beta} > 1$ to $\mathrm{Recall}$.
%, while for $\mathrm{\beta}\rightarrow\infty$ we consider purely recall.
%This work considers $\mathrm{\beta}=0.5$, thus prioritising precision over recall.

% Accuracy
Finally, as in~\cite{eval_bayesian_seg}, the misclassification detection accuracy, \Amd, is also used to evaluate uncertainty estimation.
This is defined for a given uncertainty threshold as:
\begin{equation*}
    \mathrm{A_{MD}}=\frac{\mathrm{TP}+\mathrm{TN}}{\mathrm{TP}+\mathrm{TN}+\mathrm{FP}+\mathrm{FN}}
\end{equation*}
According to this metric, the best model is the one which segments the highest proportion of \textit{all pixels} in one of two states: $(\mathrm{accurate}, \mathrm{certain})$ or $(\mathrm{inaccurate}, \mathrm{uncertain})$. % with no preference shown to either.
% For this metric to be an appropriate choice, it is important that the misclassification cost for a false positive and a false negative is the same.

% Given that a useful model should also ideally assign as many pixels as possible to the classes of interest $[\mathrm{A_{MD}}]_t$ is plotted for every threshold against the proportion of pixels that are in the state  as $\mathrm{certain}$, named $\mathrm{coverage}$, where:

% In quantifying the model's misclassification detection performance, some notion of the `safety' of the model is captured.
% It is however also important to quantify the `usefulness' of the model, which is a function of how many pixels are correctly assigned to a semantic class.

$\mathrm{A_{MD}}$ and $\mathrm{F_{0.5}}$ are plotted against the proportion of pixels that are in the state: $(\mathrm{accurate}, \mathrm{certain})$, named $\mathrm{p(a, c)}$, where:
\begin{equation*}
	\mathrm{p(a, c)} = \frac{\mathrm{TP}}{\mathrm{TP}+\mathrm{TN}+\mathrm{FP}+\mathrm{FN}}
\end{equation*}
For these plots, %seen in \cref{fig:main_results_plot}, 
the ideal model should maximise both the uncertainty metric ($\mathrm{A_{MD}}$, $\mathrm{F_{0.5}}$) and $\mathrm{p(a, c)}$, i.e. better results are closer to the top-right of the plots.
Therefore, the maximum value of \Amd and \fhalf\ -- named \maxAmd and \maxfhalf respectively -- and the value of $\mathrm{p(a, c)}$ at which they occur are also reported.%, seen in \cref{tab:max_A_MD} and \cref{tab:max_fhalf}.

%---------------------------------------------------------------------------------------------------
\subsection{Discussion: AUROC, AUPR and F-scores} \label{subsec:metric_discussion}
Given the context of this work, i.e. semantic segmentations for robotics applications, we must consider:
\begin{enumerate}
\item what the real-world costs are for misclassification in the cases of FP versus FN, and 
\item whether the evaluation should be independent of the class distribution, $\mathrm{p(accurate)}$
\end{enumerate}

The first is context-dependent to a large extent.
In general, for safety-critical contexts such as robotics, where perception directly leads to decisions and actions in the real world, the importance of $\mathrm{certain}$ being $\mathrm{accurate}$ is higher than \uncertain being $\mathrm{inaccurate}$.
To simplify, \textit{accidents arise when autonomous systems make confident but incorrect predictions about their surroundings}.
In contrast, when predictions are \uncertain but $\mathrm{accurate}$, the system is overly conservative and thus does not take action as it considers some safe actions unsafe; this is inefficient but less hazardous.
It can therefore be argued that precision is more important than recall -- we want $\mathrm{FP}=0$, i.e. no pixels $\mathrm{inaccurate}$ and \certain even if some pixels are $\mathrm{accurate}$ but \uncertain ($\mathrm{FN}>0$).

\Gls{pr} and \gls{roc} curves present the performance of a model over the full range of relative misclassification costs.
\auroc and \aupr assess how models perform in aggregate over this range of relative misclassification costs, and thus over a range of robotics contexts.
Consequently, however, these metrics do not fully represent whether a model is appropriate in a \emph{specific} context.
To represent a specific context of interest, $\mathrm{F_{\beta}}$ scores can aggregate \gls{pr} curves with a preference towards precision or recall.
In the effort to prioritise precision, as argued above, $\mathrm{F_{0.5}}$ scores are presented in this work.

% \dw{
% How is this different or better than $F_1$ score? If p(accurate) is around 50\% then accuracy is fine. If accuracy is much larger than 50\% then accuracy is biased toward models that say lots of stuff is certain, but this is fine is model is very accurate, this is a good outcome.
% }

\subsection{Discussion: Misclassification Detection Accuracy}
% Precision = Recall

Misclassification detection accuracy (named \Amd to avoid confusion with segmentation accuracy) provides an intuitive understanding of misclassification detection performance by reporting the proportion of \textit{all} pixels (the denominator considering the entire image grid) in either of the following `safe' states: $(\mathrm{accurate}, \mathrm{certain})$ or $(\mathrm{inaccurate}, \mathrm{uncertain})$; no preference between FP and FN is expressed.

Segmentation networks are typically one component of a larger system, and there are some robotics contexts where FP and FN are not drastically different in effect.
For example, semantic localisation can reject a \fp using additional processing steps in the localisation pipeline, i.e. RANSAC in geometric refinement.
%, and through this filtering ultimately ``look like'' a TN.
Also, in semantic mapping, multiple views of a location are available, which allow for additional processing, e.g. majority voting, to reduce the effect of errors. 
In both cases, where subsequent steps provide additional filtering, where a \fp is less detrimental to the broader system.

% Another cited reason is that \Amd is not independent of the class distribution.
% Given that 

% Suppose there are two models: one that can perform perfect misclassification detection i.e. $\mathrm{AUROC}=1$, $\mathrm{AUPR}=1$, but has an accuracy of \SI{60}{\%}, and one has an accuracy of \SI{99}{\%}, but classifies every pixel as certain.
% The former model would have $\mathrm{AUROC}=1$, $\mathrm{AUPR}=1$, whereas the latter would have $\mathrm{AUROC}<1$, $\mathrm{AUPR}<1$.

% While unrealistic, this scenario demonstrates that it would be useful to also factor in the semantic segmentation accuracy of the underlying model.

% For these tasks, simply reporting the classification accuracy, \Amd, is therefore more appropriate.
% \Amd simply describes the proportion of pixels in either of the following states: $(\mathrm{accurate}, \mathrm{certain})$ or $(\mathrm{inaccurate}, \mathrm{uncertain})$.

% This means you are likely to accept as many false positives as false negatives, in order to segment more pixels as $\mathrm{certain}$, and thus one of the semantic classes of interest.
% In this case, accuracy is the measure of interest.
%---------------------------------------------------------------------------------------------------

\subsection{Discussion: Class Distribution}
A characteristic of \gls{roc} and \gls{pr} curves is their insensitivity to the class distribution, defined as the proportion of pixels that are $\mathrm{accurate}$ versus $\mathrm{inaccurate}$.
% if $\mathrm{p(accurate)}$ is significantly greater than $\mathrm{p(inaccurate)}$. 
%, then these curves still effectively measure a model's misclassification detection ability.
% then metrics derived from these curves are still effective at measuring how good a model is at misclassification detection.
% In the context of this work, it is not just the case that the modeller can effect change on this class distribution, but actively want to pick a model with a higher accuracy.
Each model will have a different semantic segmentation accuracy, and so the class distributions will vary.
% In a robotics context, each model will have a different class distribution, i.e. each model will have a different semantic segmentation accuracy.
This means that \gls{roc} and \gls{pr} curves provide helpful analysis on the ability to detect misclassification, independent of segmentation accuracy.

In the context of this work, however, if two models have the same misclassification detection performance, the model with the higher $\mathrm{p(accurate)}$ should be chosen.
More specifically, the proportion of pixels assigned, with certainty, to the known semantic classes should be considered, i.e. $\mathrm{p(a,c)}$.
In the interest of jointly considering misclassification detection and semantic segmentation performance, this work presents a procedure to describe both objectives intuitively.

Let's consider $\mathrm{F_{0.5}}$ and \Amd versus $\mathrm{p(a, c)}$.
This allows us to determine the model and uncertainty threshold at which misclassification detection is performed best \emph{and} the proportion of confidently segmented pixels returned at that threshold.
These plots intuitively describe both the `introspectiveness' and the `usefulness' of the model.
Note that the point on the curve at maximum \pac corresponds to the segmentation accuracy, as all pixels are treated as certain at this threshold, and so $\mathrm{max}\mathrm{[p(a, c)]} = \mathrm{p(accurate)}$.
% A benefit of both of these curves is that most of the evaluated models exhibit peaks in $\mathrm{F_{0.5}}$ and \Amd, which represent a natural place to select a threshold for deployment.

%------------------------------------------------------------------
\section{Experimental Results} \label{sec:results}
%------------------------------------------------------------------

This section presents test metrics and qualitative examples discussing the benefits of our approach in~\cref{sec:qual_res}.

% Questions to answer:
% - How much does the quality of uncertainty estimation degrade by when distributional shift increases (all methods, can look at AUC)
% - How does our method compare in general to the benchmarks for each domain? (report AUC)
% - How does our method compare in and around p(accuracy)=p(certain), i.e. at a specific point on the curve?
% - What do the accuracies looks like for each method?
% - Maybe also, what is the effect of our training method on performance on Cityscapes?	
% - Discussion on thresholds. We can learn it without the need for any more labelled data. Is it accurate? What can you do with Cityscapes Val, or SAX val datasets to calculate when p(accuracy) = p(certain)

% The primary experiment in this work, compares the AUC between different methods for each domain.
% AUC is a measure of the quality of the uncertainty estimates, and aggregates uncertainty accuracy over the range of possible uncertainty thresholds. 
% For each domain, the accuracy of uncertainty estimates, $\mathrm{A_{MD}}$ is reported for a range of uncertainty thresholds.

% Each tested method produces either uncertainty or confidences estimates, and these are thresholded to produce binary uncertainty maps.
% For each threshold and each image, both the accuracy of the uncertainty estimate, $\mathrm{A_{MD}}$, and proportion of certain pixels, $\mathrm{coverage}$, is calculated.

\begin{figure*}[h!]
\centering
\includegraphics[width=.8\linewidth]{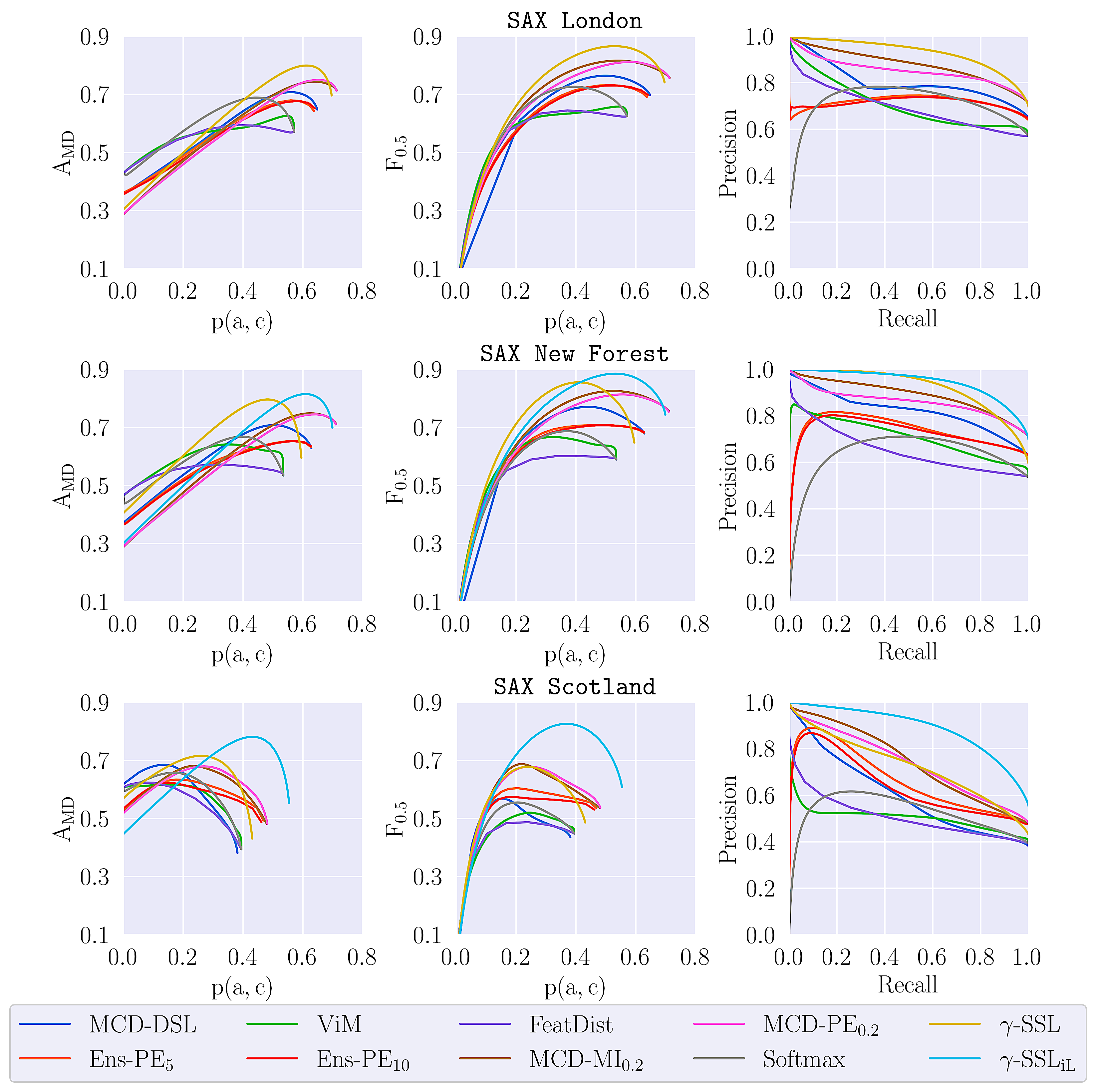}  
\caption{\label{fig:main_results_plot}
% A plot of the quality of uncertainty estimation, $\mathrm{A_{MD}}$, versus the proportion of pixels that are certain, $\mathrm{coverage}$, as judged by a threshold on estimated uncertainty. 
% $\mathrm{A_{MD}}$ is the proportion of pixels that are in either of the `safe' states: (1) accurate and certain or (2) inaccurate and uncertain.
% For a $\mathrm{coverage}=1$, all pixels are considered certain, and thus $\mathrm{A_{MD}}$ is just the segmentation accuracy. 
% As the $\mathrm{coverage}$ decreases, more pixels are considered uncertain, and if the uncertainty estimates are good, then the proportion of pixels in either of safe states should increase.
% A peak is observed for a few methods where the proportion of safe pixels it maximised at a certain $\mathrm{coverage}$.
For each SAX domain, a row of plots describes the misclassification detection performance of a series of benchmarks and the proposed methods, \gammassl and $\mathrm{\gamma}\text{-}\mathrm{SSL_{iL}}$. 
Misclassification detection accuracy, \Amd, and F-score, \fhalf, aggregate performance into a single metric, where a larger value of each represents a more `introspective' model.
They are plotted versus $\mathrm{p(a,c)}$, the proportion of pixels that are $\mathrm{accurate}$ and \certain, as this represents the amount of accurate and useful semantic information the model can extract from images; also a metric maximised by the ideal model.
Note that the maximum value of $\mathrm{p(a,c)}$ is equal to the segmentation accuracy, $\mathrm{max}\mathrm{[p(a, c)]} = \mathrm{p(accurate)}$.
Best viewed in color.
% They present how successful a model is at misclassification detection \emph{and} the quality of the semantic segmentations, thereby both how `introspective' and how 
}
\end{figure*}

% \subsection{SAX London Results}
\subsection{Source: Cityscapes, Target: SAX London}
For each plot in the first row of \cref{fig:main_results_plot}, \gammassl performs best.
Its precision is higher for nearly all values of recall, with a corresponding increase of \SI{19}{\%} and \SI{8}{\%} from the best benchmark in \auroc and \aupr respectively (see \cref{tab:auroc} and \cref{tab:aupr}).
It has the highest values of \maxAmd and \maxfhalf, returning them at \pac exceeded only by $\mathrm{MCD_{0.2}}$ (however at a lower value for \maxAmd and \maxfhalf).

The best-performing benchmarks are the $\mathrm{MCD_{0.2}}$ models, with \gls{mi} outperforming \gls{pe}.
This is because they are both more introspective as judged by \auroc and \aupr, and also have a high segmentation accuracy.
The latter suggests the dropout layers allowed for greater generalisation to the SAX domains, compared with a standard segmentation network.

On average, the performance for this domain was relatively similar for the epistemic and representation-based methods in terms of \auroc and \aupr.
However, while the representation-based methods exhibited similar \maxAmd and \maxfhalf values to epistemic methods, they do so at lower \pac on average; thus, overall, they perform less well.

Note that \gammasslil for the SAX London, KITTI and BDD domains does not exist in the results.
For SAX London \gammassl and \gammasslil are the same model, and for the latter two, it is not clear that SAX London is an intermediate domain for these target domains, see \cref{subsec:curriculum} for more details.

\begin{table}[h]
\caption{
Misclassification Detection AUROC with Source: Cityscapes
\label{tab:auroc}
}
\centering
\resizebox{0.8\columnwidth}{!}{
\begin{tabular}{ccccc>{\color{black}}c>{\color{black}}c}
\hline
& \multicolumn{1}{l}{} & \multicolumn{5}{c}{$\mathrm{AUROC}$}        \\
& Method               & LDN & NF & SCOT & KITTI & BDD \\ \hline
\multirow{7}{*}{\rotatebox[origin=c]{90}{Epistemic}} 
&  \enspefive & 0.630 & 0.645 & 0.629 & 0.845 & 0.758 \\
&  \ensmifive & 0.551 & 0.612 & 0.532 & 0.705 & 0.696 \\
& \enspeten & 0.603 & 0.629 & 0.608 & 0.828 & 0.755 \\
&  \ensmiten & 0.554 & 0.629 & 0.527 & 0.690 & 0.713 \\
&  \mcdpepointtwo & 0.727 & 0.739 & 0.735 & 0.864 & 0.801 \\
& \mcdmipointtwo & 0.755 & 0.774 & 0.725 & 0.815 & 0.801 \\
&  \mcddistil & 0.697 & 0.742 & 0.676 & 0.855 & 0.761 \\
 \hline
& -- Mean -- & 0.645 & 0.681 & 0.633 & 0.800 & 0.755 \\
  \hline
\multirow{5}{*}{\rotatebox[origin=c]{90}{Representation\ }}
& \softmax & 0.723 & 0.686 & 0.674 & 0.785 & 0.706 \\
& \softmaxA & 0.722 & 0.685 & 0.672 & 0.784 & 0.705 \\
& \featdist & 0.62 & 0.605 & 0.607 & 0.575 & 0.641 \\
&  \featdistA & 0.645 & 0.642 & 0.585 & 0.585 & 0.648 \\
&  \vim  & 0.635 & 0.700 & 0.640 & 0.714 & 0.745 \\
&  \dum & 0.483 & 0.565 & 0.510 & 0.501 & 0.502 \\
 \hline
& -- Mean -- & 0.638 & 0.647 & 0.615 & 0.657 & 0.658 \\
  \hline
 \multirow{2}{*}{\rotatebox[origin=c]{90}{Ours}} & $\mathrm{\gamma}\text{-}\mathrm{SSL}$     &  \textbf{0.895}  & 0.880   & 0.776  & \textbf{0.888} & \textbf{0.835} \\
& $\mathrm{\gamma}\text{-}\mathrm{SSL_{iL}}$   & - & \textbf{0.880} & \textbf{0.859} & - &  -\\
\hline
\end{tabular}
}
\end{table}

\begin{table}[h]
\caption{
Misclassification Detection AUPR with Source: Cityscapes
\label{tab:aupr}
}
\centering
\resizebox{0.8\columnwidth}{!}{
\begin{tabular}{ccccc>{\color{black}}c>{\color{black}}c}
\hline
& \multicolumn{1}{l}{} & \multicolumn{5}{c}{$\mathrm{AUPR}$}        \\
& Method               & LDN & NF & SCOT & KITTI & BDD \\ \hline
\multirow{7}{*}{\rotatebox[origin=c]{90}{Epistemic}} 
&  \enspefive & 0.714 & 0.733 & 0.663 & 0.956 & 0.854 \\
& \ensmifive & 0.663 & 0.737 & 0.570 & 0.910 & 0.830 \\
&  \enspeten & 0.710 & 0.724 & 0.634 & 0.952 & 0.857 \\
& \ensmiten & 0.667 & 0.752 & 0.554 & 0.905 & 0.840 \\
&  \mcdpepointtwo & 0.851 & 0.859 & 0.737 & 0.926 & 0.816 \\
&  \mcdmipointtwo & 0.878 & 0.889 & 0.744 & 0.956 & 0.904 \\
& \mcddistil & 0.785 & 0.818 & 0.602 & \textbf{0.957} & 0.851 \\
 \hline
& -- Mean -- & 0.753 & 0.787 & 0.643 & 0.937 &  0.850 \\
  \hline
\multirow{5}{*}{\rotatebox[origin=c]{90}{Representation\ }}
& \softmax & 0.711 & 0.631 & 0.525 & 0.840 & 0.686 \\
& \softmaxA & 0.711 & 0.631 & 0.524 & 0.840 & 0.686 \\
&  \featdist & 0.693 & 0.650 & 0.507 & 0.660 & 0.718 \\
& \featdistA  & 0.721 & 0.699 & 0.456 & 0.637 & 0.727 \\
& \vim & 0.705 & 0.708 & 0.502 & 0.800 & 0.797 \\
& \dum & 0.630 & 0.622 & 0.350 & 0.367 & 0.319 \\
 \hline
& -- Mean -- & 0.695 & 0.657 & 0.477 & 0.691 & 0.656 \\
  \hline
 \multirow{2}{*}{\rotatebox[origin=c]{90}{Ours}} & $\mathrm{\gamma}\text{-}\mathrm{SSL}$     &  \textbf{0.949}  &   0.921  & 0.726  & 0.951 & \textbf{0.911} \\
& $\mathrm{\gamma}\text{-}\mathrm{SSL_{iL}}$   & - & \textbf{0.942} & \textbf{0.887} & - & -\\
\hline
\end{tabular}
}
\end{table}

\begin{table}[h!]
\caption{
Maximum Accuracy $\mathrm{A_{MD}}$ and $\mathrm{p(a,c)}$ with Source: Cityscapes
\label{tab:max_A_MD}
}
\centering
\resizebox{\columnwidth}{!}{
\begin{tabular}{ccccc>{\color{black}}c>{\color{black}}c}
\hline
& \multicolumn{1}{l}{} & \multicolumn{5}{c}{$\mathrm{Max\mathrm{A_{MD}}}$ @ $\mathrm{p(a,c)}$}        \\
& Method               & LDN & NF & SCOT & KITTI & BDD \\ \hline
\multirow{7}{*}{\rotatebox[origin=c]{90}{Epistemic}} 
& \enspefive & 0.679 @ 0.560 & 0.652 @ 0.581 & 0.634 @ 0.175 & 0.818 @ 0.751 & 0.734 @ 0.607\\
& \ensmifive & 0.645 @ 0.620 & 0.643 @ 0.613 & 0.580 @ 0.076 & 0.798 @ 0.797 & 0.692 @ 0.668\\
& \enspeten & 0.678 @ 0.582 & 0.653 @ 0.564 & 0.622 @ 0.142 & 0.812 @ 0.768 & 0.728 @ 0.602\\
& \ensmiten & 0.654 @ 0.622 & 0.650 @ 0.611 & 0.578 @ 0.079 & 0.798 @ 0.798 & 0.698 @ 0.624\\
& \mcdpepointtwo & 0.750 @ 0.646 & 0.745 @ 0.645 & 0.679 @ 0.270 & 0.849 @ 0.792 & 0.765 @ 0.608\\
& \mcdmipointtwo & 0.744 @ 0.640 & 0.748 @ 0.625 & 0.681 @ 0.234 & 0.833 @ 0.823 & 0.754 @ 0.607\\
& \mcddistil & 0.708 @ 0.559 & 0.708 @ 0.503 & 0.685 @ 0.136 & 0.830 @ 0.745 & 0.738 @ 0.571\\
  \hline
& -- Mean -- & 0.694 @ 0.604 & 0.686 @ 0.592 & 0.637 @ 0.159 & 0.795 @ 0.782  & 0.730 @ 0.612 \\
  \hline
\multirow{5}{*}{\rotatebox[origin=c]{90}{Representation\ }} 
& \softmax & 0.689 @ 0.444 & 0.668 @ 0.396 & 0.657 @ 0.164 & 0.718 @ 0.434 & 0.692 @ 0.479\\
& \softmaxA & 0.689 @ 0.446 & 0.668 @ 0.401 & 0.656 @ 0.167 & 0.717 @ 0.435 & 0.692 @ 0.480\\
& \featdist & 0.594 @ 0.371 & 0.572 @ 0.334 & 0.624 @ 0.055 & 0.588 @ 0.574 & 0.620 @ 0.489\\
& \featdistA & 0.617 @ 0.449 & 0.603 @ 0.378 & 0.636 @ 0.032 & 0.564 @ 0.384 & 0.626 @ 0.502\\
& \vim & 0.626 @ 0.546 & 0.641 @ 0.353 & 0.617 @ 0.153 & 0.655 @ 0.386 & 0.687 @ 0.480\\
& \dum & 0.625 @ 0.625 & 0.618 @ 0.556 & 0.649 @ 0.000 & 0.733 @ 0.732 & 0.637 @ 0.635\\
 \hline
& -- Mean -- & 0.640 @ 0.480 & 0.628 @ 0.403 & 0.640 @ 0.095 & 0.663 @ 0.491 & 0.659 @ 0.511 \\
  \hline
% \multirow{8}{*}{\rotatebox[origin=c]{90}{Ablations}}
% & $\mathrm{NoSSL}$ & 0.727 @ 0.482 & 0.706 @ 0.424 & 0.680 @ 0.204 \\
% & $\mathrm{NoSAX}$ & 0.729 @ 0.465 & 0.717 @ 0.463 & 0.653 @ 0.003 \\
% & $\mathrm{M_{\gamma=-\infty}}$ & 0.754 @ 0.586 & 0.692 @ 0.426 & 0.685 @ 0.174 \\
% & $\mathrm{Sym}$-$\mathrm{Param}$ & 0.790 @ 0.253 & 0.596 @ 0.368 & 0.756 @ 0.053 \\
% & $\mathrm{Sym}$-$\mathrm{Non}$-$\mathrm{Param}$ & 0.752 @ 0.015 & 0.864 @ 0.000 & 0.902 @ 0.000 \\
% & $\mathrm{NoRegL}$ & 0.766 @ 0.552 & 0.748 @ 0.391 & 0.665 @ 0.201 \\
% & $\mathrm{MCD}$-$\mathrm{SSL}$ & 0.747 @ 0.594 & 0.748 @ 0.482 & 0.707 @ 0.153 \\
% & $\mathrm{p(c)}{=}{const.}$ & 0.782 @ 0.603 & 0.795 @ 0.466 & 0.758 @ 0.234 \\
% \hline
 \multirow{2}{*}{\rotatebox[origin=c]{90}{Ours}} & $\mathrm{\gamma}\text{-}\mathrm{SSL}$     &  \textbf{0.83} @ 0.625  &   0.796 @ 0.483    &   0.716 @ 0.260 & \textbf{0.856} @ 0.767 & \textbf{0.770} @ 0.568\\
& $\mathrm{\gamma}\text{-}\mathrm{SSL_{iL}}$   & - & \textbf{0.815} @ 0.608 & \textbf{0.781} @ 0.431 & - & -\\
\hline
\end{tabular}
}
\begin{tablenotes}
    \item  This represents the most conservative point of the system, given that the number of `safe' pixels is maximised.
\end{tablenotes}
\end{table}

\begin{table}[h!]
\caption{ Maximum F$_{0.5}$ Score and $\mathrm{p(a,c)}$ \label{tab:max_fhalf} with Source: Cityscapes}
\centering
\resizebox{\columnwidth}{!}{
% \begin{tabular}{ccccc>{\color{black}}c>{\color{black}}c}
\begin{tabular}{ccccc>{\color{black}}c>{\color{black}}c}
\hline
& \multicolumn{1}{l}{} & \multicolumn{5}{c}{$\mathrm{Max\mathrm{F_{0.5}}}$ @ $\mathrm{p(a,c)}$}        \\
& Method               & LDN & NF & SCOT & KITTI & BDD\\ \hline
\multirow{6}{*}{\rotatebox[origin=c]{90}{Epistemic}} 
& \enspefive & 0.732 @ 0.520 & 0.708 @ 0.460 & 0.604 @ 0.206 & 0.889 @ 0.613 & 0.801 @ 0.496\\
& \ensmifive & 0.694 @ 0.609 & 0.691 @ 0.605 & 0.528 @ 0.461 & 0.832 @ 0.795 & 0.758 @ 0.472\\
& \enspeten & 0.730 @ 0.539 & 0.707 @ 0.517 & 0.574 @ 0.168 & 0.880 @ 0.608 & 0.795 @ 0.492\\
& \ensmiten & 0.702 @ 0.603 & 0.696 @ 0.597 & 0.522 @ 0.460 & 0.833 @ 0.761 & 0.767 @ 0.488\\
& \mcdpepointtwo & 0.812 @ 0.584 & 0.813 @ 0.559 & 0.678 @ 0.248 & 0.914 @ 0.686 & 0.836 @ 0.530\\
& \mcdmipointtwo & 0.816 @ 0.541 & 0.825 @ 0.523 & 0.687 @ 0.215 & 0.891 @ 0.669 & 0.836 @ 0.500\\
& \mcddistil & 0.764 @ 0.499 & 0.770 @ 0.441 & 0.568 @ 0.155 & 0.900 @ 0.644 & 0.802 @ 0.491\\
  \hline
& -- Mean -- & 0.750 @ 0.556 & 0.744 @ 0.529 & 0.594 @ 0.273 & 0.877 @ 0.682 & 0.799 @ 0.496\\
  \hline
\multirow{5}{*}{\rotatebox[origin=c]{90}{Representation\ }} 
& \softmax & 0.726 @ 0.395 & 0.687 @ 0.370 & 0.555 @ 0.205 & 0.777 @ 0.366 & 0.735 @ 0.429\\
& \softmaxA  & 0.726 @ 0.397 & 0.687 @ 0.373 & 0.554 @ 0.207 & 0.777 @ 0.366 & 0.735 @ 0.430\\
& \featdist & 0.645 @ 0.371 & 0.601 @ 0.408 & 0.487 @ 0.240 & 0.640 @ 0.571 & 0.672 @ 0.434\\
& \featdistA & 0.672 @ 0.360 & 0.643 @ 0.296 & 0.449 @ 0.240 & 0.612 @ 0.525 & 0.680 @ 0.428\\
& \vim & 0.658 @ 0.543 & 0.667 @ 0.326 & 0.520 @ 0.243 & 0.721 @ 0.324 & 0.735 @ 0.391\\
& \dum & 0.676 @ \textbf{0.625} & 0.664 @ 0.528 & 0.419 @ 0.315 & 0.774 @ 0.731 & 0.686 @ 0.635\\
 \hline
& -- Mean -- & 0.684 @ 0.449 & 0.658 @ 0.384 & 0.497 @ 0.242 & 0.717 @ 0.481 & 0.707 @ 0.458 \\
%   \hline
% \multirow{6}{*}{\rotatebox[origin=c]{90}{Ablations}}
% & $\mathrm{NoSSL}$ & 0.790 @ 0.380 & 0.753 @ 0.364 & 0.626 @ 0.203 \\
% & $\mathrm{NoSAX}$ & 0.769 @ 0.401 & 0.756 @ 0.403 & 0.505 @ 0.220 \\
% & $\mathrm{M_{\gamma=-\infty}}$ & 0.826 @ 0.492 & 0.746 @ 0.353 & 0.602 @ 0.184 \\
% & $\mathrm{Sym}$-$\mathrm{Param}$ & 0.731 @ 0.219 & 0.608 @ 0.381 & 0.495 @ 0.057 \\
% & $\mathrm{Sym}$-$\mathrm{Non}$-$\mathrm{Param}$ & 0.394 @ 0.088 & 0.281 @ 0.030 & 0.210 @ 0.017 \\
% & $\mathrm{NoRegL}$ & 0.806 @ 0.507 & 0.781 @ 0.332 & 0.569 @ 0.241 \\
% & $\mathrm{MCD}$-$\mathrm{SSL}$ & 0.809 @ 0.518 & 0.804 @ 0.416 & 0.589 @ 0.159 \\
% & $\mathrm{p(c)}{=}{const.}$ & 0.853 @ 0.513 & 0.851 @ 0.396 & 0.706 @ 0.213 \\
\hline
 \multirow{2}{*}{\rotatebox[origin=c]{90}{Ours}} & $\mathrm{\gamma}\text{-}\mathrm{SSL}$     &  \textbf{0.893} @ 0.548  &   0.855 @ 0.407    &   0.678 @ 0.239 & \textbf{0.920} @ 0.676 & \textbf{0.843} @ 0.475\\
 & $\mathrm{\gamma}\text{-}\mathrm{SSL_{iL}}$ &  -  & \textbf{0.885} @ 0.532  & \textbf{0.826} @ 0.370 & - & -\\
\hline
\end{tabular}
}
\end{table}

% \subsubsection*{$\mathrm{AUROC}$ \& $\mathrm{AUPR}$}
% In terms of \auroc, MCD-MI$_{0.2}$, Softmax and MCD-PE$_{0.2}$ performed significantly better than the other benchmarks on this test dataset.
% As for \aupr, both MCD$-{0.2}$ methods significantly performed the rest of the benchmarks. 
% SAX London is the domain for which the two types of methods are most comparable in performance (0.637 vs. 0.638 and 0.747 vs. 0.695 for epistemic and representation methods respectively, see \cref{tab:auroc-aupr}), due to the significant decrease in performance of the representation methods as discussed above. 

% For \gammassl, the best absolute performance was achieved on the SAX London domain in terms of absolute \auroc and \aupr but also relative to the benchmarks as the percentage increase from MCD-MI$_{0.2}$ was \SI{14.3}{\%} and \SI{5.7}{\%} in \auroc and \aupr respectively.

% Each of MCD-MI$_{0.2}$, Softmax and MCD-PE$_{0.2}$ performed significantly better than the other benchmarks on this test dataset, and for both of \auroc and \aupr, the best performing benchmark on this test dataset was MCD-MI$_{0.2}$, with Softmax for \auc.
% This means that both the MCD$_{0.2}$ models performed significantly better than the other epistemic methods, and more importantly, so did the Softmax model. 
% This supports the evidence in \cite{maxsoftmax}, that a network trained with a supervised cross-entropy loss function is a strong benchmark for \gls{ood}, or as shown here, misclassification detection.

% \subsection{SAX New Forest Results}
\subsection{Source: Cityscapes, Target: SAX New Forest}
In SAX New Forest, \gammasslil and \gammassl perform similarly for \auroc and \aupr with \gammasslil having slightly higher \aupr (\SI{0.942}{} vs. \SI{0.921}{}).
% However \gammasslil performs significantly better than \gammassl in terms of \Amd and \fhalf versus \pac.
The increases from \gammassl to \gammasslil for \maxAmd and \maxfhalf are modest at \SI{2.4}{\%} and \SI{3.5}{\%} respectively, however the increases in \pac at which they occur are far larger at \SI{25.9}{\%} and \SI{30.7}{\%}.
This coincides with a large increase in segmentation accuracy from \gammassl to \gammasslil, as seen from the maximum \pac in \cref{fig:main_results_plot}.
For this domain, this confirms the hypothesis that presenting unlabelled target images in a curriculum improves semi-supervised learning both in segmentation quality and uncertainty estimation.

Compared to $\mathrm{MCD_{0.2}}$, \gammassl has higher \auroc and \aupr; however, the better method is context-dependent.
The increase in \maxAmd and \maxfhalf scores from $\mathrm{MCD_{0.2}}$ to \gammassl  are \SI{6.4}{\%} and \SI{3.64}{\%} respectively but occur at a \pac \SI{22.7}{\%} and \SI{22.2}{\%} lower. %, therefore the specifics of the robotics context will dictate which is the more ideal.
This is a possible example of a model with higher segmentation accuracy (i.e. higher \pac) preferable to a more introspective model (i.e. higher \auroc, \aupr); this demonstrates the benefit of presenting results in this way.
This is not true for \gammasslil, with an increase in \maxAmd and \maxfhalf scores of \SI{9.0}{\%} and \SI{7.3}{\%} at a change in \pac of \SI{-2.7}{\%} and \SI{1.7}{\%} compared to \mcdmipointtwo.

As in SAX London, the $\mathrm{MCD_{0.2}}$ models were the best-performing benchmarks on each metric.
In general, the epistemic methods outperformed the representation-based methods, with higher mean \auroc and \aupr.
While the mean values of \maxAmd and \maxfhalf were not significantly different, the values of \pac at which they occur were higher for epistemic methods than for representation methods.

% In SAX New Forest, the performing benchmark was again MCD-MI$_{0.2}$, in terms of both \auroc and \aupr.
% In this domain, ViM performs well compared to all epistemic methods apart from MCD$_{0.2}$, and all other representation methods, which is impressive given it has the same representation as Softmax, FeatDist.
% \gammassl still outperforms the MCD-MI$_{0.2}$, however in this case by a smaller margin than in SAX London, with percentage increases of \SI{10.2}{\%} and \SI{0.1}{\%} for \auroc and \aupr respectively.

% \subsection{SAX Scotland Results}
\subsection{Source: Cityscapes, Target: SAX Scotland}
In this domain, the increase in performance from \gammassl to \gammasslil is at its greatest, and the \gammasslil model far exceeds the performance of the other models, as in \cref{fig:main_results_plot}.
The increase over the next best model for metrics \auroc, \aupr, $\mathrm{MaxA_{MD}}\text{@}\mathrm{p(a,c)}$, $\mathrm{MaxF_{0.5}}\text{@}\mathrm{p(a,c)}$ are as follows: \SI{10.7}{\%}, \SI{19.2}{\%}, \SI{9.1}{\%} @ \SI{65.8}{\%}, \SI{20.2}{\%} @ \SI{72.1}{\%}.
Once again, the performance increase in segmentation quality \emph{and} uncertainty estimation can be attributed to the curriculum training procedure described in \cref{subsec:curriculum}.

\gammassl performs comparably to $\mathrm{MCD_{0.2}}$ in this domain, which as the distributional shift between SAX Scotland and Cityscapes so significant that it is challenging to use the semi-supervised task to improve the model's representation.

On average, epistemic methods outperform representation-based ones to a larger extent in this domain, characterised by a mean increase in \aupr of \SI{34.8}{\%} from the latter to the former.
This is because representation methods rely on a representation learned from Cityscapes, which is significantly distributionally shifted from SAX Scotland.

\begin{table}[]
\caption{
Misclassification Detection AUROC with Source: BDD
\label{tab:auroc_bdd}
}
\centering
\resizebox{0.7\columnwidth}{!}{
\begin{tabular}{ccccc}
\hline
& \multicolumn{1}{l}{} & \multicolumn{3}{c}{$\mathrm{AUROC}$}        \\
& Method               & LDN & NF & SCOT \\ \hline
\multirow{7}{*}{\rotatebox[origin=c]{90}{Epistemic}} 
&  \enspefive & 0.781 & 0.854 & 0.862  \\
&  \ensmifive & 0.744 & 0.809 & 0.818  \\
& \enspeten & 0.785 & 0.854 & \textbf{0.866} \\
&  \ensmiten & 0.761 & 0.827 & 0.861 \\
&  \mcdpepointtwo  & 0.834 & 0.841 & 0.822 \\
& \mcdmipointtwo & 0.808 & 0.816 & 0.739  \\
&  \mcddistil & 0.821 & 0.803 & 0.805  \\
 \hline
& -- Mean -- & 0.791  & 0.829 & 0.825 \\
  \hline
\multirow{5}{*}{\rotatebox[origin=c]{90}{Representation\ }}
& \softmax & 0.825 & 0.819 & 0.810  \\
& \softmaxA & 0.825 & 0.819 & 0.810  \\
& \featdist & 0.605 & 0.599 & 0.563 \\
&  \featdistA & 0.625 & 0.631 & 0.598  \\
&  \vim  & 0.663 & 0.664 & 0.560  \\
&  \dum & 0.480 & 0.431 & 0.398 \\
 \hline
& -- Mean -- & 0.671 & 0.661 & 0.623 \\
  \hline
 \multirow{2}{*}{\rotatebox[origin=c]{90}{Ours}} & $\mathrm{\gamma}\text{-}\mathrm{SSL}$     &  \textbf{0.889}  & 0.876   & 0.833  \\
& $\mathrm{\gamma}\text{-}\mathrm{SSL_{iL}}$   & - & \textbf{0.882} & 0.855 \\
\hline
\end{tabular}

}
\end{table}

\begin{table}[]
\caption{
Misclassification Detection AUPR with Source: BDD
\label{tab:aupr_bdd}
}
\centering
\resizebox{0.7\columnwidth}{!}{
\begin{tabular}{ccccc}
\hline
& \multicolumn{1}{l}{} & \multicolumn{3}{c}{$\mathrm{AUPR}$}        \\
& Method               & LDN & NF & SCOT \\ \hline
\multirow{7}{*}{\rotatebox[origin=c]{90}{Epistemic}} 
&  \enspefive & 0.896 & 0.941 & 0.885  \\
&  \ensmifive & 0.876 & 0.915 & 0.837 \\
& \enspeten & 0.899 & \textbf{0.942} & 0.880 \\
&  \ensmiten & 0.884 & 0.925 & 0.870 \\
&  \mcdpepointtwo & 0.903 & 0.881 & 0.798  \\
& \mcdmipointtwo & 0.880 & 0.863 & 0.728  \\
&  \mcddistil & 0.902 & 0.914 & 0.849  \\
 \hline
& -- Mean -- & 0.891 & 0.912 & 0.835 \\
  \hline
\multirow{5}{*}{\rotatebox[origin=c]{90}{Representation\ }}
    & \softmax & 0.901 & 0.904 & 0.816  \\
& \softmaxA & 0.901 & 0.905 & 0.816  \\
& \featdist & 0.724 & 0.731 & 0.547  \\
&  \featdistA & 0.758 & 0.770 & 0.604  \\
&  \vim  & 0.799 & 0.797 & 0.547  \\
&  \dum & 0.549 & 0.530 & 0.321 \\
 \hline
& -- Mean -- & 0.772 & 0.773 & 0.608 \\
  \hline
 \multirow{2}{*}{\rotatebox[origin=c]{90}{Ours}} & $\mathrm{\gamma}\text{-}\mathrm{SSL}$     &  \textbf{0.936}  & 0.935   & 0.858  \\
& $\mathrm{\gamma}\text{-}\mathrm{SSL_{iL}}$   & - & 0.928 & \textbf{0.900} \\
\hline
\end{tabular}
}
\end{table}

\begin{table}[]
\caption{
Maximum Accuracy $\mathrm{A_{MD}}$ and $\mathrm{p(a,c)}$ with Source: BDD
\label{tab:max_A_MD_bdd}
}
\centering
\resizebox{\columnwidth}{!}{
\begin{tabular}{ccccc}
\hline
& \multicolumn{1}{l}{} & \multicolumn{3}{c}{$\mathrm{Max\mathrm{A_{MD}}}$ @ $\mathrm{p(a,c)}$}        \\
& Method               & LDN & NF & SCOT  \\ \hline
\multirow{7}{*}{\rotatebox[origin=c]{90}{Epistemic}} 
& \enspefive & 0.739 @ 0.630 & 0.783 @ 0.621 & 0.790 @ 0.347\\
& \ensmifive & 0.716 @ 0.623 & 0.766 @ 0.637 & 0.753 @ 0.343 \\
& \enspeten & 0.741 @ 0.619 & 0.783 @ 0.627 & 0.792 @ 0.331 \\
& \ensmiten & 0.727 @ 0.619 & 0.774 @ 0.629 & \textbf{0.797} @ 0.351\\
& \mcdpepointtwo & 0.753 @ 0.488 & 0.760 @ 0.390 & 0.766 @ 0.217 \\
& \mcdmipointtwo & 0.729 @ 0.454 & 0.742 @ 0.362 & 0.729 @ 0.176 \\
& \mcddistil & 0.756 @ 0.549 & 0.761 @ 0.638 & 0.727 @ 0.387 \\
  \hline
& -- Mean -- & 0.737 @ 0.569 &  0.767 @ 0.558 &  0.765 @ 0.307 \\
  \hline
\multirow{5}{*}{\rotatebox[origin=c]{90}{Representation\ }} 
& \softmax & 0.761 @ 0.553 & 0.760 @ 0.573 & 0.755 @ 0.275 \\
& \softmaxA & 0.761 @ 0.554 & 0.761 @ 0.574 & 0.754 @ 0.275 \\
& \featdist & 0.673 @ 0.609 & 0.677 @ 0.674 & 0.603 @ 0.154 \\
& \featdistA & 0.706 @ 0.646 & 0.727 @ 0.668 & 0.590 @ 0.265 \\
& \vim & 0.662 @ 0.662 & 0.683 @ 0.608 & 0.588 @ 0.114 \\
& \dum & 0.562 @ 0.562 & 0.596 @ 0.596 & 0.604 @ 0.000 \\
 \hline
& -- Mean -- & 0.688 @ 0.598 & 0.701 @ 0.616 & 0.649 @ 0.181 \\
  \hline
% \multirow{8}{*}{\rotatebox[origin=c]{90}{Ablations}}
% & $\mathrm{NoSSL}$ & 0.727 @ 0.482 & 0.706 @ 0.424 & 0.680 @ 0.204 \\
% & $\mathrm{NoSAX}$ & 0.729 @ 0.465 & 0.717 @ 0.463 & 0.653 @ 0.003 \\
% & $\mathrm{M_{\gamma=-\infty}}$ & 0.754 @ 0.586 & 0.692 @ 0.426 & 0.685 @ 0.174 \\
% & $\mathrm{Sym}$-$\mathrm{Param}$ & 0.790 @ 0.253 & 0.596 @ 0.368 & 0.756 @ 0.053 \\
% & $\mathrm{Sym}$-$\mathrm{Non}$-$\mathrm{Param}$ & 0.752 @ 0.015 & 0.864 @ 0.000 & 0.902 @ 0.000 \\
% & $\mathrm{NoRegL}$ & 0.766 @ 0.552 & 0.748 @ 0.391 & 0.665 @ 0.201 \\
% & $\mathrm{MCD}$-$\mathrm{SSL}$ & 0.747 @ 0.594 & 0.748 @ 0.482 & 0.707 @ 0.153 \\
% & $\mathrm{p(c)}{=}{const.}$ & 0.782 @ 0.603 & 0.795 @ 0.466 & 0.758 @ 0.234 \\
% \hline
 \multirow{2}{*}{\rotatebox[origin=c]{90}{Ours}} & $\mathrm{\gamma}\text{-}\mathrm{SSL}$     &  \textbf{0.818} @ 0.599  &   0.805 @ 0.567    &   0.762 @ 0.331 \\
& $\mathrm{\gamma}\text{-}\mathrm{SSL_{iL}}$   & - & \textbf{0.809} @ 0.578 & 0.769 @ 0.417 \\
\hline
\end{tabular}
}
\begin{tablenotes}
    \item  This represents the most conservative point of the system, given that the number of `safe' pixels is maximised.
\end{tablenotes}
\end{table}

\begin{table}[]
\caption{ Maximum F$_{0.5}$ Score and $\mathrm{p(a,c)}$ \label{tab:max_fhalf_bdd} with Source: BDD}
\centering
\resizebox{\columnwidth}{!}{
\begin{tabular}{ccccc}
\hline
& \multicolumn{1}{l}{} & \multicolumn{3}{c}{$\mathrm{Max\mathrm{F_{0.5}}}$ @ $\mathrm{p(a,c)}$}        \\
& Method               & LDN & NF & SCOT \\ \hline
\multirow{6}{*}{\rotatebox[origin=c]{90}{Epistemic}} 
& \enspefive & 0.810 @ 0.488 & 0.872 @ 0.500 & 0.829 @ 0.307 \\
& \ensmifive & 0.792 @ 0.484 & 0.839 @ 0.528 & 0.777 @ 0.290 \\
& \enspeten & 0.814 @ 0.488 & 0.872 @ 0.500 & 0.825 @ 0.284 \\
& \ensmiten & 0.802 @ 0.483 & 0.855 @ 0.519 & 0.815 @ 0.309 \\
& \mcdpepointtwo & 0.833 @ 0.403 & 0.814 @ 0.328 & 0.748 @ 0.199 \\
& \mcdmipointtwo & 0.820 @ 0.380 & 0.803 @ 0.311 & 0.685 @ 0.163 \\
& \mcddistil & 0.829 @ 0.460 & 0.839 @ 0.525 & 0.769 @ 0.319 \\
  \hline
& -- Mean -- & 0.814 @ 0.455 & 0.842 @ 0.459 & 0.778 @ 0.267 \\
  \hline
\multirow{5}{*}{\rotatebox[origin=c]{90}{Representation\ }} 
& \softmax & 0.833 @ 0.465 & 0.835 @ 0.483 & 0.764 @ 0.241 \\
& \softmaxA  & 0.833 @ 0.468 & 0.836 @ 0.484 & 0.764 @ 0.241 \\
& \featdist & 0.730 @ 0.564 & 0.731 @ 0.566 & 0.534 @ 0.191 \\
& \featdistA & 0.755 @ 0.614 & 0.778 @ 0.621 & 0.592 @ 0.285 \\
& \vim & 0.732 @ 0.433 & 0.747 @ 0.541 & 0.511 @ 0.214 \\
& \dum & 0.616 @ 0.562 & 0.648 @ 0.596 & 0.451 @ 0.396 \\
 \hline
& -- Mean -- & 0.750 @ 0.518 & 0.763 @ 0.549 & 0.603 @ 0.261 \\
%   \hline
% \multirow{6}{*}{\rotatebox[origin=c]{90}{Ablations}}
% & $\mathrm{NoSSL}$ & 0.790 @ 0.380 & 0.753 @ 0.364 & 0.626 @ 0.203 \\
% & $\mathrm{NoSAX}$ & 0.769 @ 0.401 & 0.756 @ 0.403 & 0.505 @ 0.220 \\
% & $\mathrm{M_{\gamma=-\infty}}$ & 0.826 @ 0.492 & 0.746 @ 0.353 & 0.602 @ 0.184 \\
% & $\mathrm{Sym}$-$\mathrm{Param}$ & 0.731 @ 0.219 & 0.608 @ 0.381 & 0.495 @ 0.057 \\
% & $\mathrm{Sym}$-$\mathrm{Non}$-$\mathrm{Param}$ & 0.394 @ 0.088 & 0.281 @ 0.030 & 0.210 @ 0.017 \\
% & $\mathrm{NoRegL}$ & 0.806 @ 0.507 & 0.781 @ 0.332 & 0.569 @ 0.241 \\
% & $\mathrm{MCD}$-$\mathrm{SSL}$ & 0.809 @ 0.518 & 0.804 @ 0.416 & 0.589 @ 0.159 \\
% & $\mathrm{p(c)}{=}{const.}$ & 0.853 @ 0.513 & 0.851 @ 0.396 & 0.706 @ 0.213 \\
\hline
 \multirow{2}{*}{\rotatebox[origin=c]{90}{Ours}} & $\mathrm{\gamma}\text{-}\mathrm{SSL}$     &  \textbf{0.887} @ 0.515  &   0.873 @ 0.495    &   0.797 @ 0.284 \\
 & $\mathrm{\gamma}\text{-}\mathrm{SSL_{iL}}$ &  -  & \textbf{0.885} @ 0.501  & \textbf{0.831} @ 0.348 \\
\hline
\end{tabular}
}
\end{table}

\subsection{Effect of increasing distributional shift}
As discussed in \cref{subsec:data}, the following domains are in order of ascending distributional shift: London, New Forest, Scotland, as evidenced by the corresponding reduction in segmentation accuracy -- $0.571$, $0.538$, $0.394$ -- for a network trained solely on Cityscapes. 
% \Cref{tab:domain_shift} shows that the extent to which misclassification detection performance degrades as distributional shift increases appears to depend on what type of method is used.
\Cref{tab:domain_shift} shows that the method type has an effect on the extent to which misclassification detection performance degrades as distributional shift increases.

\subsubsection*{Epistemic Methods}
For all epistemic methods, \auroc, \aupr increases from London to New Forest -- \SI{6.0}{\%} and \SI{5.0}{\%} on average, respectively.
This suggests that these techniques perform better uncertainty estimation as the proportion of errors related to distributional shift increases, which aligns with the stated motivation of the methods. 
However, as distributional uncertainty significantly increases, not every method improves further.
In fact, both \auroc and \aupr significantly decrease from New Forest to Scotland, \SI{-7.4}{\%, \SI{-18.3}{\%}} on average.
% and there were only two small increases for Ens-MI$_5$ and MCD-PE$_{0.2}$ on \auc.
This suggests that there is a limit beyond which these methods start to degrade significantly, as was also reported for a range of epistemic uncertainty estimation methods in \cite{bad_uncertainty}.

\subsubsection*{Representation methods}
% Relative to the epistemic methods, these methods decrease more rapidly in performance as distributional shift increases, with an average change of \SI{-6.7}{\%} and \SI{-5.4}{}
In general, these methods are less robust to distributional shift than the epistemic ones, with a difference in \aupr of \SI{-5.4}{\%} and \SI{-27.3}{\%} for London to New Forest, and New Forest to Scotland respectively. 
The changes were less uniform for \auroc; however, the majority of models tested decreased in performance for \textit{both} shifts.
%, unlike the epistemic methods were there was a uniform increase from London to New Forest.
Additionally, these methods have a significant reduction in the \pac at which \maxAmd and \maxfhalf occur compared with epistemic methods.
% The decrease in segmentation accuracy means the distributional shift causes the networks to represent the known classes less separably, and these results suggest the same is true for in-distribution vs. \gls{ood} data.

\subsubsection*{$\gamma$-SSL methods}
% \subsubsection*{$\mathrm{\gamma}\text{-}\mathrm{SSL}$ methods}
Much like many of the representation methods, the performance of \gammassl decreases from London to New Forest, and again to Scotland.
\gammassl decreases less on average in terms of \aupr, but more in terms of \auroc.
Independent of the increase in segmentation accuracy, the use of a curriculum for \gammasslil reduces the effects of large distributional shifts on uncertainty estimation, as evidenced by the \gammasslil having by far the smallest reduction in \aupr from New Forest to Scotland. 
Factoring in the increased segmentation accuracy, \gammasslil also exhibits a much smaller reduction in the value of \pac at which \maxAmd and \maxfhalf occur when compared to \gammassl.

% TODO: needs comment about MaxAue and Coverage -> and needs comparison to optimal model for this 
% For both types of methods, the $\mathbm{coverage}$ at which \Amd is maximised decreases significantly from New Forest to Scotland. 
% This means that the point at which most pixels are correctly assigned

\begin{table}[]
\caption{\auroc and \aupr Percentage Change at Increasing Distributional Shift, $\% \Delta \mathrm{ROC}$ and $\% \Delta \mathrm{PR}$ Respectively}
\label{tab:domain_shift}
\centering
\begin{tabular}{cccccc}
\hline
& \multicolumn{1}{l}{} & \multicolumn{2}{c}{LDN $\rightarrow$ NF}  & \multicolumn{2}{c}{NF $\rightarrow$ SCOT}  \\
& Method               & $\% \Delta \mathrm{ROC}$ & $\% \Delta \mathrm{PR}$ & $\% \Delta \mathrm{ROC}$ &  $\% \Delta \mathrm{PR}$ \\ 
\hline
\multirow{6}{*}{\rotatebox[origin=c]{90}{Epistemic}} 
 & \enspefive & 2.3 & 2.7 & -2.5 & -9.6 \\
 & \ensmifive & 11.1 & 11.2 & -13.0 & -22.7 \\
 & \enspeten & 4.2 & 2.0 & -3.2 & -12.4 \\
 & \ensmiten & 13.7 & 12.8 & -16.3 & -26.3 \\
 & \mcdpepointtwo & 1.8 & 0.9 & -0.6 & -14.2 \\
 & \mcdmipointtwo & 2.6 & 1.3 & -6.4 & -16.4 \\
 & \mcddistil & 6.5 & 4.2 & -8.9 & -26.4 \\
 \hline
 & -- Mean -- & 6.0  & 5.0 & -7.3 & -18.3 \\
 \hline
\multirow{6}{*}{\rotatebox[origin=c]{90}{Representation}} 
 & \softmax & -5.1 & -11.2 & -1.8 & -16.8 \\
 & \softmaxA & -5.1 & -11.2 & -1.9 & -17.0 \\
 & \featdist & -2.3 & -6.1 & 0.3 & -22.1 \\
 & \featdistA & -0.4 & -3.1 & -9.0 & -34.8 \\
 & \vim & 10.3 & 0.5 & -8.5 & -29.2 \\
 & \dum & 16.9 & -1.4 & -9.7 & -43.6 \\
 \hline
 & -- Mean --  & 2.4 & -5.4 & -5.1 & -27.3 \\
 \hline
 \multirow{2}{*}{\rotatebox[origin=c]{90}{Ours}} & $\mathrm{\gamma}\text{-}\mathrm{SSL}$     & -1.7 & -3.0 & -11.8 & -21.2 \\
 % & $\mathrm{\gamma}\text{-}\mathrm{SSL}\text{-}\mathrm{iL}$     & -1.2 & -4.0 & -7.3 & -20.0 \\
 & $\mathrm{\gamma}\text{-}\mathrm{SSL_{iL}}$     & -1.7 & -0.7 & -2.4 & -5.8 \\
%  \multirow{6}{*}{\rotatebox[origin=c]{90}{Ablations\ }} 
% & $\mathrm{NoSSL}$ & 0.785 & - & - & 0.837 & - & - \\
% & $\mathrm{M_{\gamma=-\infty}}$ & 0.836 & - & - & 0.896 & - & - \\
% & $\mathrm{NoSAX}$ & 0.841 & - & - & 0.886 & - & - \\
% & $\mathrm{Sym}$-$\mathrm{NoP}$ & 0.675 & - & - & 0.386 & - & - \\
% & $\mathrm{Sym}$-$\mathrm{P}$ & 0.736 & - & - & 0.578 & - & - \\
% & $\mathrm{NoRegL}$ & 0.793 & - & - & 0.766 & - & - \\
% \hline
%  \rotatebox[origin=c]{90}{\ Ours\ } & $\gamma$-SSL &  \textbf{0.882}  &  \textbf{0.849}   &  & \textbf{0.936} & \textbf{0.877} &    \\
\hline
\end{tabular}
\begin{tablenotes}
   \item \textit{A negative value represents a decrease in misclassification detection performance as distributional shift increases.}
\end{tablenotes}
\end{table}

\subsection{Source: Cityscapes, Target: KITTI \& BDD}   \label{subsec:kitti_bdd_experiments}
The results for KITTI and BDD can be found in \cref{tab:auroc}, \cref{tab:aupr}, \cref{tab:max_A_MD}, \cref{tab:max_fhalf}.
The accuracy of \gammassl for KITTI, SAX London, BDD and SAX New Forest are as follows: 0.817, 0.703, 0.684, 0.595, therefore KITTI is the least distributionally shifted domain w.r.t. Cityscapes, and BDD is more distributionally shifted than SAX London.

\subsubsection*{KITTI}
The \auroc, \maxAmd and \maxfhalf metrics for our \gammassl method exceeded that of all of the benchmarks, with only \aupr exceeded by the \mcddistil method.
Epistemic methods significantly outperformed representation-based methods, as epistemic method outperformed the latter on almost all of the metrics.
% The higher mean values of \pac reported for each method suggest that KITTI is more similar to Cityscapes than any of the SAX domains.
This experiment therefore provides extra evidence that our method performs well in target domains close to the source domain.
\subsubsection*{BDD}
\gammassl is the best performing model for each metric with BDD as the target domain, with epistemic methods on average outperforming the representation-based methods.
Despite being less distributionally shifted than SAX New Forest according to segmentation accuracy, each of the uncertainty estimation metrics are lower for BDD than SAX New Forest in a break from the trend.
The hypothesis for why this is that there is significantly more diversity in BDD compared with SAX New Forest, and learning a representation for uncertainty estimation is more difficult in a more diverse domain.

\subsection{Source: BDD}    \label{subsec:source_bdd}
In order to investigate the generality of this approach, this work also conducts secondary experiments with \gls{bdd} as the source domain.
The suite of results for this are found: \cref{tab:auroc_bdd}, \cref{tab:aupr_bdd}, \cref{tab:max_A_MD_bdd}, \cref{tab:max_fhalf_bdd}.
For both epistemic and representation based methods, using BDD as the source domain improves uncertainty estimation performance across all metrics.
This benefit typically increases as the domain shift increases, e.g. for \maxfhalf the average percentage increases from Cityscapes to BDD for the benchmarks are: 9.1\%, 14.6\%, 26.2\% for SAX London, New Forest and Scotland respectively.
The BDD labelled dataset is larger and more diverse, therefore leading to a more general representation of the semantic classes.
\subsubsection*{SAX London}
For this target domain, \gammassl performs the best on each metric, with a \SI{6.5}{\%} increase in \maxfhalf over the best performing benchmark.
Across the board, the results for \gammassl with Cityscapes as the source domain are better than in this experiment.
% It is possible that if BDD is more diverse, then a narrower definition of the source domain helps to improve quality of uncertainty estimation
The higher uncertainty estimation metrics coincide with a higher accuracies using Cityscapes as the source domain than BDD (0.703 and 0.688 respectively), suggesting that Cityscapes could be more similar to this domain than BDD.
\subsubsection*{SAX New Forest}
The accuracy of the \gammassl models are 0.666 and 0.595 for BDD and Cityscapes respectively, suggesting a larger domain shift between Cityscapes and SAX New Forest, than for BDD and SAX New Forest.
In all but \aupr, both the \gammassl and \gammasslil models outperform all benchmarks.
The results for using BDD as source are similar or slightly better than using Cityscapes.
Better uncertainty estimation results for BDD would be predicted by the lesser domain shift, however this is not consistently shown.
This is perhaps because BDD is more diverse, and a narrower definition of the source domain allows a greater separation of source and target, and simpler uncertainty estimation.
\subsubsection*{SAX Scotland}
The results for \gammassl are significantly better with source domain as BDD, than as Cityscapes, accompanied by similar but smaller improvements for \gammasslil.
Segmentation accuracies for \gammassl of 0.495 and 0.431 for BDD and Cityscapes respectively suggest that the magnitude of the domain shifts would explain this.
It is however also true in this experiment that the epistemic methods are significantly better resulting in comparable performance between \gammasslil and these methods for these metrics, i.e. which method is better depends on the metric considered.

The results demonstrate that while a different source domain has an effect on the quality of uncertainty estimation, our method still exceeds or is competitive with the results of the best benchmarks considered, while still having the low latency required for robotic deployment.

\subsection{Optimal Threshold Calculation Testing} \label{subsec:threshold_tests}
\begin{figure}[h!]
\centering
\includegraphics[width=\columnwidth]{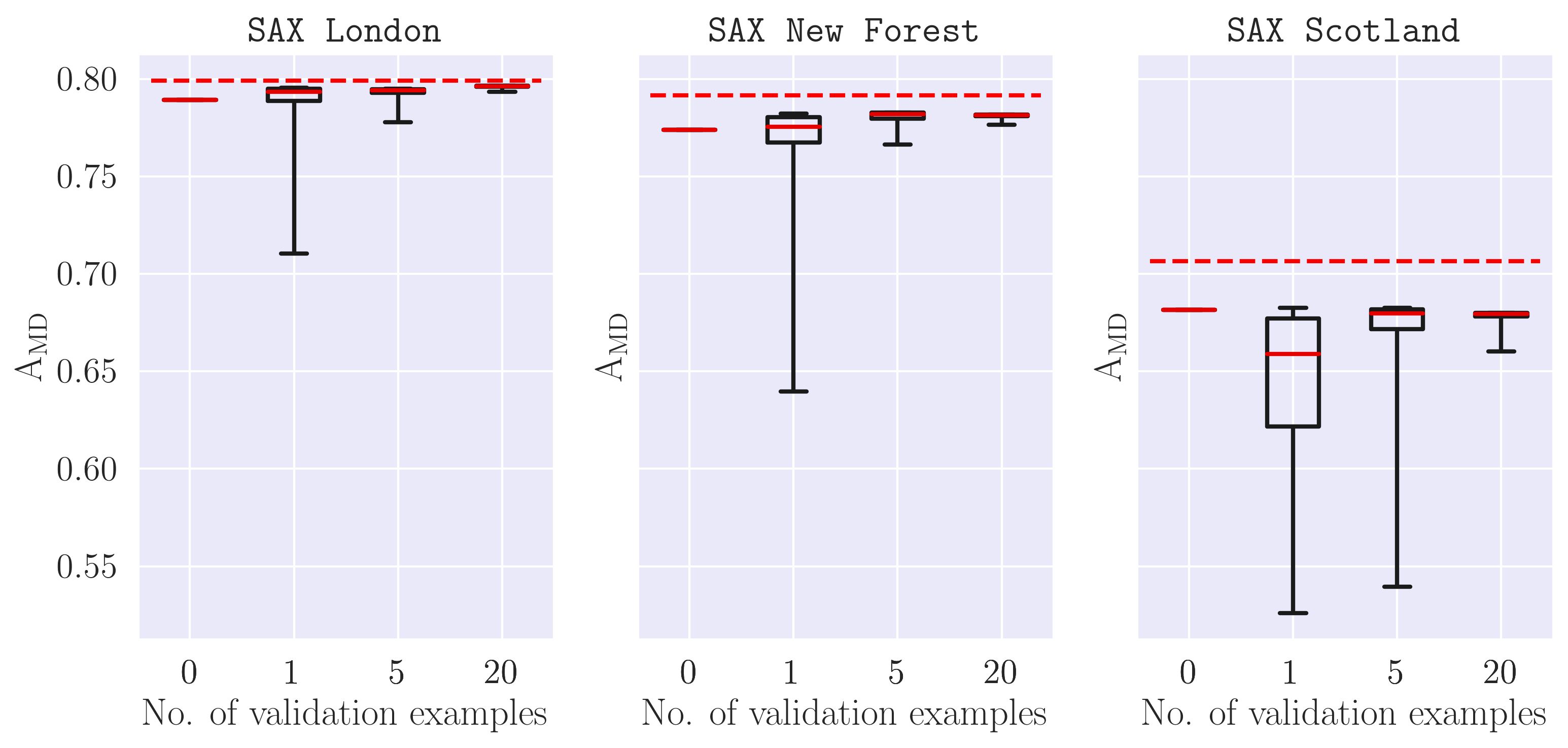}  
\caption{
\label{fig:threshold_tests_plot}
Box plot representing the achieved \Amd by calculating the uncertainty threshold with varying numbers of validation examples for a \gammassl model.
The dashed lines represent the values of \Amd achieved when using the entire test dataset to calculate the optimal threshold, before then testing on it.
% by testing on all validation examples, when also using all validation examples to calculate the optimal threshold.
}
\end{figure}

It is clear from \cref{fig:main_results_plot} that there typically exists a threshold such that accurate and inaccurate pixels are optimally separated (according to \Amd and \fhalf).
Calculation of this threshold typically requires a set of validation images from the test domain, which reduces the number of images available for testing.
Additionally, if this small set of validation images is not representative of the test dataset, then the calculated threshold will result in significantly worse misclassification detection performance.
% It is thus clear that the smaller the required number of validation examples the better.

The effect of the number of validation examples on misclassification detection performance for the \gammassl models is investigated by calculating the metrics discussed in \cref{subsec:metric_definitions} for a given withheld validation set on the remaining test images.
The results for this are averaged over 100 trials, where for each the withheld validation set is selected randomly.
Given that the method in this work calculates a threshold parameter during training, it is also possible for it to use this threshold during testing, such that \emph{zero} validation examples are required.
The mean and variability of these metrics is presented as a box plot in \cref{fig:threshold_tests_plot}.

Firstly, these plots demonstrate that the smaller the number of chosen validation images, the less well they represent the test dataset; therefore, the more variable the test performance.
More importantly, they show that there is a minimal decrease in performance between using 20 validation images and calculating the threshold with 0, as per \cref{subsec:calculating_gamma}.
This means the \gammassl methods are successfully able to calculate an appropriate threshold for the misclassification detection task without using \emph{any} validation examples.

% As the distributional shift increases, the optimality of the thresholds calculated with each number of validation images decreases.

\subsection{Cross-Domain Threshold Testing}     \label{subsec:cross_domain_thresholds}
The aim of this work is to propose a model that can estimate its mistakes with a feature-space distance threshold as the data distribution changes.
It is therefore important that this threshold calculated for one domain is effective for all domains, rather than needing a specifically optimal threshold for each domain. 

In order to investigate this (see \cref{tab:cross_domain_thresholds}), we compare the \fhalf results for the optimal threshold value for a test domain and the value calculated from the target domain the model is trained on, e.g. for $\mathrm{\gamma}\text{-}\mathrm{SSL_{iL}}\text{-}\mathrm{LDN}$ (trained on unlabelled SAX London data) the threshold corresponding to the maximum \fhalf score is used testing on the SAX New Forest and Scotland datasets and the corresponding \fhalf scores are shown.

\cref{tab:cross_domain_thresholds} shows that a threshold optimal for one domain degrades performance only very slightly for another domain.

\begin{table}[]
\caption{Cross-Domain Threshold Testing Results\label{tab:cross_domain_thresholds}}
\begin{subtable}{\columnwidth}\
\vspace{-0.3cm}
\centering
\caption{$\mathrm{\gamma}\text{-}\mathrm{SSL_{iL}}\text{-}\mathrm{LDN}$}
\vspace{0.1cm}
\begin{tabular}{cccc}
\hline
      & \maxfhalf & \fhalf with $\gamma$-{\text{LDN}} & $\Delta$ \\ \hline
LDN  &      0.8930    &        0.8930         &           0\%          \\
NF   &     0.8827     &       0.8822          &          $-$0.058\%           \\
SCOT &      0.7853    &        0.7839         &         $-$0.175\%           
\end{tabular}
\label{tab:cross_domain_thresholds_ldn}
\end{subtable}

\begin{subtable}{\columnwidth}\
\centering
\caption{$\mathrm{\gamma}\text{-}\mathrm{SSL_{iL}}\text{-}\mathrm{NF}$}
\vspace{0.1cm}
\begin{tabular}{cccc}
\hline
& \maxfhalf & \fhalf with $\gamma$-{\text{NF}} & $\Delta$ \\ \hline
LDN  &     0.8834     &        0.8832         &            $-$0.017\%          \\
NF   &      0.8849    &         0.8849        &            0\%         \\
SCOT &      0.7952     &        0.7949         &            $-$0.038\%        
\end{tabular}
\label{tab:cross_domain_thresholds_nf}
\end{subtable}

\begin{subtable}{\columnwidth}\
\centering
\caption{$\mathrm{\gamma}\text{-}\mathrm{SSL_{iL}}\text{-}\mathrm{SCOT}$}
\vspace{0.1cm}
\begin{tabular}{cccc}
\hline
& \maxfhalf & \fhalf with $\gamma$-{\text{SCOT}}  & $\Delta$ \\ \hline
LDN  &    0.8768      &        0.8764         &           $-$0.038\%          \\
NF   &     0.8806     &        0.8805         &              $-$0.007\%       \\
SCOT &      0.8257    &        0.8257         &          0\%          
\end{tabular}
\label{tab:cross_domain_thresholds_scot}
\end{subtable}
\end{table}

\subsection{WildDash Results} \label{subsec:wilddash_results}
So far in this work, the \gammassl and \gammasslil models have been trained on the SAX domains with operating conditions different to that of Cityscapes, and tested on these same SAX domains.
% By training on the SAX unlabelled datasets, the \gammassl and \gammasslil models have been exposed to domains with operating conditions different to that of Cityscapes, 
In this section, the models are also tested on the WildDash dataset \cite{wilddash}, in order to investigate how the models generalise to a test dataset with different operating conditions to both the labelled \emph{and} unlabelled training data.
This dataset does not define a single domain but includes images from a diverse set of domains, including: different weather conditions, day/night, and geographic locations from across the world.
Misclassification detection performance on this dataset is, therefore, a measure of how well these models can detect error due to \gls{ood} instances unlike anything seen in the labelled or unlabelled training datasets, or how specific they are to the domain of the unlabelled training data.

\begin{figure*}[h!]
\centering
\includegraphics[width=0.9\linewidth]{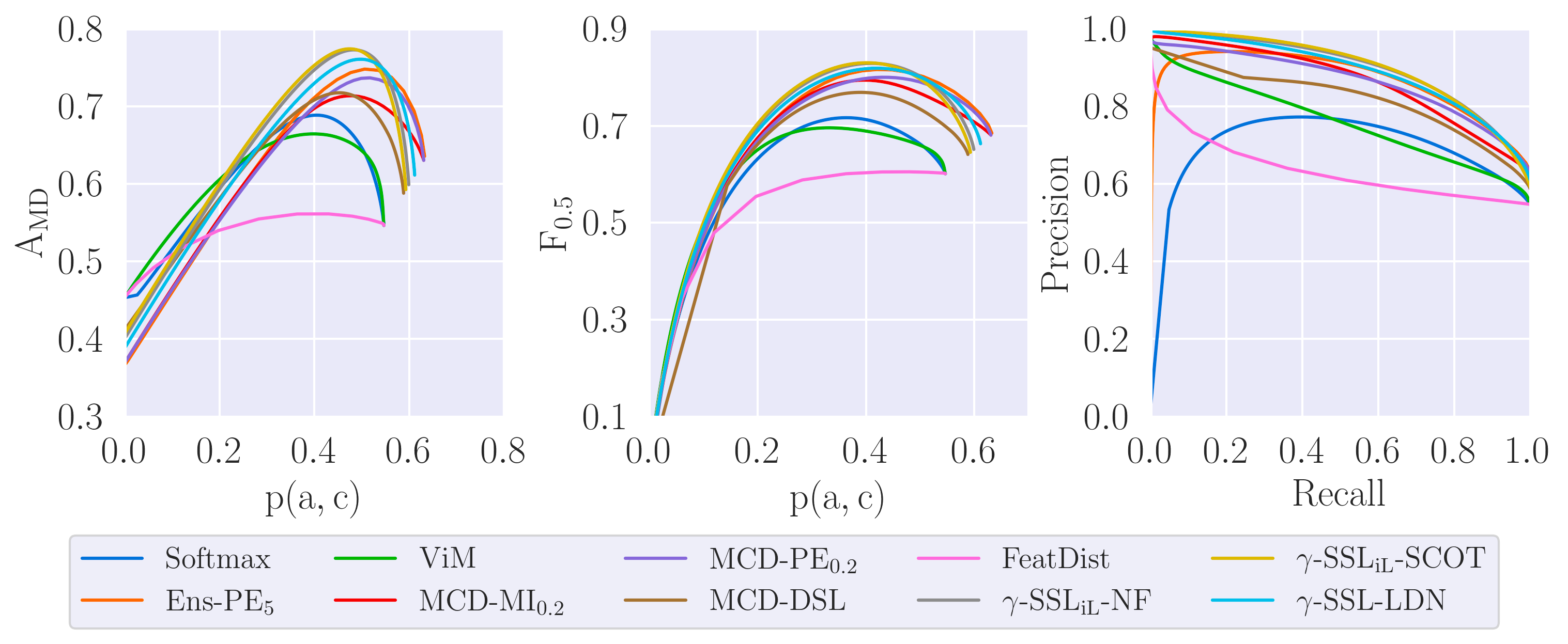}  
\caption{
\label{fig:wilddash_plot}
Misclassification detection results on the WildDash Dataset \cite{wilddash}.
$\mathrm{\gamma}$-$\mathrm{SSL}$-LDN refers to a \gammassl trained on the SAX London unlabelled dataset, whereas $\mathrm{\gamma}$-$\mathrm{SSL_{iL}}$-NF, $\mathrm{\gamma}$-$\mathrm{SSL_{iL}}$-SCOT refer to \gammasslil models that are trained on the SAX New Forest and SAX Scotland unlabelled datasets, while also using SAX London as part of a curriculum.
Best viewed in color.
}
\end{figure*}

As shown in \cref{fig:wilddash_plot}, $\mathrm{\gamma}$-$\mathrm{SSL_{iL}}$-$\mathrm{SCOT}$ outperform all benchmarks, with an \auroc and \aupr of \SI{0.852}{} and \SI{0.896}{}, compared with the best benchmark, $\mathrm{Ens}\text{-}\mathrm{PE}_5$, returning \SI{0.803}{} and \SI{0.868}{}.

Firstly, this demonstrates that although the \gammasslil models have been trained to mitigate error in specific operating conditions based on geographic location, they can also effectively detect error due to never-before-seen conditions.
Secondly, given that these values for $\mathrm{\gamma}$-$\mathrm{SSL_{iL}}$-$\mathrm{SCOT}$ are lower than the values for the SAX test domains, it also demonstrates that the best performance is reached when an unlabelled training dataset is collected from the same domain as the test data.

\subsection{Timing Results}  \label{subsec:latency_results}
In \cref{tab:timing_results}, the frequency at which each method can operate is presented on differing hardware, as a low latency is a key characteristic for robotic perception systems.
The GPU tested upon is a NVIDIA V100 GPU, while the CPU is on Macbook Pro with M2 Pro CPU. 
The $\mathrm{Vanilla}$ method is a DeepLabV3+ segmentation network \cite{deeplabv3plus} detailed in \cref{subsec:net_arch}, with the difference from $\mathrm{Ours}$ being that it uses a convolutional layer instead of prototype segmentation, and therefore also does not contain a projection network.
The timing results for $\mathrm{Vanilla}$ therefore represent all of the representation-based methods described in in \cref{subsec:benchmarks}.
The $\mathrm{MCD}$ and $\mathrm{Ens}$ methods are the same as those described in \cref{subsec:benchmarks}, (i.e. 5 and 10 member ensembles and \SI{8}{} samples for the $\mathrm{MCD}$ method).
The superscript $\mathrm{LM}$ and $\mathrm{HM}$ relate to different inference methods for the ensembles, with the former (\textbf{L}ow \textbf{M}emory) only loading one network into GPU memory at a time, while the latter (\textbf{H}igh \textbf{M}emory) loads every member network at once, while still performing inference sequentially.

These results demonstrate that our proposed method operates at significantly lower latency than the Monte Carlo Dropout and Ensemble methods, and is no slower than the segmentation network from which our method is built.

\begin{table}[h]
\centering
\caption{Timing Results}
\begin{tabular}{ccc}
\hline
Method  & GPU {[}Hz{]} & CPU {[}Hz{]} \\ \hline
$\mathrm{Vanilla}$ & 159.12       & 1.53        \\
$\mathrm{MCD}$     & 19.62        & 0.46        \\
$\mathrm{Ens}$-$\mathrm{5}^{\mathrm{LM}}$   & 5.27  & 0.75       \\
$\mathrm{Ens}$-$\mathrm{5}^{\mathrm{HM}}$   & 27.22 & 0.75       \\
$\mathrm{Ens}$-$\mathrm{10}^{\mathrm{LM}}$ & 2.99 & 0.38      \\
$\mathrm{Ens}$-$\mathrm{10}^{\mathrm{HM}}$ & 16.64 & 0.38      \\
$\mathrm{Ours}$    & 183.37       & 1.52        \\ \hline
\end{tabular}
\label{tab:timing_results}
\end{table}

\subsection{Qualitative results}\label{sec:qual_res}

\begin{figure*}
    \centering
    \begin{subfigure}{0.66\textwidth}
        \includegraphics[width=\textwidth]{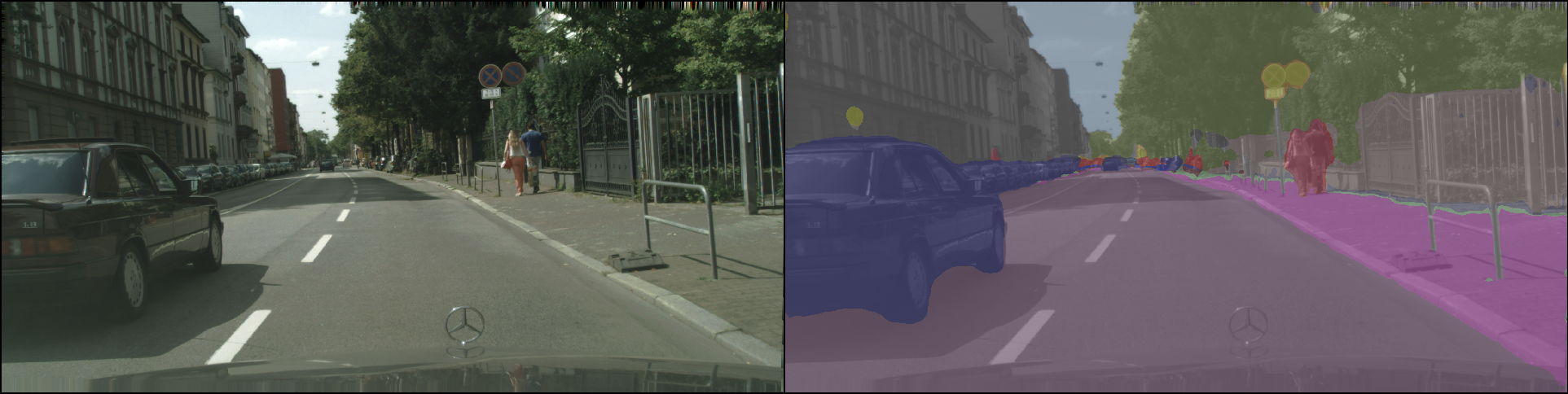}
        \caption{Cityscapes}
    \end{subfigure}
    
    \begin{subfigure}{0.66\textwidth}
        \includegraphics[width=\textwidth]{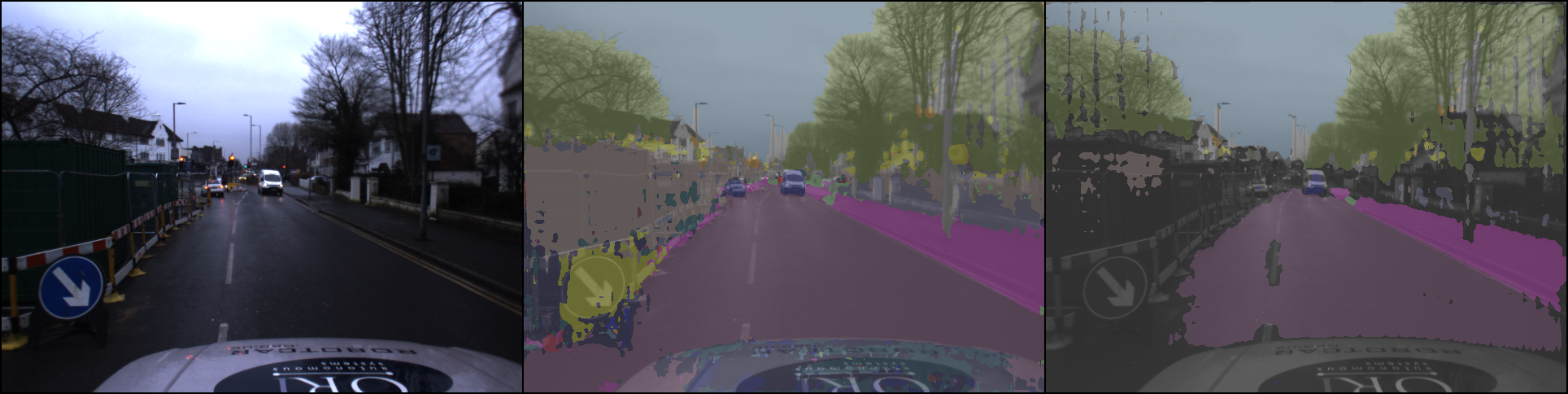}
    \end{subfigure}
    
    \begin{subfigure}{0.66\textwidth}
        \includegraphics[width=\textwidth]{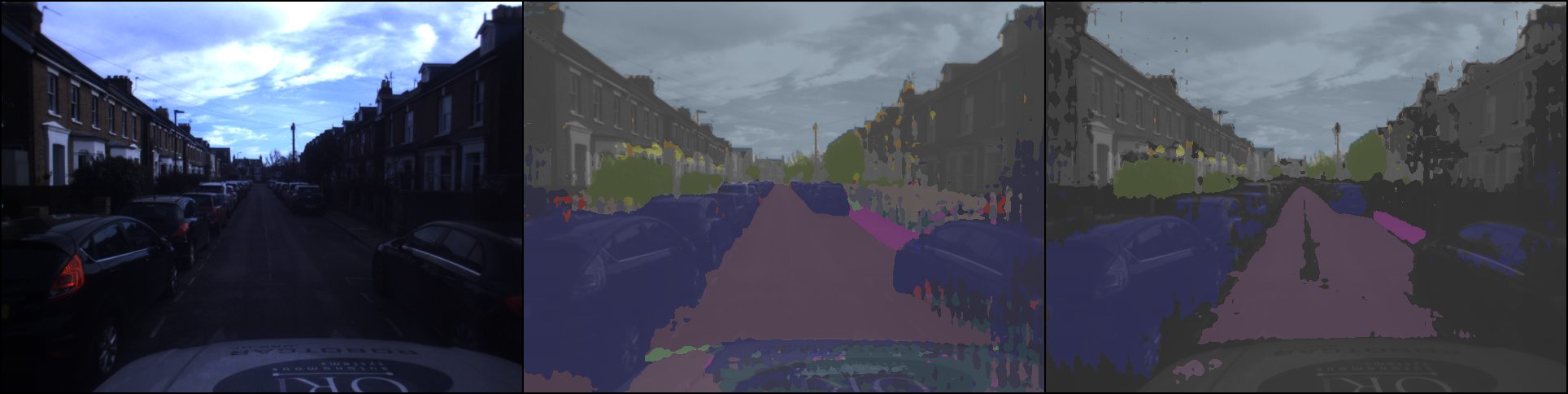}
        \caption{London}
    \end{subfigure}
    
    \begin{subfigure}{0.66\textwidth}
        \includegraphics[width=\textwidth]{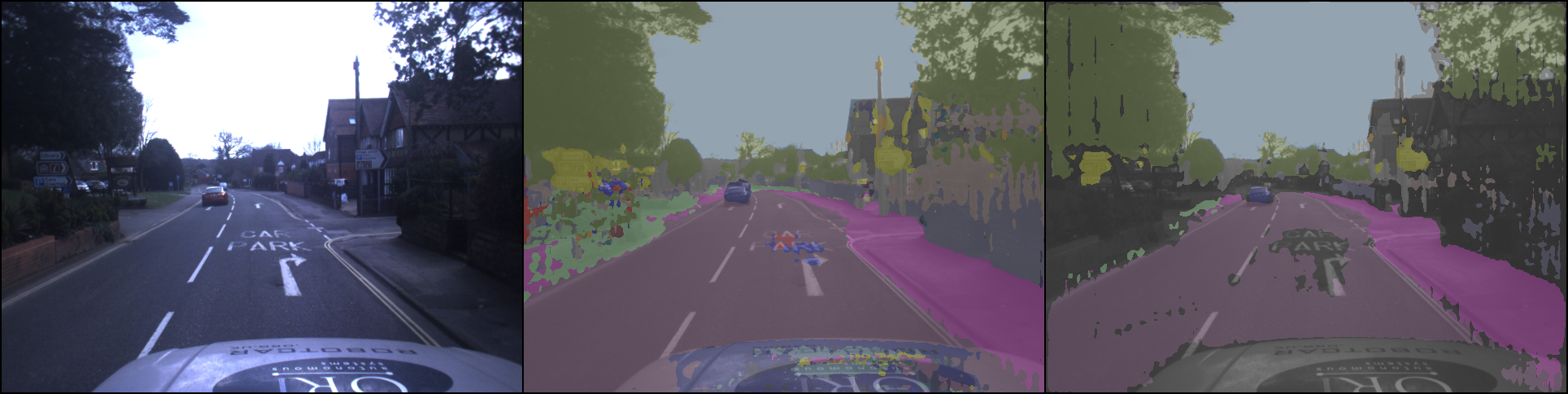}
    \end{subfigure}
    
    \begin{subfigure}{0.66\textwidth}
        \includegraphics[width=\textwidth]{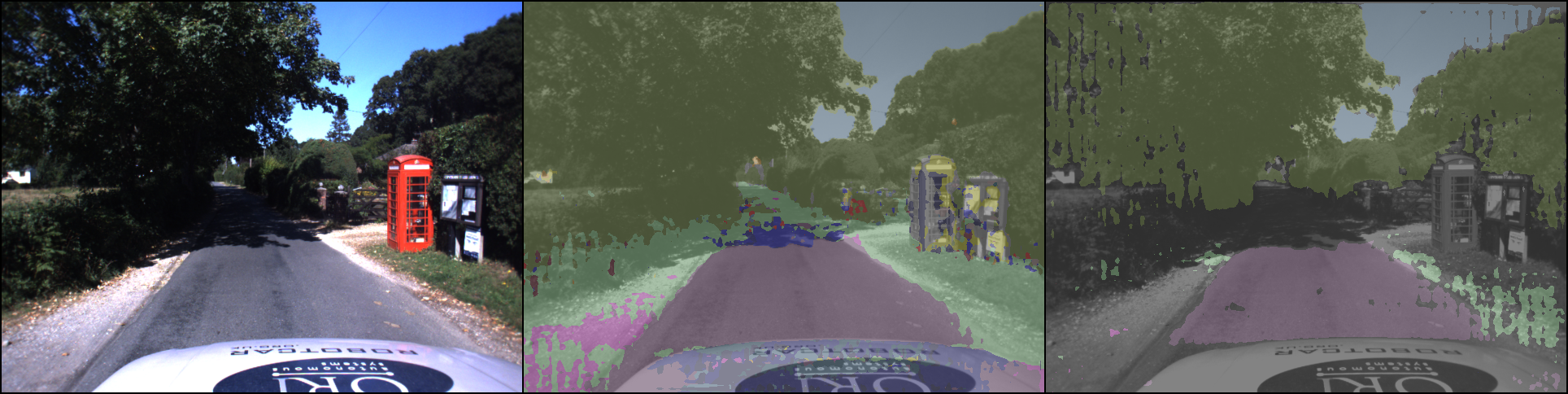}
        \caption{New Forest}
    \end{subfigure}
    
    \begin{subfigure}{0.66\textwidth}
        \includegraphics[width=\textwidth]{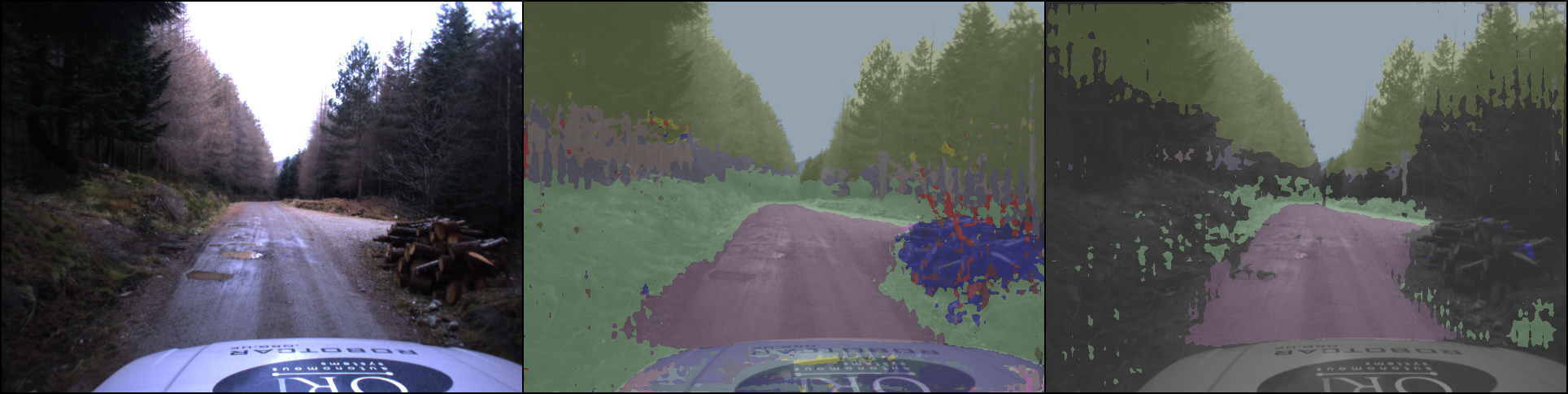}
    \end{subfigure}
    
    \begin{subfigure}{0.66\textwidth}
        \includegraphics[width=\textwidth]{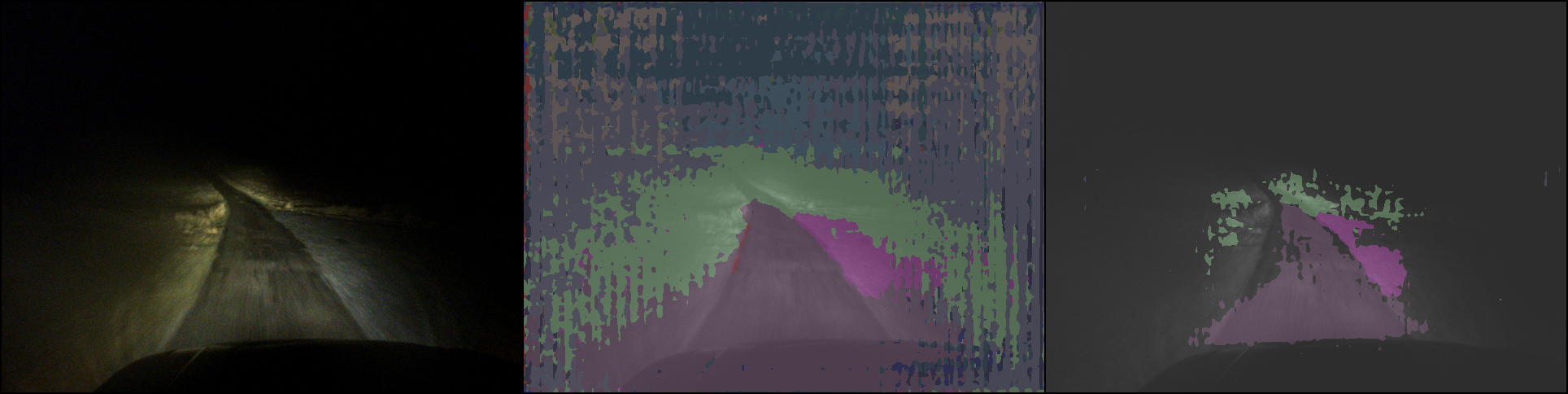}
        \caption{Scotland}
    \end{subfigure}
    
    \caption{Qualitative results for Cityscapes and the SAX domains. As the SAX RGB images (left) become more dissimilar from Cityscapes (from top to bottom), the corresponding semantic segmentations (centre) decrease in quality. \emph{However}, for these poorly segmented regions, high uncertainty is largely expressed over them, shown in black (right).}
    \label{fig:qual_results}
\end{figure*}

Presentation of qualitative results can be found in \Cref{fig:qual_results}.
Again, labels are only available in the source domain -- Cityscapes -- for which segmentations are very high-quality.
% is \textit{segmented extremely well}.
The segmentation performance degrades with increasing domain shift as expected from London $\rightarrow$ New Forest $\rightarrow$ Scotland.
Correspondingly, however, uncertainty mitigates these erroneous predictions.
Consider several examples from these samples:
\begin{enumerate}
\item The lane-obstructed street sign in the first London example, not amongst the known signs in the Cityscapes dataset. 
\item The telephone box in the second New Forest example (not in the Cityscapes domain or list of known classes),
\item The pile of timber in the first Scotland example -- classified as vehicle,
\end{enumerate}
All of these are correctly assigned high uncertainty.

\section{Ablation Studies} \label{sec:ablation_studies}
This section investigates the effect of each component of the system on misclassification-detection performance by removing key aspects of the system during training.
Results from these ablation experiments can be found in \cref{tab:ablation,fig:abl_plot}.

\begin{table}[]
    \caption{
    Ablations Misclassification Detection
    % Definitions for the names of these ablations, and further discussion of the results are found in \cref{sec:ablation_studies}.
    \label{tab:ablation}}
    \centering
    \begin{subtable}{\columnwidth}
    \resizebox{\textwidth}{!}{
    \begin{tabular}{ccccc|ccc}
    \hline
    & \multicolumn{1}{l}{} & \multicolumn{3}{c}{\auroc}  & \multicolumn{3}{c}{\aupr}   \\
    & Method               & LDN & NF & SCOT & LDN & NF & SCOT \\ 
    \hline
    \multirow{7}{*}{\rotatebox[origin=c]{90}{Ablations}} 
    & $\mathrm{NoSSL}$ & 0.803 & 0.774 & 0.72 & 0.872 & 0.829 & 0.686 \\
    & $\mathrm{NoSAX}$ & 0.793 & 0.771 & 0.669 & 0.824 & 0.794 & 0.455 \\
    & $\mathrm{M_{\gamma=-\infty}}$ & 0.805 & 0.761 & 0.711 & 0.9 & 0.827 & 0.646 \\
    & $\mathrm{Sym}$-$\mathrm{Param}$ & 0.851 & 0.598 & 0.622 & 0.8 & 0.539 & 0.486 \\
    & $\mathrm{Sym}$-$\mathrm{Non}$-$\mathrm{Param}$ & 0.643 & 0.643 & 0.621 & 0.382 & 0.235 & 0.165 \\
    & $\mathrm{NoRegL}$ & 0.815 & 0.827 & 0.72 & 0.882 & 0.847 & 0.621 \\
    & $\mathrm{MCD}$-$\mathrm{SSL}$ & 0.776 & 0.81 & 0.705 & 0.794 & 0.857 & 0.592 \\
    % & $\mathrm{p(c)}{=}{const.}$ & 0.835 & 0.879 & 0.806 & 0.913 & 0.917 & 0.741 \\
    \hline
     \multirow{2}{*}{\rotatebox[origin=c]{90}{Ours}} & $\mathrm{\gamma}\text{-}\mathrm{SSL}$ &  \textbf{0.895}  &  {0.88}   & {0.776} & \textbf{0.949} & {0.921} &  0.726  \\
     & $\mathrm{\gamma}\text{-}\mathrm{SSL_{iL}}$ &  -  &  \textbf{0.88}   & \textbf{0.859} & - & \textbf{0.942} &  \textbf{0.887}  \\
     \hline
    \end{tabular}
    }
    \caption{\label{tab:abl_auroc_aupr}}
    \end{subtable}
    \vspace{5pt}
    
    \begin{subtable}{\columnwidth}
    \resizebox{\textwidth}{!}{
    {\begin{tabular}{ccccc}
    \hline
    & \multicolumn{1}{l}{} & \multicolumn{3}{c}{$\mathrm{Max\mathrm{A_{MD}}}$ @ $\mathrm{p(a,c)}$}        \\
    & Method               & LDN & NF & SCOT \\ \hline
    \multirow{7}{*}{\rotatebox[origin=c]{90}{Ablations}}
    & $\mathrm{NoSSL}$ & 0.727 @ 0.482 & 0.706 @ 0.424 & 0.680 @ 0.204 \\
    & $\mathrm{NoSAX}$ & 0.729 @ 0.465 & 0.717 @ 0.463 & 0.653 @ 0.003 \\
    & $\mathrm{M_{\gamma=-\infty}}$ & 0.754 @ 0.586 & 0.692 @ 0.426 & 0.685 @ 0.174 \\
    & $\mathrm{Sym}$-$\mathrm{Param}$ & 0.790 @ 0.253 & 0.596 @ 0.368 & 0.756 @ 0.053 \\
    & $\mathrm{Sym}$-$\mathrm{Non}$-$\mathrm{Param}$ & 0.752 @ 0.015 & \st{0.864 @ 0.000} & \st{0.902 @ 0.000} \\
    & $\mathrm{NoRegL}$ & 0.766 @ 0.552 & 0.748 @ 0.391 & 0.665 @ 0.201 \\
    & $\mathrm{MCD}$-$\mathrm{SSL}$ & 0.747 @ 0.594 & 0.748 @ 0.482 & 0.707 @ 0.153 \\
    % & $\mathrm{p(c)}{=}{const.}$ & 0.782 @ 0.603 & 0.795 @ 0.466 & 0.758 @ 0.234 \\
    \hline
    \multirow{2}{*}{\rotatebox[origin=c]{90}{Ours}} & $\mathrm{\gamma}\text{-}\mathrm{SSL}$     &  \textbf{0.83} @ 0.625  &   0.796 @ 0.483   &   0.716 @ 0.260 \\
    & $\mathrm{\gamma}\text{-}\mathrm{SSL_{iL}}$   & - & \textbf{0.815} @ 0.608 & \textbf{0.781} @ 0.431 \\
    \hline
    \end{tabular}
    }
    }
    \begin{tablenotes}
        \fix{\item \textit{Struck through results are discounted as no pixels that are both accurate and certain are found.}}
    \end{tablenotes}
    \caption{\label{tab:abl_Amd}}
    \end{subtable}
    \vspace{5pt}
    
    \begin{subtable}{\columnwidth}
    \resizebox{\textwidth}{!}{
    \begin{tabular}{ccccc}
    \hline
    & \multicolumn{1}{l}{} & \multicolumn{3}{c}{$\mathrm{Max\mathrm{F_{0.5}}}$ @ $\mathrm{p(a,c)}$}        \\
    & Method               & LDN & NF & SCOT \\ \hline
    \multirow{7}{*}{\rotatebox[origin=c]{90}{Ablations}}
    & $\mathrm{NoSSL}$ & 0.790 @ 0.380 & 0.753 @ 0.364 & 0.626 @ 0.203 \\
    & $\mathrm{NoSAX}$ & 0.769 @ 0.401 & 0.756 @ 0.403 & 0.505 @ 0.220 \\
    & $\mathrm{M_{\gamma=-\infty}}$ & 0.826 @ 0.492 & 0.746 @ 0.353 & 0.602 @ 0.184 \\
    & $\mathrm{Sym}$-$\mathrm{Param}$ & 0.731 @ 0.219 & 0.608 @ 0.381 & 0.495 @ 0.057 \\
    & $\mathrm{Sym}$-$\mathrm{Non}$-$\mathrm{Param}$ & 0.394 @ 0.088 & 0.281 @ 0.030 & 0.210 @ 0.017 \\
    & $\mathrm{NoRegL}$ & 0.806 @ 0.507 & 0.781 @ 0.332 & 0.569 @ 0.241 \\
    & $\mathrm{MCD}$-$\mathrm{SSL}$ & 0.809 @ 0.518 & 0.804 @ 0.416 & 0.589 @ 0.159 \\
    % & $\mathrm{p(c)}{=}{const.}$ & 0.853 @ 0.513 & 0.851 @ 0.396 & 0.706 @ 0.213 \\
    \hline
    \multirow{2}{*}{\rotatebox[origin=c]{90}{Ours}} & $\mathrm{\gamma}\text{-}\mathrm{SSL}$     &  \textbf{0.893} @ 0.548  &   0.855 @ 0.407    &   0.678 @ 0.239 \\
    & $\mathrm{\gamma}\text{-}\mathrm{SSL_{iL}}$ &  -  & \textbf{0.885} @ 0.532  & \textbf{0.826} @ 0.370 \\
    \hline
    \end{tabular}
    }
    \caption{\label{tab:abl_fhalf}}
    \end{subtable}
    
\end{table}

\subsection{Is distance to prototypes a good measure of uncertainty?}
This is investigated by updating the model using only the source images through the supervised loss $L_{\text{s}}$, i.e. no semi-supervised learning takes place.
In this way, the mechanism for generating uncertainty estimates, i.e. calculating the cosine similarity with the prototypes, can be investigated.
Notably, except for \gls{dum}, the same model weights are shared between our method and the representation-based benchmarks.
Comparing the results in \cref{tab:auroc}, \cref{tab:aupr} to \cref{tab:abl_auroc_aupr}, this ablation, named \nossl, outperforms the other representation-based methods in \auroc and \aupr.

These results support our argument that the method for calculating $M_{\gamma}$ is a good inductive bias and helps to set up the positive feedback loop discussed in \cref{sec:training}.
The fact that \nossl performs significantly worse than \gammassl and \gammasslil for each metric suggests that the method's success relies on the representation learned on unlabelled target domain data.

\subsection{Is the target domain data required?}
In this investigation, Cityscapes data is used for the semi-supervised task instead of the SAX unlabelled data (hence is called \nosax). 
The objective is to determine whether the proposed method is leveraging the unlabelled target domain data or our uncertainty estimation results are due only to the semi-supervised task and objectives.

Compared with \gammassl and \gammasslil, the \auroc and \aupr results in \cref{tab:abl_auroc_aupr} show a worse misclassification detection performance for \nosax on the SAX test datasets.
This confirms the utility of collecting large, diverse datasets containing near-distribution and \gls{ood} instances, as this ablation empirically shows that using such a dataset during training improves the detection of \gls{ood} instances during testing.

% \nossl, no semi-supervised learning is performed, and the model is only updated using $L_{\text{s}}$.
% This means that the mechanism for generating uncertainty estimates in this work is tested with respect to the benchmarks, independent of the learned representation.
% \cref{tab:auroc-aupr} shows that \nossl performs comparably to \softmax, and significantly better than \featdist.
% The latter suggests that using cosine distance performs better than Mahalanobis distance.
% The former suggests that the success of the \gammassl model is in it learned representation.

% \subsection{Learned loss attenuation}
\subsection{Is $M_\gamma$ required to learn uncertainty estimation?}
% \subsection{Effect of maximising consistency un-selectively}
This experiment, named \nofiltering, trains a model by maximising the consistency between \emph{all} pixels.
This, therefore, investigates the loss function in \cref{eq:Lc} and whether standard semi-supervised training is sufficient to learn a good representation for uncertainty estimation.

\begin{figure*}[h!]
\centering
\includegraphics[width=.8\linewidth]{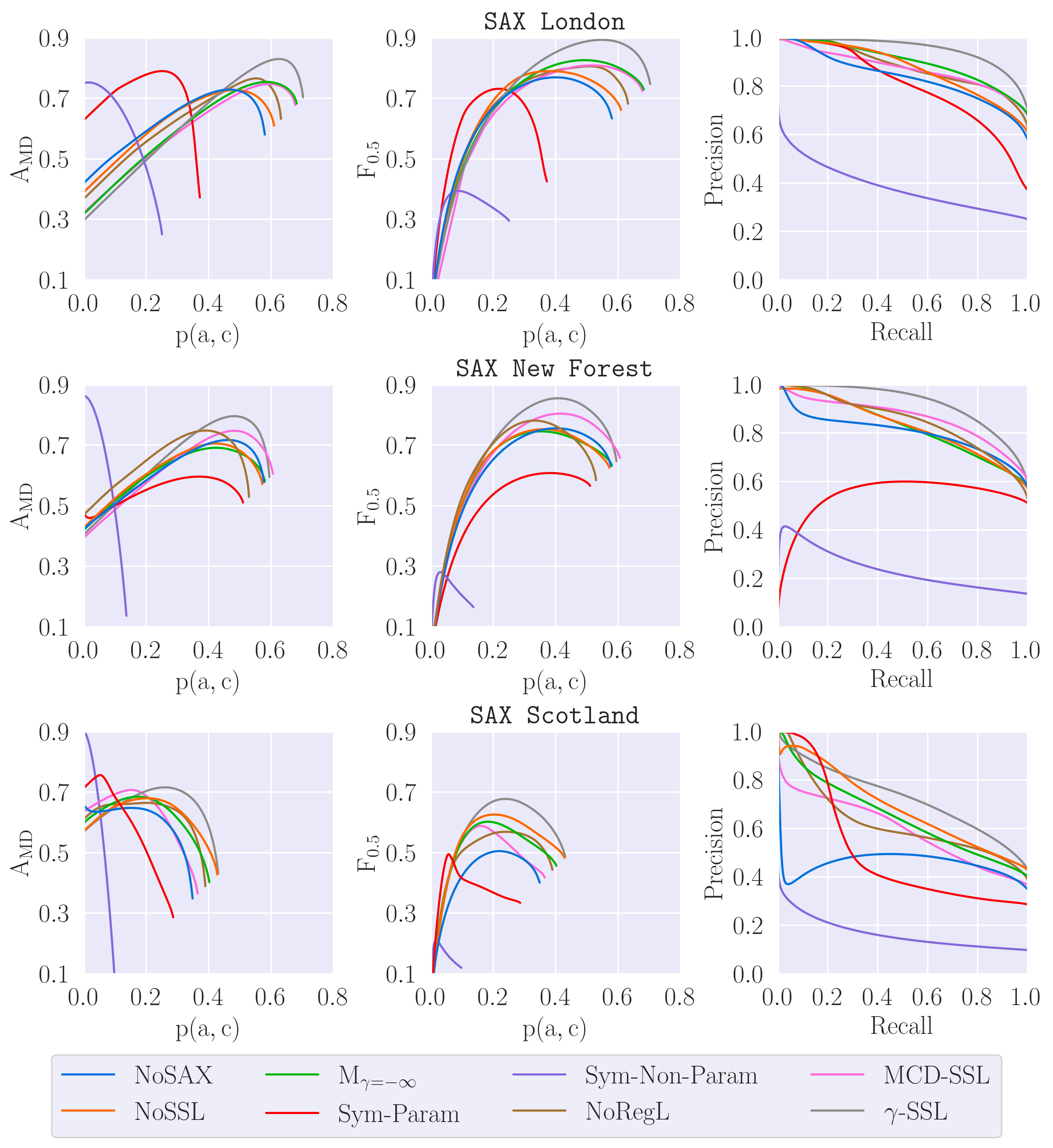}  
\caption{
\label{fig:abl_plot}
Misclassification results for ablated \gammassl models. Given that the ablations are performed on the \gammassl models, the \gammasslil models are not plotted. See \cref{sec:ablation_studies} for details.
Best viewed in color.
}
\end{figure*}

The results consistently show that the proposed \gammassl performs better at misclassification detection on each metric.
This suggests that attenuating the loss for \uncertain pixels facilitates learning a representation suitable for uncertainty estimation.
Additionally, the results in \Cref{fig:abl_plot} show that the segmentation accuracy is lower for this ablation (remembering that $\mathrm{max}\mathrm{[p(a, c)]}$ is the segmentation accuracy), showing it is beneficial to use $M_{\gamma}$ to filter out noise in the semi-supervised consistency task introduced by the challenging, uncurated nature of the unlabelled training images.

% \subsection{Effect of training only on unlabelled source data} 
% \nosax: the system is trained with Cityscapes images as the target images $x_T$. 
% \cref{tab:abl_max_A_MD} shows an almost 8\% increase in \maxAmd with respect to \nossl despite using the same training data.
% The semi-supervised training with $\gamma$ therefore massively improves uncertainty estimation performance, independent of the target training data used.
% Also shown is that the additional target data from the test domain does increase \maxAmd by another 3\%.

\subsection{Does branch asymmetry prevent feature collapse?}
$\mathrm{Sym}$-$\mathrm{Non}$-$\mathrm{Param}$ and $\mathrm{Sym}$-$\mathrm{Param}$ produce segmentations using only non-parametric prototype segmentation and a parameterised segmentation head respectively, and are thus both symmetric, unlike our system (see \cref{subsec:asymm}).

For both methods, the models nearly always suffered feature collapse, where each feature is embedded near a single class prototype ($\mathrm{Road}$ in this case).
The exception is for $\mathrm{Sym}$-$\mathrm{Proto}$ on the SAX New Forest dataset.
When collapse occurs, the key observation is that segmentation accuracy greatly deteriorates, even if \auroc and \aupr look acceptable.
Firstly, this ablation provides evidence that the asymmetric branches successfully prevent this type of failure.
Secondly, it confirms that looking at \auroc and \aupr alone is clearly insufficient to fully evaluate the model, and that integrating segmentation quality into the metrics is useful.
% The latter is because misclassification detection becomes an easier task if almost all pixels are incorrect.

\subsection{Do the additional losses provide useful regularisation?}
By removing both $L_{\text{u}}$ and $L_{\text{p}}$ for $\mathrm{NoRegL}$, the effect of these additional losses can be investigated.
The result is not a complete feature collapse but a deterioration of misclassification-detection and segmentation performance on every metric.
This suggests that by spreading out features and prototypes on the unit-hypersphere, distance to prototypes is a better measure of uncertainty, and the classes in the target domain become more separable.

% $\mathrm{NoRegL}$: removes both $L_{\text{u}}$ and $L_{\text{p}}$.
% This also experiences feature collapse, however in this case $\mathrm{p(certain)} \to 0$, thus attenuating $L_\text{c}$ fully, and so the representation is not improved by the target data.

% \subsection{Model perturbation vs. Data Augmentation}
\subsection{Is data augmentation the best way of obtaining a distribution over possible segmentations?}
\Cref{sec:which_uncertainty} discusses how distributional uncertainty can be treated as inherent to the data rather than the model, thus motivating using data augmentation rather than model perturbation to obtain a distribution over possible segmentations of an image.
This ablation, $\mathrm{MCD}\text{-}\mathrm{SSL}$, investigates training the model using dropout instead of data augmentation to calculate $M_{c}$.
The dropout probability used was \SI{0.2}{}, as this performed best for both misclassification-detection and segmentation performance for the $\mathrm{MCD_{0.2}}$ benchmarks.

$\mathrm{MCD}\text{-}\mathrm{SSL}$ does not achieve as good misclassification-detection performance as \gammassl, as demonstrated on every metric in \cref{tab:ablation}.
This provides evidence that data augmentation is a good method for obtaining segmentation distributions representing the likelihood of correct class assignments.

\section{Additional Ablation Studies}
In this section differing training procedures are experimented with in order to investigate their performance relative to our proposed method.
The experiments involve using the Cityscapes dataset as the labelled dataset, and SAX London data as the unlabelled dataset, with test results reported on the SAX London Test dataset.

\subsection{Does a ``soft'' $M_\gamma$ help training?}      \label{subsec:soft_m_gamma}

In this experiment, the certainty mask is no longer binary, but instead the confidence is expressed as the max softmax score, $M_\gamma^{(i)} = \mathrm{norm}(\max [ \sigma_{\tau}(s^{(i)}) ]) \in \{0,1\}$, where $\mathrm{norm}$ is a function that normalises a batch of certainty masks such that the lowest pixel confidence is 0, and the highest is 1, see \cref{subsec:proto_seg} for more details.

% We trained a model in this way on Cityscapes labelled data, and SAX London unlabelled data, and then evaluated on SAX London test data.
The uncertainty estimation results for this are \maxfhalf is 0.862 at a \pac of 0.421, and with a segmentation accuracy of 0.576.
The results for an equivalent model using a binary thresholded $M_\gamma$ are as follows: 0.893 @ 0.548 for \maxfhalf @ \pac and segmentation accuracy of  0.70, thereby showing a significant drop in segmentation accuracy in the target domain, and also a drop in uncertainty estimation performance.

These results suggest that the soft $M_\gamma$ introduces noise into the consistency task on the unlabelled domain, and prevents the learning of a high-quality representation of this domain.

% \dw{Why?}

\subsection{Should each class prototype have a different threshold?}        \label{subsec:gamma_per_class}
% In this training experiment, the binary certainty mask, $M_\gamma$ is calculated by applying a different threshold $\gamma$ for each class.
This experiment evaluates whether learned uncertainty estimation can be improved by thresholding cosine distance between a class prototype and a feature with a different value for each class.

The rationale for this is that datasets often represent different classes with different levels of occurrence and diversity, therefore the statistics of the each class's representation may also vary, e.g. more diversely represented classes may have a greater variance.

In order to account for this in this experiment, the certainty mask $M_\gamma$ is calculated from per-class thresholds $\Gamma = [\gamma_1, \gamma_2, ...\gamma_K]$ as follows:
\begin{equation}\label{eq:certainMask}
    % M_\gamma^{(i)} = \mathbbm{1}[s_{1:K}^{(i)}>\gamma]
    % M_\gamma^{(i)} = \Gamma [\mathrm{argmax}(s^{(i)})]
    % M_\gamma^{(i)} = \mathbbm{1}[s_{1:K}^{(i)} > \gamma ]
    M_\gamma^{(i)} = \mathbbm{1}[\mathrm{max}(s^{(i)}) > \gamma_{k=\mathrm{argmax}(s^{(i)})} ]
\end{equation}
where $s^{(i)} \in \R^{K}$ are the segmentation scores for a pixel $i$.

The per-class thresholds $\Gamma$ are calculated such \fixM{that}{2.5\label{comm:2.5_1}} the per-class consistency $[p_c]_k$ is equal to the per-class certainty $[p_\gamma]_k$, with both calculated as follows:

\fixM{
\begin{equation}
[p_\gamma]_k =  \sum_{i}^{NHW}  \frac{\mathbbm{1}[\mathrm{argmax}(s^{(i)}) = k] \odot M_\gamma^{(i)}}{\sum_{j}^{NHW}\mathbbm{1}[\mathrm{argmax}(s^{(j)}) = k]}
\end{equation}
}{2.4\label{comm:2.4_1}}

\fixM{
\begin{equation}
[p_c]_k =  \sum_{i}^{NHW}  \frac{\mathbbm{1}[\mathrm{argmax}(s^{(i)}) = k] \odot M_c^{(i)}}{\sum_{j}^{NHW} \mathbbm{1}[\mathrm{argmax}(s^{(j)}) = k]}
\end{equation}
}{2.4\label{comm:2.4_2}}

The uncertainty estimation results achieved with a model trained in this way are significantly worse than our proposed method, with a \maxfhalf @ \pac of 0.772 @ 0.432 (versus 0.893 @ 0.548 for our proposed model) and does this with a segmentation accuracy of 0.651 versus 0.703.
This suggests that solving for a per-class threshold during training negatively affects both segmentation quality and uncertainty estimation, and it is therefore preferable to solve for a single threshold as per our method.

\subsection{Is a large batch size required for calculating the prototypes during training?} \label{subsec:proto_batch_size}

For efficiency during training, the prototypes are calculated from features extracted from a batch of labelled images, whereas in testing prototypes can be calculated from the entire dataset.
This therefore raises the question as to whether the training prototypes are too noisy if the batch size becomes too small, and whether a large amount of GPU memory is required for this method.
Our proposed training procedure only uses a batch size of 12, and mitigates one aspect of this problem by using the most recent class prototype if a class is not present in the batch of labelled images (shown in \cref{tab:prototype_batch_size} as Use history?).

In order to investigate whether small batch sizes are a problem, we train a model with a smaller number of images from which to compute prototypes, while keeping the batch size for the other aspects the same.
During testing, prototypes are calculated from all available labelled images from the labelled domain.
We report metrics for both the segmentation quality (segmentation accuracy, Seg. Acc.) and uncertainty estimation performance ($\mathrm{Max\mathrm{F_{0.5}}}$ @ $\mathrm{p(a,c)}$) in \cref{tab:prototype_batch_size}.
These results show that, while still using previous prototypes where needed, reducing the prototype batch size \textit{does not} significantly affect segmentation or uncertainty estimation quality.
However when not using the history of prototypes, uncertainty estimation quality (in the form of $\mathrm{Max\mathrm{F_{0.5}}}$ @ $\mathrm{p(a,c)}$) does degrade at lower batch sizes, thereby demonstrating the usefulness of this method.

\begin{table}[h]
\caption{Results for varying Training Prototype Batch Size on SAX London}
\centering
\begin{tabular}{cccc}
\hline
  $\mathrm{Batch\ Size}$  & Use history? & $\mathrm{Max\mathrm{F_{0.5}}}$ @ $\mathrm{p(a,c)}$ & Seg. Acc. \\ \hline
12 & Yes & 0.893 @ 0.548        &  0.703 \\
8  & Yes & 0.888 @ 0.559       & 0.719 \\
6  & Yes & 0.892 @ 0.546      & 0.712 \\
2  & Yes & 0.882 @ 0.518     & 0.693 \\
2  & No & 0.827 @ 0.538     &  0.690 \\ \hline
\end{tabular}
\label{tab:prototype_batch_size}
\end{table}

%------------------------------------------------------------------
\section{Conclusion} \label{sec:concl}
%------------------------------------------------------------------
Firstly, this work presents a segmentation network that mitigates misclassification on challenging distributionally shifted test data via uncertainty estimation.
It achieves this by learning a feature representation, where pixel embeddings corresponding to accurate and inaccurate segmentations are separable by a single global threshold around prototypical class features. 
By leveraging a large quantity of uncurated unlabelled data from the deployment domain, the constraint of having labelled data from that domain is relaxed, and thus a small labelled dataset from a distinct domain can be used.
% Only a small quantity of labelled data from a domain shifted from the deployment domain is required for training, due to leveraging a large quantity of uncurated unlabelled data.
Secondly, it presents a novel semantic segmentation test benchmark, comprising a set of \SI{700}{} pixel-wise labels from three distinct domains and metrics to measure quality of uncertainty estimation.
Upon evaluation on this challenging benchmark, the presented network outperforms epistemic uncertainty estimation and out-of-distribution detection methods, and does so without increasing the computational footprint of a standard segmentation network.
% %------------------------------------------------------------------
% \section*{Acknowledgements}
% %------------------------------------------------------------------
% 
% This work was supported by the Assuring Autonomy International Programme, a partnership between Lloyd’s Register Foundation and the University of York, as well as by the EPSRC Programme Grant ``From Sensing to Collaboration'' (EP/V000748/1).

%------------------------------------------------------------------
% \newpage
\bibliographystyle{IEEEtran}
\bibliography{biblio}
%------------------------------------------------------------------

\end{document}